\documentclass{article}
\PassOptionsToPackage{numbers, compress}{natbib}
\usepackage[final]{neurips_2022}
\usepackage{natbib}
\setcitestyle{square}

\usepackage[auth-lg]{authblk}
\usepackage{blindtext}

\usepackage[utf8]{inputenc} % allow utf-8 input
\usepackage[T1]{fontenc}    % use 8-bit T1 fonts
\usepackage{hyperref}       % hyperlinks
\usepackage{url}            % simple URL typesetting
\usepackage{booktabs}       % professional-quality tables
\usepackage{amsfonts}       % blackboard math symbols
\usepackage{nicefrac}       % compact symbols for 1/2, etc.
\usepackage{microtype}      % microtypography
\usepackage{xcolor}         % colors
\usepackage{bbm}
\usepackage{booktabs}
\usepackage{bm}
\usepackage{amstext}
\usepackage{subfigure}
\usepackage{graphicx}
\usepackage{mathtools}
\usepackage{mathtools}
\usepackage{amsmath}
\usepackage{enumitem}
\usepackage{amsthm}
\usepackage{amssymb}
\usepackage{multirow}
\usepackage{wrapfig}
\usepackage{booktabs}
\usepackage{tcolorbox}
\usepackage[ruled,linesnumbered]{algorithm2e}

\setcitestyle{nonatbib}
\newtheorem{Definition}{\textbf{Definition}}
\newtheorem{Theorem}{\textbf{Theorem}}

\newtheorem{Lemma}{\textbf{Lemma}}
\newtheorem{Assumption}{\textbf{Assumption}}

\usepackage{xcolor}
\usepackage{color, soul}

\newcommand{\nop}[1]{}

\title{Revisiting Graph Contrastive Learning from the Perspective of Graph Spectrum}

\author[1]{Nian Liu}
\author[ ]{Xiao Wang$^{1,2*}$}
\author[1]{Deyu Bo}
\author[ ]{Chuan Shi$^{1,2}$\thanks{Corresponding authors.}}
\author[3]{Jian Pei}
\affil[1]{Beijing University of Posts and Telecommunications}
\affil[2]{Peng Cheng Laboratory}
\affil[3]{Simon Fraser University}
\affil[ ]{ {\{nianliu, xiaowang, bodeyu, shichuan\}@bupt.edu.cn, jpei@cs.sfu.ca}}

\begin{document}
\maketitle

\begin{abstract}
\setcounter{footnote}{0}
Graph Contrastive Learning (GCL), learning the node representations by augmenting graphs, has attracted considerable attentions. Despite the proliferation of various graph augmentation strategies, some fundamental questions still remain unclear: what information is essentially encoded into the learned representations by GCL? Are there some general graph augmentation rules behind different augmentations? If so, what are they and what insights can they bring? In this paper, we answer these questions by establishing the connection between GCL and graph spectrum. By an experimental investigation in spectral domain, we firstly find the General grAph augMEntation (GAME) rule for GCL, i.e., the difference of the high-frequency parts between two augmented graphs should be larger than that of low-frequency parts. This rule reveals the fundamental principle to revisit the current graph augmentations and design new effective graph augmentations. Then we theoretically prove that GCL is able to learn the invariance information by contrastive invariance theorem, together with our GAME rule, for the first time, we uncover that the learned representations by GCL essentially encode the low-frequency information, which explains why GCL works. Guided by this rule, we propose a spectral graph contrastive learning module (SpCo\footnote{Code available at https://github.com/liun-online/SpCo}), which is a general and GCL-friendly plug-in. We combine it with different existing GCL models, and extensive experiments well demonstrate that it can further improve the performances of a wide variety of different GCL methods.
\end{abstract}
% !TEX root =  ../main.tex

\section{Introduction}

Graph Neural Networks (GNNs) learn the node representations in a graph mainly by message passing. GNNs have attracted significant interest and found many applications~\citep{gcn, shi2020point, linmei2019heterogeneous}. Training the high quality GNNs heavily relies on task-specific labels, while it is well known that manually annotating nodes in graphs is costly and time-consuming~\citep{hu2019strategies}. Therefore, Graph Contrastive Learning (GCL) is developed as a typical technique for self-supervised learning without the explicit usage of labels~\citep{dgi, mvgrl, gca}.

\begin{figure}[ht]
\centering
\includegraphics[scale=0.18]{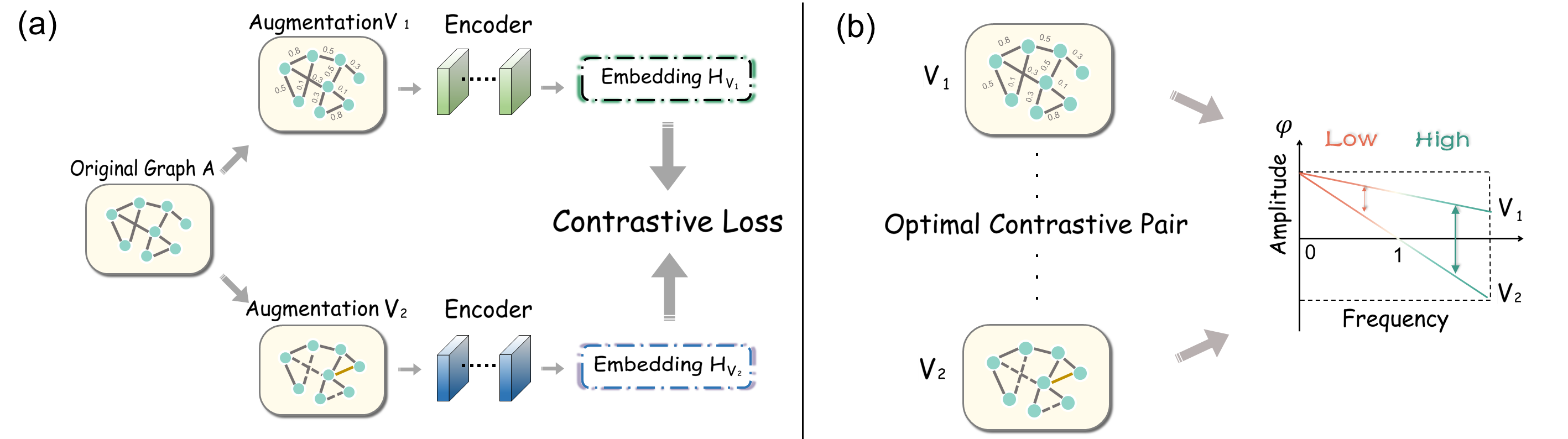}
\caption{(a) The general framework of GCL. (b) The illustration of our findings in the empirical study (Section~\ref{case study}). If two contrasted graphs have a larger margin between high-frequency part than low-frequency part, they will boost the GCL. We call such two graphs as optimal contrastive pair.}
\label{gcl_frame}
\end{figure}

The traditional GCL framework (Fig.~\ref{gcl_frame} (a)) mainly includes three components: graph augmentation, graph representation learning by an encoder, and contrastive loss. In essence, GCL aims to maximize agreement between augmentations to learn invariant representations~\citep{gca}. Typical GCL methods have sought to elaborately design different graph augmentation strategies. For example, the heuristic based methods including node or edge dropping~\citep{graphcl}, feature masking~\citep{cca}, and diffusion~\citep{mvgrl}; and the learning based methods  including InfoMin~\citep{adgcl, infogcl}, disentanglement~\citep{dgcl}, and adversarial training~\citep{yang2021graph}. Although various graph augmentation strategies are proposed, the fundamental augmentation mechanism is not well understood. \textit{What information should we preserve or discard in an augmented graph? Are there some general rules across different graph augmentation strategies? How to use those general rules to validate and improve the current GCL methods?}  This paper explores those questions.

Essentially, an augmented graph is obtained by changing some components in the original graph and thus strength of frequencies~\citep{gsp} in graph spectrum. This natural and intuitive connection between graph augmentation and graph spectrum inspires us to explore the effectiveness of augmentations from the spectral domain. We start with an empirical study (Section~\ref{case study}) to understand the importance of low-frequency and high-frequency information in GCL. 
%, where we construct the augmented graphs by retaining the low-frequency and high-frequency information of original graph with different proportions. Then we utilize the augmented graph and original graph to perform graph contrastive learning. We find 
Our findings indicate that both the lowest-frequency information and the high-frequency information are important in GCL.  Retaining more high-frequency information is particularly helpful to improve the performance of GCL. 
However, as shown in Fig.~\ref{gcl_frame} (b), the way of handling high-frequency information in two contrasted graphs $V_1$ and $V_2$ should be different, which can be finally summarized as a general graph augmentation (GAME) rule: the difference of amplitudes of high frequencies in two contrasted graphs should be larger than that of low frequencies.

%This further reveals the general augmentation rule: the difference of high-frequency information in the two graphs should be larger than that of low-frequency information for a better GCL.

%(2) However, the difference of amplitudes of high-frequency information in two graphs should be larger than those of low-frequency information in two graphs.

%\todo{A reader who does not read the rest of the paper may not be able to understand this paragraph and thus the rest of Section 1. Please revise.}
% Why is that? The answer is closely related with what information is learned by GCL. Because the general augmentation rule implies that the low-frequency information will be the common pattern between the two graphs. Furthermore, we propose the contrastive invariance theorem~\ref{case_theo}, which theoretically proves that GCL is able to learn the invariance information from original and augmented graphs. With the above analysis, we can find that the invariance information learned by GCL corresponds to the low-frequency information in the two graphs, and the usefulness of low-frequency information has been well demonstrated~\citep{entezari2020all}. The general augmentation rule not only explains why GCL works, but also provides a clear and concise demonstration of which current augmentation strategy is better. The experiments, shown in Section~\ref{The General Graph Augmentation Ru}, on nine recently proposed augmentation strategies well verify our proposed general augmentation rule.

To explain the GAME rule, we need to understand what information is encoded into the learned representations by GCL. We propose the contrastive invariance (Theorem~\ref{case_theo}), which, for the first time, theoretically proves that GCL can learn the invariance information from two contrasted graphs. Meanwhile, as can be seen in Fig.~\ref{gcl_frame} (b), because the difference of amplitudes of lowest-frequency information is much smaller than that of high-frequency information, the lowest-frequency information will be the approximately invariant pattern between the two graphs $V_1$ and $V_2$. Therefore, with such two augmentations $V_1$ and $V_2$, we can conclude that the information learned by GCL is mainly the low-frequency information, whose usefulness has been well demonstrated~\citep{entezari2020all}. This not only explains why GCL works, but also provides a clear and concise demonstration of which augmentation strategy is better, as verified by the experiments in Section~\ref{The General Graph Augmentation Ru}.

Based on our findings and theoretical analysis, we define two augmentations satisfying the GAME rule are called an optimal contrastive pair. Then, we propose a novel \textbf{sp}ectral graph \textbf{co}ntrastive learning (SpCo), a general GCL framework, which can boost existing GCL methods with optimal contrastive pairs. Specifically, to ensure that the learned augmented graph is an optimal contrastive pair with the original adjacency matrix, we need to make the amplitude of its high frequency ascend while keeping the low frequency the same as the original structure. We model this process as an optimization objective based on matrix perturbation theory, which can be solved by Sinkhorn's Iteration \citep{sinkhorn} and finally obtain the augmented structure used for the following target GCL model. 

% Our contributions are summarized as follows. Firstly, to the best of our knowledge, we are the first attempt to fundamentally explore the augmentation strategies for GCL from spectral domain. We not only reveal the general graph augmentation rule behind different augmentation strategies, but also explain why GCL works by proposing the contrastive invariance theorem. We show that the augmentation rule provides a novel insight to estimate the current augmentation strategies. Secondly, we propose a novel concept optimal contrastive pair and theoretically derive a general GCL framework SpCo, which is able to improve the performance of existing GCL methods. Last, we choose three typical GCL methods as target methods, and plug SpCo into them. Extensive experiments on five public datasets demonstrate that the plugged SpCo can consistently gain improvements compared with those original target methods.

Our contributions are summarized as follows. Firstly, we answer the question ``what information is learned by GCL and whether there exists a general augmentation rule''. To the best of our knowledge, this is the first attempt to fundamentally explore the augmentation strategies for GCL from spectral domain. We not only reveal the general graph augmentation rule behind different augmentation strategies, but also explain why GCL works by proposing the contrastive invariance theorem. Our work provides deeper understanding on the nature of GCL. Secondly, we answer the question ``how to utilize the augmentation rule for GCL''. We show that the augmentation rule provides a novel insight to estimate the current augmentation strategies. We propose a novel concept optimal contrastive pair and theoretically derive a general framework SpCo, which is able to improve the performance of existing GCL methods. Last, we choose three typical GCL methods as target methods, and plug SpCo into them. We validate the effectiveness of SpCo on five datasets. We consistently gain improvements compared with those target methods.

% \begin{itemize}
%     \item 
%     We answer ``what information is learned by GCL and whether there exists general augmentation rule".
%     To our best knowledge, this is the first attempt to fundamentally explore the augmentation strategies for GCL from spectral domain. We propose the contrastive invariance theorem of GCL, explain why GCL works, and discover the general augmentation rule, providing deeper understanding on the nature of GCL.
    
%     %and we prove this can encourage low-frequency information to be captured. And we formally define them as "optimal contrastive pair".
%     \item 
%     We answer ``how to utilize the augmentation rule for GCL". We show that the augmentation rule provides a novel insight to estimate the current augmentation strategies. Meanwhile, we propose a novel concept ``optimal contrastive pair", and theoretically derive a general framework SpCo, which is able to improve the performance of existing GCL methods. 
    
%     %We learn a new augmented structure from original structure by solving a matrix scaling problem with Sinkhorn's Iteration, and view this new structure and original structure to construct optimal contrastive pair.
%     \item We choose three typical GCL methods as target methods, and plug the SpCo into them. We validate the effectiveness of additional SpCo on five datasets, where we consistently gain improvements in all cases compared with those target methods.
% \end{itemize}
% !TEX root =  ../main.tex

\section{Preliminaries}
\label{pre}

Let $\mathcal{G=(\mathcal{V}, \xi)}$ represent an undirected attributed graph, where $\mathcal{V}$ is the set of $N$ nodes and $\xi\subseteq \mathcal{V}\times\mathcal{V}$ is the set of edges. All edges formulate an adjacency matrix $\bm{A}\in\{0,1\}^{N\times N}$, where $\bm{A}_{ij}\in\{0,1\}$ denotes the relation between nodes $i$ and $j$ in $\mathcal{V}$. The node degree matrix $\bm{D}=diag(d_1,\dots.d_n)$, where $d_i=\sum_{j\in \mathcal{V}}\bm{A}_{ij}$ is the degree of node $i \in \mathcal{V}$. Graph $\mathcal{G}$ is often associated with a node feature matrix $\bm{X}=[x_1, x_2,\dots, x_N]\in\mathbb{R}^{N\times d}$, where $x_i$ is a $d$ dimensional feature vector of node $i \in \mathcal{V}$. Let $\mathcal{\bm{L}}=\bm{D}-\bm{A}$ be the unnormalized graph Laplacian of $\mathcal{G}$. If we set symmetric normalized adjacency matrix as $\hat{\bm{A}}=\bm{D}^{-\frac{1}{2}}\bm{A}\bm{D}^{-\frac{1}{2}}$, then $\hat{\mathcal{\bm{L}}}=\bm{I_n}-\hat{\bm{A}}=\bm{D}^{-\frac{1}{2}}(\bm{D}-\bm{A})\bm{D}^{-\frac{1}{2}}$ is the symmetric normalized graph Laplacian.

%\subsection{Graph Spectrum} 

%From Graph Signal Processing (GSP), \textbf{graph spectrum} reflects amplitudes of different frequency components of a given graph in the spectral domain, indicating which parts of frequency are enhanced or weaken~\citep{gsp}.
Since $\hat{\mathcal{\bm{L}}}$ is symmetric normalized, its eigen-decomposition is $\bm{U}\bm{\Lambda}\bm{U}^\top$, where $\bm{\Lambda}=diag(\lambda_1,\dots,\lambda_N)$ and $\bm{U}=[\bm{u_1^\top},\dots,\bm{u_N^\top]}\in\mathbb{R}^{N\times N}$ are the eigenvalues and eigenvectors of $\hat{\mathcal{\bm{L}}}$, respectively. Without loss of generality, assume $0\leq\lambda_1\leq\cdots\leq \lambda_N<2$ (where we approximate $\lambda_N\approx 2$ \citep{gcn}). Denote by $\mathcal{F_L}=\{\lambda_1,\ldots,\lambda_{\lfloor N/2 \rfloor}\}$ the amplitudes of \textit{low-frequency components} and by $\mathcal{F_H}=\{\lambda_{\lfloor N/2 \rfloor+1},\ldots,\lambda_N\}$ the amplitudes of \textit{high-frequency components}. The \textbf{graph spectrum} is defined as these amplitudes of different frequency components, denoted as $\phi(\lambda)$, indicating which parts of frequency are enhanced or weakened~\citep{gsp}. Additionally, we rewrite $\hat{\bm{\mathcal{L}}}=\lambda_1\cdot \bm{u_1u_1}^\top+\dots+\lambda_N\cdot \bm{u_Nu_N}^\top$, where we define term $\bm{u_iu_i}^\top\in\mathbb{R}^{N\times N}$ as the eigenspace related to $\lambda_i$, denoted as $\bm{S_i}$.

% \subsection{Graph Convolutional Networks}
% As one typical architecture, GCN~\citep{gcn} performs exceedingly on the graph related tasks, and therefore it is usually selected as the backbone encoder in many classical methods, for example graph contrastive learning~\citep{dgi, mvgrl}. From the view of spectral domain, GCN elaborates first-order approximation of Fourier transformation on graphs based on Chebyshev polynomial expansion~\citep{chebnet}. Formally, the $k$th GCN layer can be written as:
% \begin{equation}
%     \label{gcn}
%     \textbf{Feature\ Transformation}: \bm{T^{(k)}} = \bm{H^{(k-1)}}\bm{W}^k,\ \textbf{Aggregation}: \bm{H^{(k)}}=\sigma(\hat{\bm{A}}\cdot\bm{T^{(k)}}),
% \end{equation}
% where $\bm{W}^k$ is the $k$th learnable parameters matrix, and $\sigma$ is non-linear activation function. $\bm{H^{(k-1)}}$ and $\bm{H^{(k)}}$ are the node embeddings obtained from $k-1$th and $k$th layers, and $\bm{H^{(0)}}=\bm{X}$.

\textbf{Graph Contrastive Learning} (GCL) \citep{dgi, mvgrl, cca} aims to learn discriminative embeddings without supervision, whose pipeline is shown in Fig.~\ref{gcl_frame}(a). We summarize the representative GCL in Appendix~\ref{related work}. Specifically, two augmentations are randomly extracted from $\bm{A}$ in a predefined way and are encoded by GCN~\citep{gcn} to obtain the node embeddings under these two augmentations. Then, for one target node, its embedding in one augmentation is learned to be close to the embeddings of its positive samples in the other augmentation and be far away from those of its negative samples. Models built in this way are capable of discriminating similar nodes from dissimilar ones. For example, some graph contrastive methods~\citep{gca, graphcl, grace} use classical InfoNCE loss~\citep{cpc} as the optimization objective:
\begin{equation}
    \label{contra}
    \mathcal{L}(\bm{h_i^{V_1}},\bm{ h_i^{V_2}})=\log\frac{\exp(\theta\bm{(h_i^{V_1}}, \bm{h_i^{V_2}})/\tau)}{\exp(\theta(\bm{h_i^{V_1}}, \bm{h_i^{V_2}})/\tau)+\sum\limits_{k\neq i}\exp(\theta(\bm{h_i^{V_1}}, \bm{h_k^{V_2}})/\tau)},
\end{equation}
where $\bm{h_i^{V_1}}$ and $\bm{h_i^{V_2}}$ are the embeddings of node $i$ under augmentations $V_1$ and $V_2$, respectively, $\theta$ is the similarity metric, such as cosine similarity, and $\tau$ is a temperature parameter. The total loss is $\mathcal{L}_{InfoNCE}=\sum\limits_i \frac{1}{2}\left(\mathcal{L}(\bm{h_i^{V_1}}, \bm{h_i^{V_2}})+\mathcal{L}(\bm{h_i^{V_2}}, \bm{h_i^{V_1}})\right)$.
% !TEX root =  ../main.tex
\setcounter{footnote}{0}
\section{Impact of Graph Augmentation: An Experimental Investigation}
\label{case study}

\begin{wrapfigure}[9]{r}{0.4\textwidth}
    \centering
    \vspace{-15pt}
    \includegraphics[width=0.38\textwidth,height=0.2\textwidth]{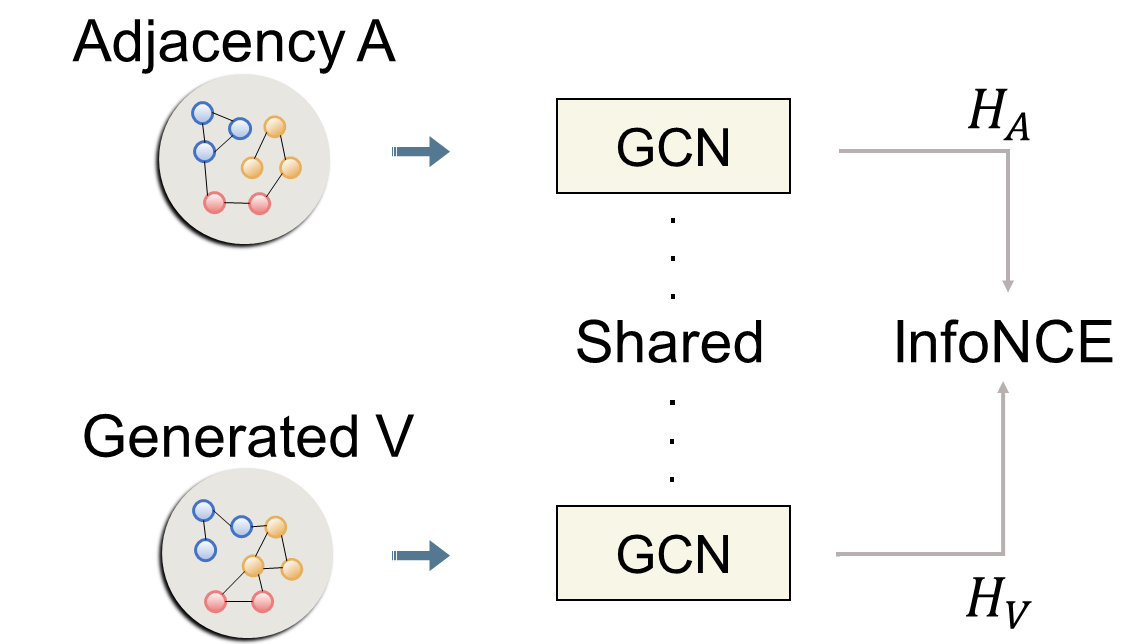}
    \caption{The case study model.}\label{case_study_model}
\end{wrapfigure}
In this section, we aim to explore what information should be considered in two contrasted augmentations from the perspective of graph spectrum. Specifically, we design a simple GCL framework shown in Fig.~\ref{case_study_model}. Two input augmentations are adjacency matrix $\bm{A}$ and generated $\bm{V}$. Then, we utilize a shared GCN with one layer as the encoder to encode $\bm{A}$ and $\bm{V}$ and get their nodes embeddings as $\bm{H_A}$ and $\bm{H_V}$.We train the GCN by utilizing InfoNCE loss as in Eq.~\eqref{contra}. More experimental settings can be found in appendix~\ref{Experimental settingss}.

% \begin{figure}[htbp]
% \centering
% \begin{minipage}[h]{0.4\linewidth}
% \centering
% \includegraphics[scale=0.15, height=3cm]{fig/case_model.png}
% \caption{The case study model.}\label{case_study_model}
% \end{minipage}
% \hspace{30pt}
% \begin{minipage}[h]{0.4\linewidth}
% \centering
% \includegraphics[scale=0.18]{fig/genev.png}
% \caption{The generation of $\bm{V}$.}\label{genev}
% \end{minipage}
% \end{figure}

\begin{wrapfigure}[9]{r}{0.4\textwidth}
    \centering
    \includegraphics[width=0.4\textwidth]{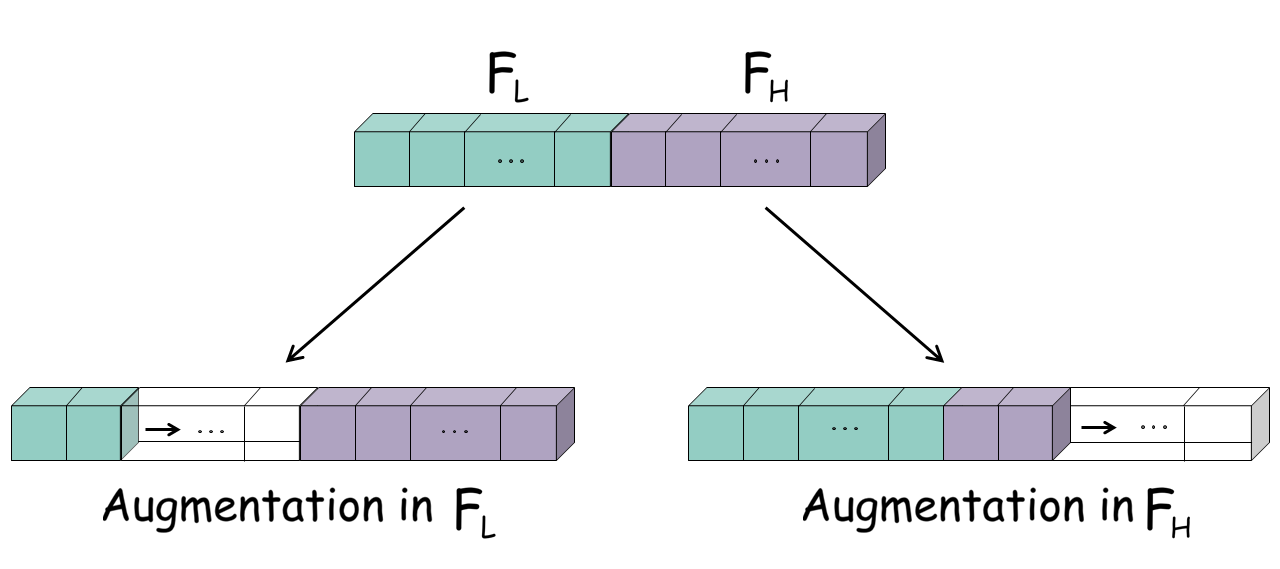}
    \caption{The generation of $\bm{V}$.}\label{genev}
\end{wrapfigure}
\textbf{Generating augmentation $\bm{V}$} \label{gene} We construct the augmented graph by extracting information with different frequencies from the original graph, so that we can analyze the effect of different information. This process is shown in Fig.~\ref{genev}. Specifically, we divide the eigenvalues of $\bm{\mathcal{L}}$ into $\mathcal{F_L}$ and $\mathcal{F_H}$ parts, and conduct augmentations in these two parts, respectively. Taking the augmentation in $\mathcal{F_L}$ for example, we keep the high-frequency part as $\bm{u_{N/2}u_{N/2}}^\top+\dots+\bm{u_Nu_N}^\top$. Then, we gradually add the eigenspaces in $\mathcal{F_L}$ back with rates $[20\%, 40\%, 60\%, 80\%]$, starting from the lowest frequency. Therefore, $\bm{V}$ augmenting 20\% in $\mathcal{F_L}$ is $\bm{u_1u_1}^\top+\dots+\bm{u_{0.2*N/2}u_{0.2*N/2}}^\top+\bm{u_{N/2}u_{N/2}}^\top+\dots+\bm{u_Nu_N}^\top$. Similarly, $\bm{V}$ augmenting 20\% in $\mathcal{F_H}$ is $\bm{u_1u_1}^\top+\dots+\bm{u_{N/2}u_{N/2}}^\top+\bm{u_{(N+1)/2}u_{(N+1)/2}}^\top+\dots+\bm{u_{0.7N}u_{0.7N}}^\top$. Please note that we set graph spectrum of $\bm{V}$, $\phi_{\bm{V}}(\lambda)=1, \forall \lambda\in[0,2]$ above, in that we just want to test the effect of different $\bm{u_iu_i}^\top$ and avoid the influence from eigenvalues $\lambda$ \citep{Revisiting}.

\begin{figure}[ht]
\centering
\subfigure[Cora]
{
    \begin{minipage}[b]{.23\linewidth}
        \centering
        \includegraphics[scale=0.2]{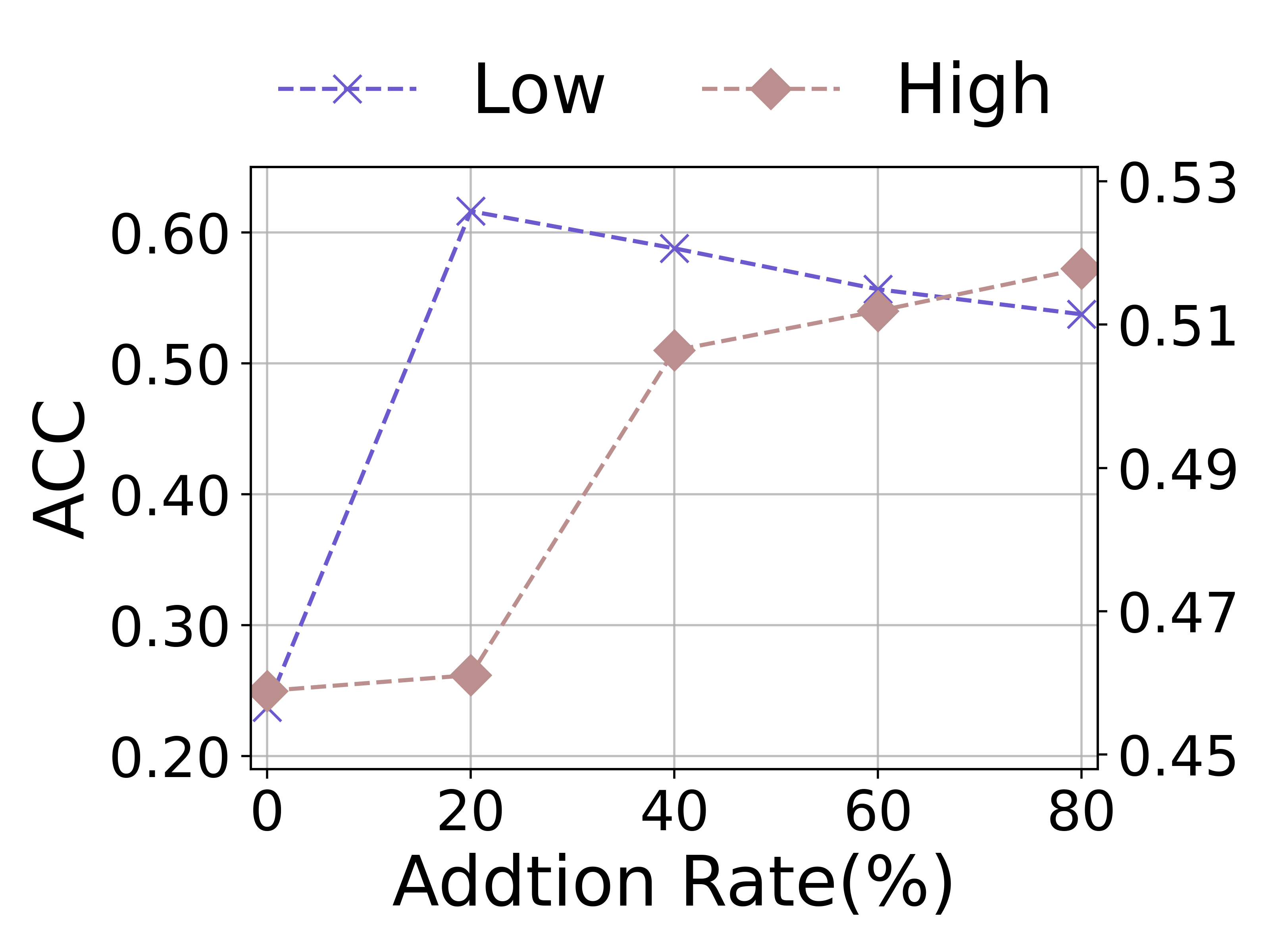}
    \end{minipage}
}
\subfigure[Citeseer]
{
 	\begin{minipage}[b]{.23\linewidth}
        \centering
        \includegraphics[scale=0.2]{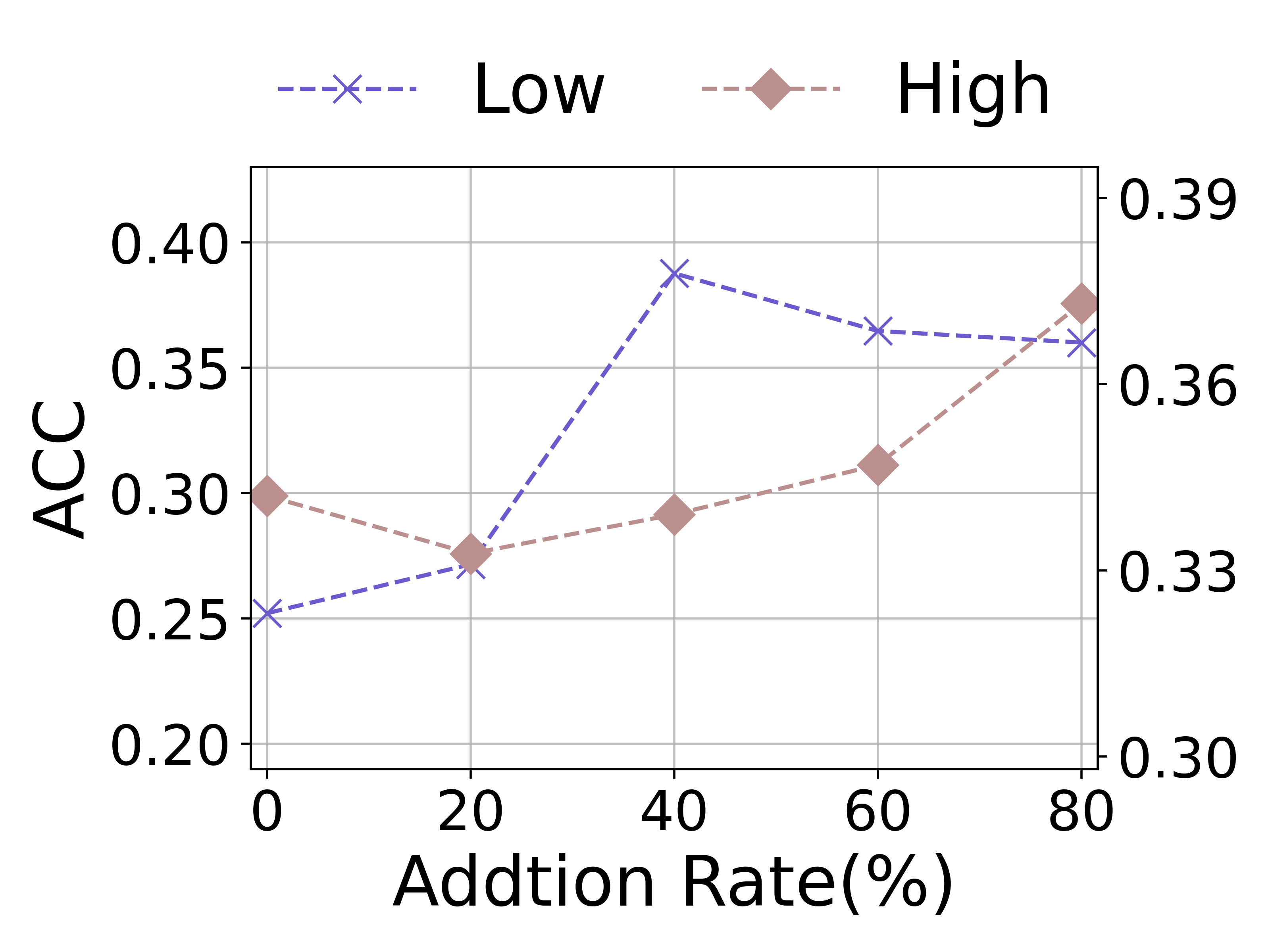}
    \end{minipage}
}
\subfigure[BlogCatalog]
{
 	\begin{minipage}[b]{.23\linewidth}
        \centering
        \includegraphics[scale=0.2]{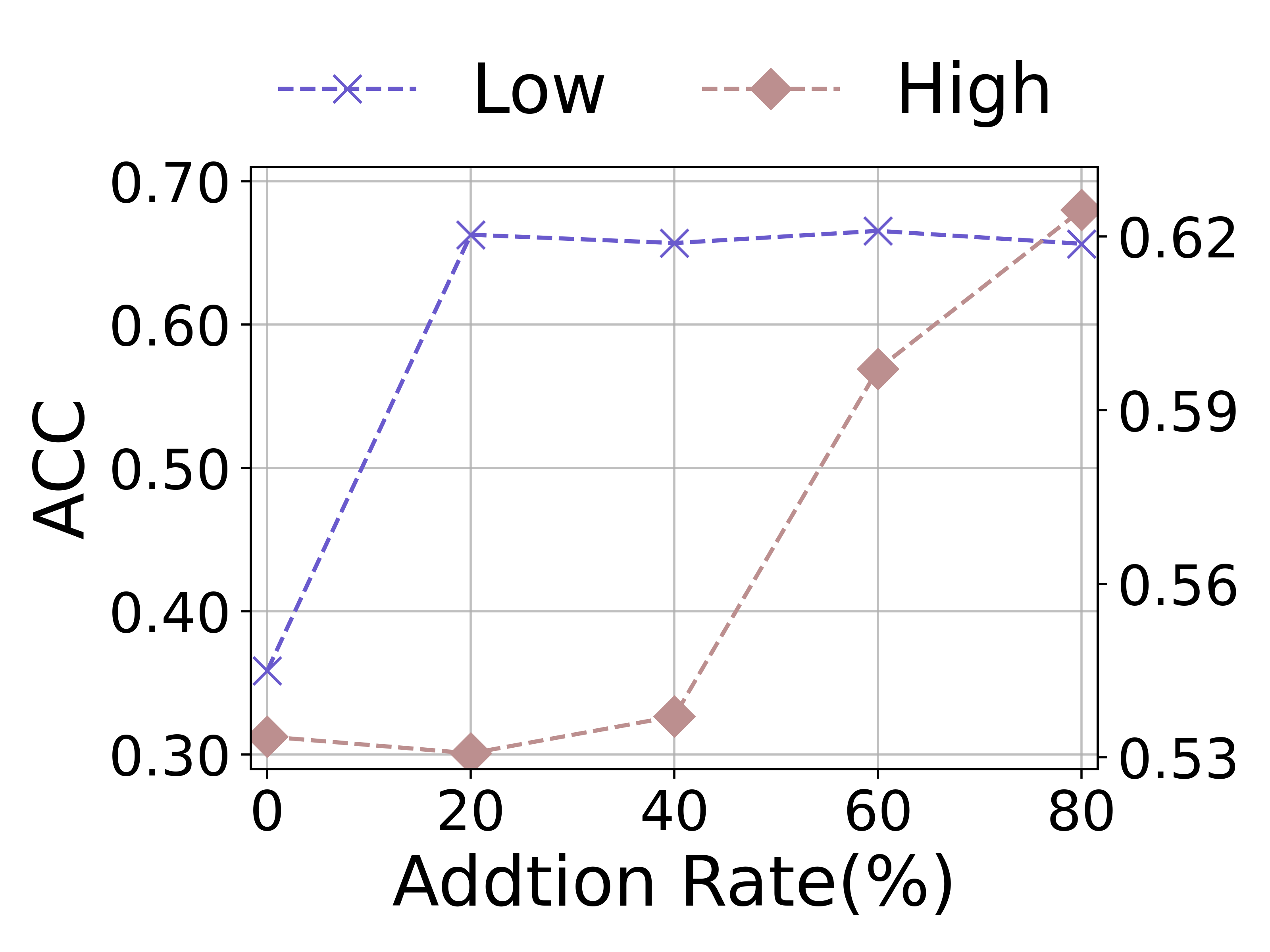}
    \end{minipage}
}
\subfigure[Flickr]
{
 	\begin{minipage}[b]{.23\linewidth}
        \centering
        \includegraphics[scale=0.2]{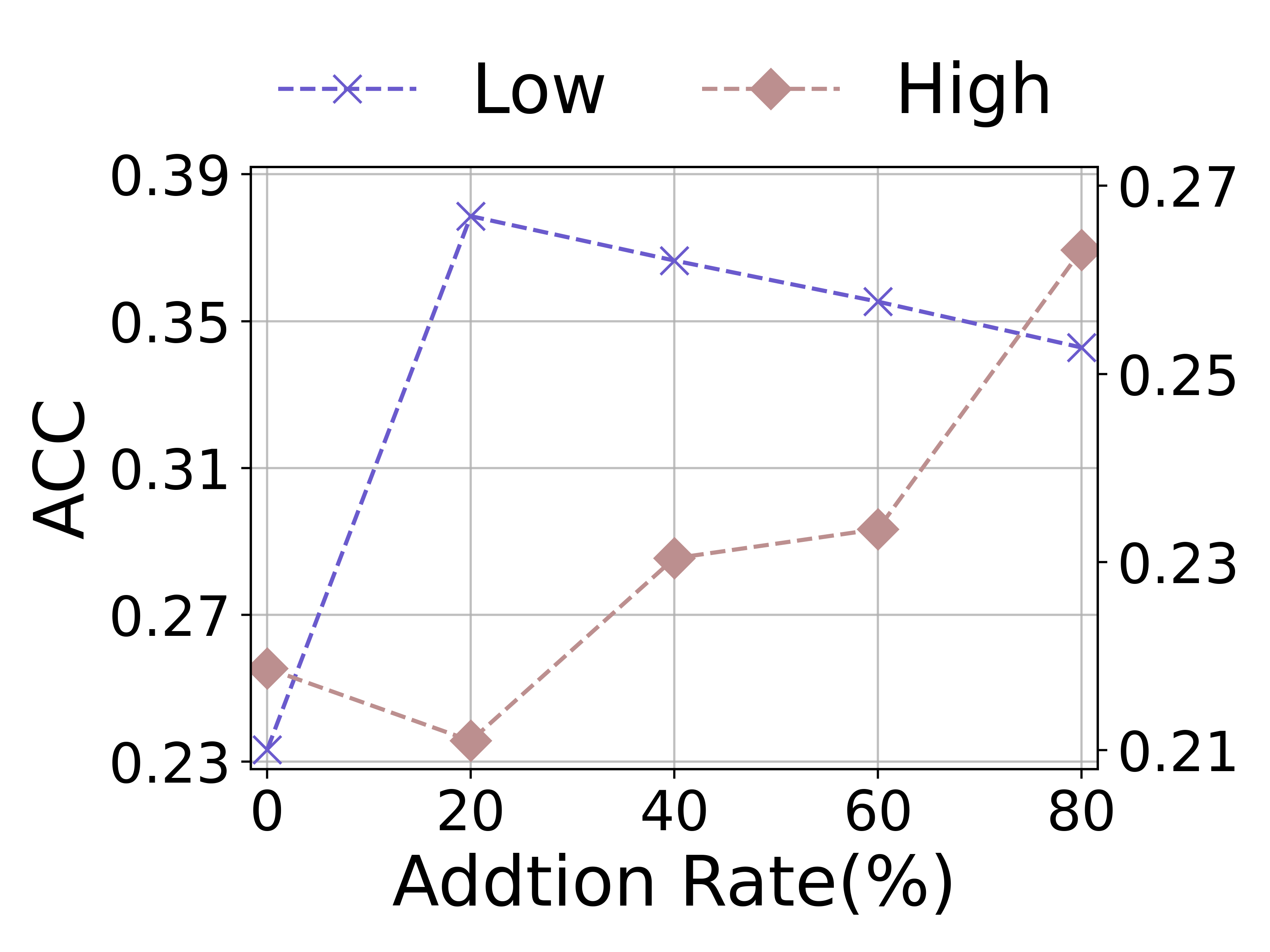}
    \end{minipage}
}
\caption{The results of case study on four datasets. The x-axis means different addition rate of different frequency interval, and y-axis means the performance on ACC. The performance of augmentations in $\mathcal{F_L}$ are plotted on the left y-axis, and in $\mathcal{F_H}$ are plotted on the right y-axis.}
\label{case_results}
\end{figure}

\begin{wrapfigure}[12]{r}{0.25\textwidth}
    \centering
    \includegraphics[width=0.25\textwidth]{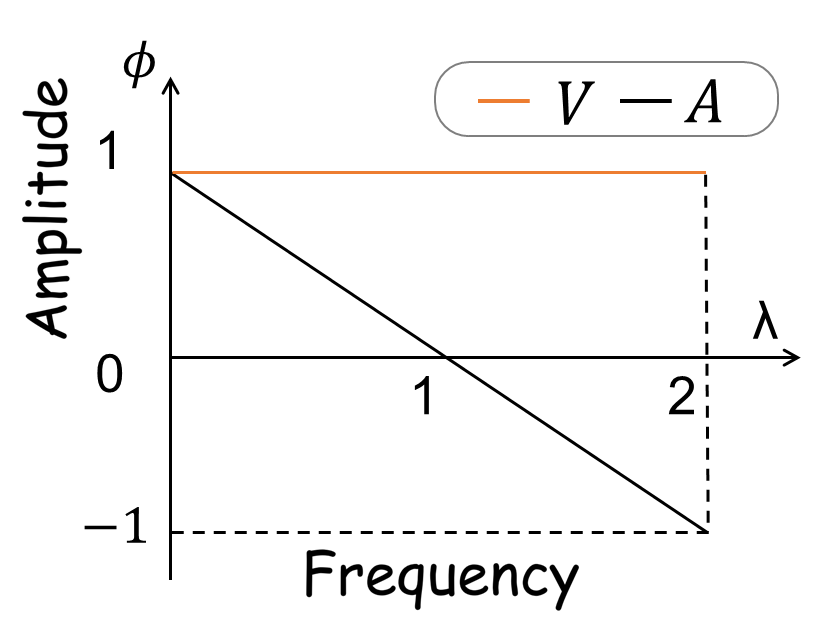}
    \caption{The spectrum of $\bm{A}$ and $\bm{V}$.}\label{spectrum}
\end{wrapfigure}
\textbf{Results and analyses} We conduct the node classification on four datasets: Cora, Citeseer \citep{gcn}, BlogCatalog, and Flickr \citep{meng2019co}. The accuracy (ACC) is shown in Fig.~\ref{case_results}. In appendix \ref{More results of case study}, we also report the results when both high and low frequency components are added back in the high-to-low frequency order.
\textbf{Results.} For each dataset, in generated $\bm{V}$, (1) when the lowest part of frequencies are kept, the best performance is achieved; (2) when more frequencies in $\mathcal{F_H}$ are involved, the performance generally rises. \textbf{Analyses.} From the graph spectra of $\bm{A}$ and $\bm{V}$ shown in Fig.~\ref{spectrum}, we can see that in generated $\bm{V}$, (1) when the lowest part of frequencies are kept, the difference of amplitude, i.e., the graph spectrum, in $\mathcal{F_L}$ between $\bm{A}$ and $\bm{V}$ becomes smaller; (2) when more frequencies in $\mathcal{F_H}$ are involved, the margin of graph spectrum in $\mathcal{F_H}$ between $\bm{A}$ and $\bm{V}$ becomes larger. Combining results and observations, we propose the following general \textbf{G}raph \textbf{A}ug\textbf{ME}ntation rule, called GAME rule\footnote{Although this rule is derived from contrasting $\bm{A}$ and $\bm{V}$, the selection of certain views does not curb the generality of GAME rule. Considering that most of augmentations are obtained from the raw adjacency matrix $\bm{A}$, it is a natural setting that one view is fixed as $\bm{A}$ and the other is an augmented one.}:
\begin{center}
\begin{tcolorbox}[
                  colback=yellow!10,
                  colframe=black!55,
                  width=\textwidth,
                  arc=1mm, auto outer arc,
                  boxrule=0.5pt,
                  title=The General Graph Augmentation Rule ,
                 ]
Given two random augmentations $\bm{V_1}$ and $\bm{V_2}$, their graph spectra are $\phi_{V_1}(\lambda)$ and $\phi_{V_2}(\lambda)$. Then,  $\forall$ $\lambda_m$ $\in$ [1,2] and $\lambda_n$ $\in$ [0,1], $\bm{V_1}$ and $\bm{V_2}$ are an effective pair of graph augmentations if the following condition is satisfied: \\
\centerline{\bm{$\left|\phi_{V_1}(\lambda_m)-\phi_{V_2}(\lambda_m)\right| > \left|\phi_{V_1}(\lambda_n)-\phi_{V_2}(\lambda_n)\right|$}.}\\
We define such pair of augmentations as optimal contrastive pair.
\end{tcolorbox}
\end{center}

\setcounter{footnote}{0}
\section{Analysis of The General Graph Augmentation Rule}
\label{The General Graph Augmentation Ru}

In this section, we aim to verify the correctness of GAME rule that whether two contrasted augmentations satisfying GAME rule can perform better in downstream tasks from experimental and theoretical analysis.

\begin{figure}[ht]
  \centering
  \includegraphics[width=0.8\textwidth]{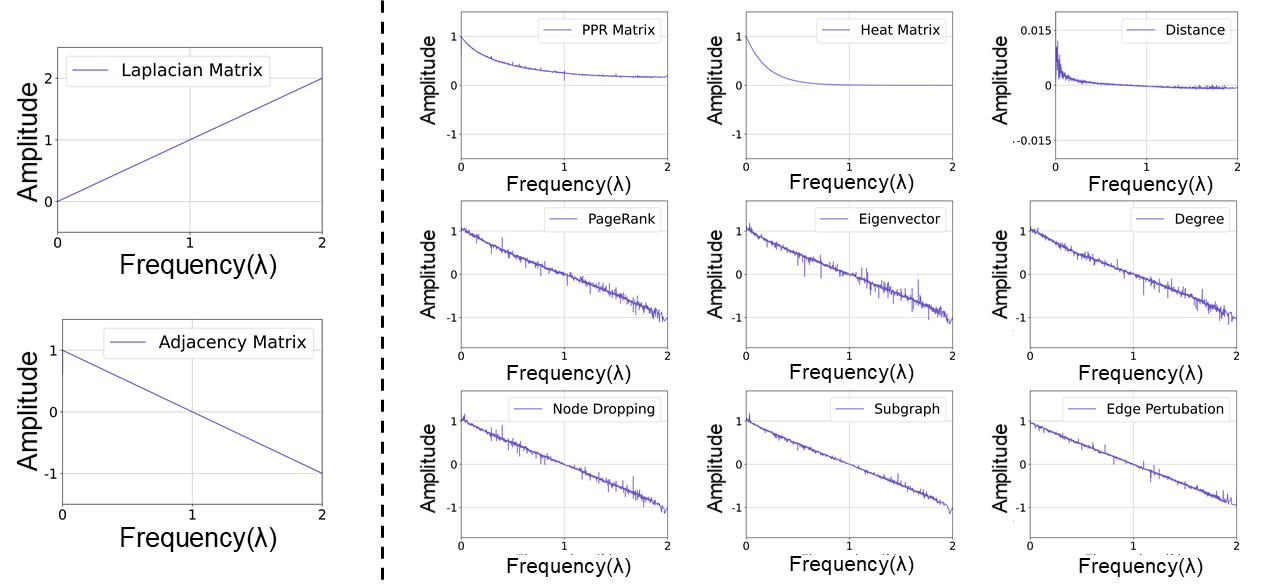}
  \caption{The graph spectra of laplacian, adjacency matrix and nine existing augmentations.} 
  \label{view_exist} 
\end{figure}

\textbf{Experimental analysis} We substitute existing augmentations proposed by MVGRL \citep{mvgrl}, GCA \citep{gca} and GraphCL \cite{graphcl} for augmentation $\bm{V}$ in the case. Specifically, MVGRL proposes PPR matrix, heat diffusion matrix and pair-wise distance matrix. GCA mainly randomly drops edges based on Degree, Eigenvector and PageRank. GraphCL adopts random node dropping, edge perturbation and subgraph sampling. The nine augmentations almost cover the mainstream augmentations in GCL. To accurately depict the change of the amplitude after these augmentations for some $\lambda_i$, we turn to matrix perturbation theory \footnote{Here, we do not use eigenvalue decomposition to obtain $\lambda'$ of $\bm{{A}'}$, because the obtained $\lambda'$ are unordered compared with previous $\lambda$ of $\bm{A}$. That is to say, for certain $\lambda_i$ of $\bm{A}$, we cannot figure out which eigenvalue of $\bm{{A}'}$ matches to it after decomposition, so we cannot calculate the change $\Delta \lambda_i$ for $\lambda_i$ in this case.} \citep{mtxper}:
\begin{equation}
    \label{pertu}
    \Delta \lambda_i=\lambda_i'-\lambda_i=\bm{u}_i^\top\Delta\bm{A}\bm{u}_i-\lambda_i\bm{u}_i^\top\Delta\bm{D}\bm{u}_i+\mathcal{O}(||\Delta\bm{A}||),
\end{equation}
where $\lambda_i'$ is the eigenvalue after change, $\Delta \bm{{A}}=\bm{{A}'}-\bm{{A}}$ represent the modification of edges after augmentation, and $\Delta \bm{D}$ is the respective change in degree matrix. With Eq.~\eqref{pertu}, we calculate the eigenvalues on Cora after each augmentation, and plot their graph spectra in Fig.~\ref{view_exist}. Simultaneously, we use the GCL framework in Section~\ref{case study} to separately contrast adjacency matrix $\bm{A}$ and these augmentations, and results are shown in Table~\ref{view_table}. As shown in Fig.~\ref{view_exist}, PPR matrix, Heat diffusion matrix and Distance matrix better accord with GAME rule, where they have small difference with $\bm{A}$ in $\mathcal{F_L}$, and have a large difference in $\mathcal{F_H}$. Therefore, they outperform other augmentations in Table~\ref{view_table}. 
%Then, all of augmentations proposed by GCA and GraphCL have the roughly similar spectra with that of $\bm{A}$, while the spectra of augmentations by GCA have a fiercer oscillation in $\mathcal{F_H}$ than that of augmentations by GraphCL, and thus they have a slightly larger difference with $\bm{A}$ in $\mathcal{F_H}$. Therefore, the performances of three augmentations by GCA are better than that by GraphCL.

\begin{table}[h]
  \centering
  \caption{Performance of different existing augmentations to verify the GAME rule.}
  \label{view_table}
  \resizebox{\textwidth}{!}{
  \begin{tabular}{c|c|c|c|c|c|c|c|c|c}
  \bottomrule
     Methods   & \multicolumn{3}{c|}{GraphCL} &  \multicolumn{3}{c|}{GCA} &  \multicolumn{3}{c}{MVGRL} \\
     \hline
     Type & Subgraph & Node dropping & Edge perturbation & Degree & PageRank & Eigenvector & PPR & Heat & Distance \\
     \hline
     Results & 34.9$\pm3.5$ & 29.8$\pm2.3$ & 37.7$\pm4.4$ & 40.2$\pm4.1$ & 38.5$\pm5.0$ & 42.1$\pm4.9$ & \textbf{58.0$\pm1.6$} &\textbf{49.9$\pm4.2$} & \textbf{46.1$\pm7.5$} \\
     \hline
  \end{tabular}}
\end{table}
We also test the GAME rule in another circumstance, where we contrast among three cases: $\bm{A}$ and $\bm{A}^2$ (two-hop of $\bm{A}$), $\bm{A}$ and $\bm{A}$, and $\bm{A}^2$ and $\bm{A}^2$. The results are given in Appendix~\ref{another}.

\textbf{Theoretical analysis} 
% In this part, we aim to answer that \textit{given two augmentations meeting GAME rule, how they manipulate the learning process of GCL?} Combined with the case and proper simplification and assumption,
We have the following theorem to depict the learning process of the GCL.
\begin{Theorem}
\label{case_theo}
\textbf{(Contrastive Invariance)} Given adjacency matrix $\bm{A}$ and the generated augmentation $\bm{V}$, the amplitudes of $i$-th frequency of $\bm{A}$ and $\bm{V}$ are $\lambda_i$ and $\gamma_i$, respectively. With the optimization of InfoNCE loss $\mathcal{L}_{InfoNCE}$, the following upper bound is established: \\
\centerline{$\mathcal{L}_{InfoNCE} \leq \frac{1+N}{2}\sum\limits_i \theta_i\left[2-\left(\lambda_i-
\gamma_i\right)^2\right],$} \\
where $\theta_i$ is an adaptive weight of the $i$th term.
\end{Theorem}

The proof is given in the Appendix~\ref{proof_1}, where we simplify GCN without the activation function. Theorem~\ref{case_theo} indicates an upper bound of GCL loss, implying that maximizing the contrastive loss equals to maximize the upper bound. So, larger $\theta_i$ will be assigned to the smaller $\left(\lambda_i-\gamma_i\right)^2$, or $\lambda_i\approx\gamma_i$. Meanwhile, if $\lambda_i\approx\gamma_i$, these two contrasted augmentations are regarded to share the invariance at $i$th frequency. Therefore, with contrastive learning, the encoder will emphasize the invariance between two contrasted augmentations from spectrum domain. To our best knowledge, theorem~\ref{case_theo}, for the first time, theoretically proves that GCL can capture the invariance between two augmentations. Please recall that GAME rule suggests that the difference between two augmentations in $\mathcal{F_L}$ is smaller. Thus, under the guidance of GAME rule, GCL attempts to capture the common low-frequency information of two augmentations. Thus, GAME rule points out a general augmentation strategy to manipulate encoder to capture low-frequency information, which achieves a better performance. 
\setcounter{footnote}{0}
\section{Spectral Graph Contrastive Learning}
\label{mmmmmmmm}
\begin{wrapfigure}[12]{r}{0.5\textwidth}
    \centering
    \vspace{-15pt}
    \includegraphics[scale=0.19]{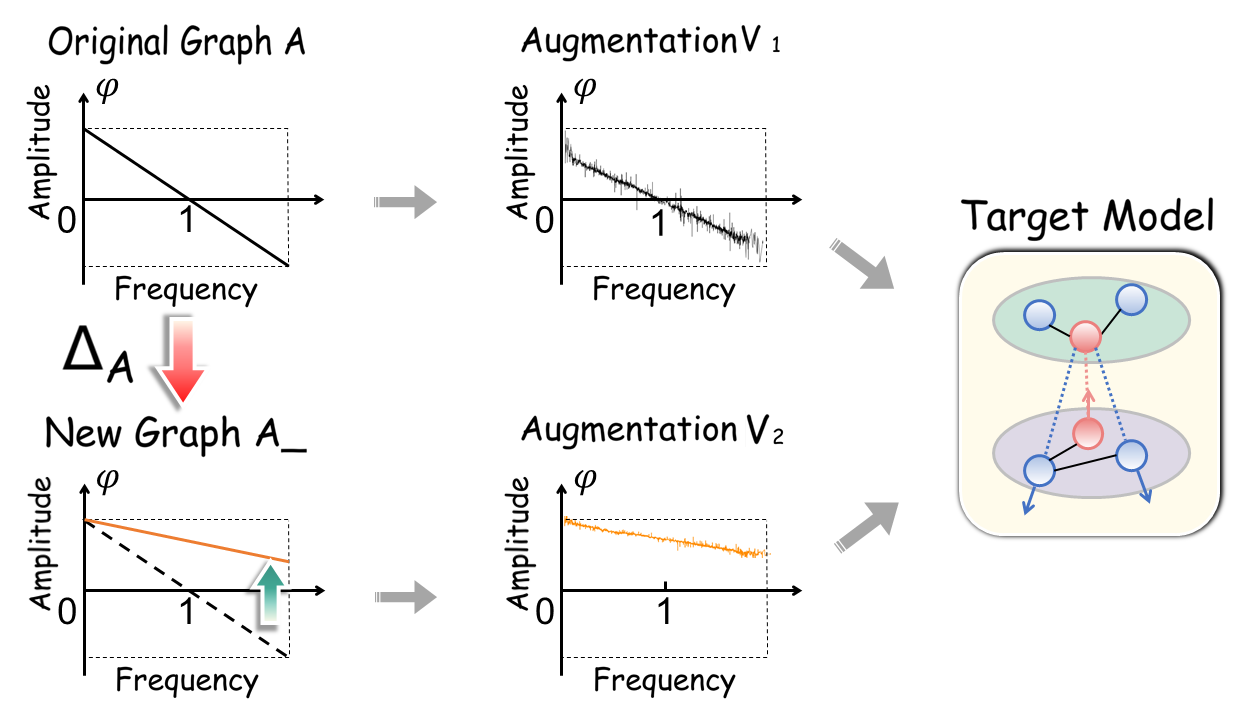}
    \caption{Combine SpCo with existing GCL.}\label{spco}
\end{wrapfigure}
Based on the GAME Rule, we mainly aim to learn a general and GCL-friendly transformation $\Delta_{\bm{A}}$ from adjacency matrix $\bm{A}$ to a new augmentation $\bm{A\_}$ (or $\Delta_{\bm{A}}=\bm{A\_}-\bm{A}$), where $\bm{A}$ and $\bm{A\_}$ are required to be an optimal contrastive pair. Then, they are fed into existing GCL method $\Phi$, i.e. augmenting with the same strategies of $\Phi$ to generate $\bm{V_1}$ and $\bm{V_2}$ and training with the corresponding contrastive loss, shown in Fig.~\ref{spco}. The whole pipeline is our proposed spectral graph contrastive learning (SpCo), which can boost existing GCL methods.

% In this section, we attempt to propose a general and GCL-friendly plug-in based on the GAME Rule, called spectral graph contrastive learning (SpCo), which can boost existing GCL methods. As shown in Fig.~\ref{spco}, we mainly aim to learn a transformation $\Delta_{\bm{A}}$ from adjacency matrix $\bm{A}$ to a new augmentation $\bm{A\_}$ (or $\Delta_{\bm{A}}=\bm{A\_}-\bm{A}$), where $\bm{A}$ and $\bm{A\_}$ are required to be an optimal contrastive pair. Then, they are fed into combined GCL method $\Phi$, i.e. augmenting with the same strategies of $\Phi$ to generate $\bm{V_1}$ and $\bm{V_2}$ and training with the corresponding contrastive loss.

Firstly, we separate $\Delta\bm{A}=\Delta_{\bm{A+}}-\Delta_{\bm{A-}}$, where $\Delta_{\bm{A+}}$ and $\Delta_{\bm{A-}}$ indicate which edge is added and deleted, respectively. Next, we indicate how to learn $\Delta_{\bm{A+}}$, while the calculation of $\Delta_{\bm{A-}}$ is similar. Based on our theoretical derivation in Appendix.~\ref{Derivation of optimization objective}, the following optimization objective of $\Delta_{\bm{A+}}$ should be maximized:
\begin{equation}
    \label{target}
    \mathcal{J} = \underbrace{<\bm{\mathcal{C}}, \ \Delta_{\bm{A+}}>^2}_{\text{\small Matching Term}}+  \underbrace{\epsilon H(\Delta_{\bm{A+}})}_{\text{\small Entropy Reg.}}+\underbrace{<\bm{f},\Delta_{\bm{A+}}\mathbbm{1}_n-\bm{a}>+<\bm{g},\Delta_{\bm{A+}}^\top\mathbbm{1}_n-\bm{b}>}_{\text{\small Lagrange Constraint Conditions}},
\end{equation}
This objective consists of three components: (1) $\textbf{Matching Term}$. $\forall \ \bm{P},\bm{Q}\in\mathbb{R}^{N\times N}$, $<\bm{P},\bm{Q}>=\sum_{ij}\bm{P}_{ij}\bm{Q}_{ij}$. To maximize $<\bm{\mathcal{C}}, \ \Delta_{\bm{A+}}>^2$, $\Delta_{\bm{A+}}$ should learn to "match" or be similar to $\bm{\mathcal{C}}$. In~\ref{Derivation of optimization objective}, we define $\bm{\mathcal{C}}=\bm{U}g(\lambda)\bm{U}^\top$, where $\bm{U}$ and $g(\lambda)$ are eigenvector matrix and some function about eigenvalues of $\bm{A}$. According to GAME rule, we set $\phi_\Delta(\lambda)=|\phi_{\bm{A}}(\lambda)-\phi_{\bm{A\_}}(\lambda)|$, and we need $\phi_\Delta(\lambda_m)>\phi_\Delta(\lambda_n)$, $\forall$ $\lambda_m$ $\in$ [1,2] and $\lambda_n$ $\in$ [0,1]. Therefore, we stipulate that $\phi_\Delta(\lambda)$ should be a monotone increasing function. Since $\bm{\mathcal{C}}$ will guide $\Delta_{\bm{A+}}$ to capture the change of difference between graph spectra ($\phi_\Delta(\lambda)$), we naturally set $g(\lambda)$ of $\bm{\mathcal{C}}$ also a monotone increasing function. Furthermore, we notice that the graph spectrum of Laplacian $\mathcal{\bm{L}}$ does meet our need about $g(\lambda)$ (shown in Fig.~\ref{view_exist}), so we simply set $\bm{\mathcal{C}}=\Theta\bm{\mathcal{L}}$, where $\Theta$ is a parameter updating in training. (2) $\textbf{Entropy Regularization}$. Here, $H(\bm{P})=-\sum_{i, j}\bm{P}_{i, j}(\log(\bm{P}_{i, j})-1)$ \citep{ot}, and $\epsilon$ is the weight of this term. This term aims to increase the uncertainty of the learnt $\Delta_{\bm{A+}}$, which encourages more edges (entries in $\Delta_{\bm{A+}}$) to join in optimization. (3) $\textbf{Lagrange Constraint Conditions}$. $\bm{f}\in \mathbb{R}^{N\times 1}$ and $\bm{g}\in \mathbb{R}^{N\times 1}$ are Lagrange multipliers, and $\bm{a}\in \mathbb{R}^{N\times 1}$ and $\bm{b}\in \mathbb{R}^{N\times 1}$ are distributions\footnote{We define $\bm{a}$ and $\bm{b}$ are both node degree distribution in this paper.}. This term restrains the row and column sums of $\Delta_{\bm{A+}}$ within some limitation.

Next, we expound how to solve eq.~\eqref{target}. The partial of $\mathcal{J}$ with respect to $\Delta_{\bm{A+}}$ is as following:
\begin{subequations}
\begin{align} 
\partial \mathcal{J}\ /\ \partial (\Delta_{\bm{A+}})_{ij}&=2<\bm{\mathcal{C}}, \ \Delta_{\bm{A+}}>\bm{\mathcal{C}}_{ij}-\epsilon \log (\Delta_{\bm{A+}})_{ij} + \bm{f}_i+\bm{g}_j  \label{aaa}\\
&=m_{ij}+2\bm{\mathcal{C}}_{ij}^2(\Delta_{\bm{A+}})_{ij}-\epsilon \log (\Delta_{\bm{A+}})_{ij}+\bm{f}_i+\bm{g}_j, \label{bbb}
\end{align}
\end{subequations}
where we separate $2\bm{\mathcal{C}}_{ij}^2(\Delta_{\bm{A+}})_{ij}$ from $2<\bm{\mathcal{C}}, \ \Delta_{\bm{A+}}>\bm{\mathcal{C}}_{ij}$, and set the rest part as $m_{ij}$. The next theorem points out when $\mathcal{J}$ can get the maximal value in the domain of definition $(\Delta_{\bm{A+}})_{ij}\in(0, 1)$:
\begin{Theorem}
\label{theorem 4}
Given\ $(\Delta_{\textbf{A+}})_{ij}\in(0,1)$,\ $\mathcal{J}$\ exists\ the\ maximal\ value,\ iff\\
$\quad\text{(1)}$ $\bm{\mathcal{C}}_{ij}^2 < -\frac{\bm{f}_i+\bm{g}_j+m_{ij}}{2}$,\text{ and}\ $\bm{f}_i+\bm{g}_j+m_{ij}<0$,\ or\\
$\quad\text{(2)}$ $\frac{\epsilon}{2}<\bm{\mathcal{C}}_{ij}^2<\frac{\epsilon}{2}\exp(-\frac{\bm{f}_i+\bm{g}_j+m_{ij}+\epsilon}{2})$,\text{ and}\ $\bm{f}_i+\bm{g}_j+m_{ij}+\epsilon<0$.
\end{Theorem}
We provide the proof in the Appendix~\ref{proof_4}. Normally, we should let eq.~\eqref{bbb} equal to zero and get the analytical solution of $(\Delta_{\bm{A+}})_{ij}$. However, eq.~\eqref{bbb} is a transcendental equation because of the coexistence of linear term and logarithm. 
Thus, we require eq.~\eqref{aaa} to equal to 0. As the training goes on, $\Delta_{\bm{A+}}$ does not change sharply. So, we firstly rewrite eq.~\eqref{aaa} as follows:
\begin{equation}
\label{rewrite}
    \partial \mathcal{J}\ /\ \partial (\Delta_{\bm{A+}})_{ij}\approx2<\bm{\mathcal{C}}, \ \Delta_{\bm{A+}}'>\bm{\mathcal{C}}_{ij}-\epsilon \log (\Delta_{\bm{A+}})_{ij} + \bm{f}_i+\bm{g}_j. 
\end{equation}
Compared with eq.~\eqref{aaa}, eq.~\eqref{rewrite} only replaces $\Delta_{\bm{A+}}$ with $\Delta_{\bm{A+}}'$ in the first term, where $\Delta_{\bm{A+}}'$ is obtained from the last training epoch, and frozen at the current epoch. In this case, the matrix form of solution of current epoch is:
\begin{equation}
\label{solution}
    \Delta_{\bm{A+}} = diag(\bm{u})\exp\left(2<\bm{\mathcal{C}}, \Delta_{\bm{A+}}'>\bm{\mathcal{C}}\ /\ \epsilon\right)diag(\bm{v})=\bm{U_{+}K_{+}V_{+}},
\end{equation}
where $\bm{U_{+}}=diag(\bm{u}_i)=diag\big(\exp\big(\frac{\bm{f}_i}{\epsilon}\big)\big)$ and $\bm{V_{+}}=diag(\bm{v}_j)=diag\big(\exp\big(\frac{\bm{g}_j}{\epsilon}\big)\big)$. To further calculate $\bm{U_{+}}$ and $\bm{V_{+}}$, we restrain the row and column sums of $\Delta_{\bm{A+}}$ according to Lagrange Constraint Conditions: $\bm{u}*\left(\bm{K_{+}v}\right)=\bm{a}$ and $ \bm{v}*\left(\bm{K_{+}}^\top \bm{u}\right)=\bm{b}$. We solve this matrix scaling problem \citep{nemirovski1999complexity} by Sinkhorn's Iteration \citep{sinkhorn}, which is shown in Algorithm~\ref{Sinkhorn's Iteration Process} \citep{cuturi2013sinkhorn}. There exists a upper bound of the difference between $\Delta_{\bm{A+}}$ and $\Delta_{\bm{A+}}'$:
\begin{Theorem}
\label{theorem 5}
After Sinkhorn's Iteration, the bound between $\Delta_{\bm{A+}}$ and $\Delta_{\bm{A+}}'$ is: \\
\centerline{$\left|\alpha(\Delta_{\bm{A+}})_{ij}-(\Delta_{\bm{A+}})_{ij}'\right|\leq\frac{\alpha}{\epsilon^2(1-\gamma)}\{d(r^{(0)},\bm{a})+d(c^{(0)},\bm{b})\}+\alpha(1+\frac{|m_{ij}|}{\epsilon})$,} \\
where $\alpha=\frac{\epsilon}{2\bm{\mathcal{C}}_{ij}^2}$. $\forall (x, x')\in(\mathbb{R}^n_{+})^2$, $d(x, x')$ is the Hilbert’s projective metric \citep{bushell1973hilbert} on $\mathbb{R}^n_{+}$. $\gamma$ is $\kappa({\bm{K_{+}}})$, and $\kappa$ is contraction ratio \citep{franklin1989scaling}. $r^{(0)}$ and $c^{(0)}$ are the row and column sum vectors of $\bm{K_{+}}$.
\end{Theorem}
The proof is given in Appendix~\ref{proof_5}. The calculation of $\Delta_{\bm{A-}}$ is similar as $\Delta_{\bm{A+}}$ shown as follows:
\begin{equation}
\label{solution_}
    \Delta_{\bm{A-}} = diag(\bm{u}')\exp\left(-2<\bm{\mathcal{C}}, \Delta_{\bm{A-}}'>\bm{\mathcal{C}}\ /\ \epsilon\right)diag(\bm{v}')=\bm{U_{-}K_{-}V_{-}},
\end{equation}
where $diag(\bm{u}')$, $diag(\bm{v}')$ and $\Delta_{\bm{A-}}'$ have the similar meanings as in eq.~\eqref{solution}.

Finally, we get the solution $\Delta_{\bm{A}}=\Delta_{\bm{A+}}-\Delta_{\bm{A-}}$, utilizing eq.~\eqref{solution} and eq.~\eqref{solution_}. With learnt transformation $\Delta_{\bm{A}}$, we can obtain the new augmentation $\bm{A\_}$ as:
\begin{equation}
\begin{aligned}
\label{ete}
    \bm{A\_} = \bm{A}+\eta\cdot\mathbb{S}*\Delta_{\bm{A}},
\end{aligned}
\end{equation}
where '*' means element-wise product, and $\eta$ is the combination coefficient. To make $\Delta_{\bm{A}}$ sparse, we use scope matrix $\mathbb{S}$ to limit our focus, e.g. one-hop neighbors for each node. The whole algorithm is given in Algorithm~\ref{The proposed SpCo}.

\begin{minipage}{.43\textwidth}
\vspace{-7pt}
\begin{algorithm}[H]
\label{Sinkhorn's Iteration Process}
\caption{Sinkhorn's Iteration}
\LinesNumbered
\SetKwInOut{Input}{\textbf{Input}}
\SetKwInOut{Params}{\textbf{Params}}
\SetKwInOut{Output}{\textbf{Output}}

\Input{Matrix $\bm{K}$, distribution $\bm{a}\in \mathbb{R}^{N\times 1}$ and $\bm{b}\in \mathbb{R}^{N\times 1}$}
\Params{Iteration number $Iter$}
\Output{$\Delta_{\bm{A+}}$ (or $\Delta_{\bm{A-}}$)}
\BlankLine
Initialize $\bm{u}=[1/N, 1/N, \dots, 1/N]_{1\times N}$; \\
$\overline{\bm{K}}=diags(1./\bm{a})\bm{K}$;\\
\For{$i=1$ to $Iter$}{
$\bm{u}=1./\overline{\bm{K}}\left(\bm{b}/\bm{K}^\top\bm{u} \right)$;
}
$\bm{v}=\bm{b}/\bm{K}^\top\bm{u}$;\\
$\Delta_{\bm{A+}} / \Delta_{\bm{A-}} = diag(\bm{u})\bm{K}diag(\bm{v})$;\\
\Return $\Delta_{\bm{A+}} / \Delta_{\bm{A-}}$\;
\end{algorithm}
\end{minipage}
\hspace{16pt}
\begin{minipage}{.52\textwidth}
\begin{algorithm}[H]
\label{The proposed SpCo}
\caption{The proposed SpCo}
\LinesNumbered
\SetKwInOut{Input}{\textbf{Input}}
\SetKwInOut{Params}{\textbf{Params}}
\SetKwInOut{Output}{\textbf{Output}}
\Input{$\Phi$, augmentation $\bm{Aug}_{\Phi}$, $\bm{A}$, $\mathcal{\bm{L}}$, $\bm{X}$}
\Params{Total epochs $T$, update epochs $\Omega$, $\mathbb{S}$, $\eta$, $\Theta$, $\epsilon$, $\bm{a}$ and $\bm{b}$}
\BlankLine
\For{$i=1$ to $T$}{
$\bm{\mathcal{C}}=\Theta\mathcal{\bm{L}}$;\\
Calculate $\bm{K}_{+}$ / $\bm{K}_{-}$ in eq.~\eqref{solution}, \eqref{solution_};\\
Get $\Delta_{\bm{A+}}$ / $\Delta_{\bm{A-}}$ through Algorithm~\ref{Sinkhorn's Iteration Process};\\
$\bm{A\_}=\bm{A}+\eta(\Delta_{\bm{A+}}-\Delta_{\bm{A-}})\mathbb{S}$ with eq.~\eqref{ete};\\
Update $\Theta$;\\
\For{$j=1$ to $\Omega$}{
$\bm{V_1}, \bm{V_2}$ = $\bm{Aug}_{\Phi}(\bm{A}), \bm{Aug}_{\Phi}(\bm{A\_})$;\\
Train $\Phi(\bm{V_1}, \bm{V_2}, \bm{X})$ \;
}
}
\end{algorithm}
\end{minipage}

\section{Experiments}
\label{experiments}
In this section, we mainly evaluate the performance of proposed SpCo on five datasets: Cora, Citeseer, Pubmed \citep{gcn}, BlogCatalog and Flickr \citep{meng2019co}. Details of datasets are in Appendix~\ref{app_dataset}. We select two categories of baselines: semi-supervised GNN models \{GCN \citep{gcn}, GAT \citep{gat}\} and six representative graph contrastive learning methods \{DGI \citep{dgi}, MVGRL \citep{mvgrl}, GRACE \citep{grace}, GCA \citep{gca}, GraphCL \citep{graphcl}, CCA-SSG \citep{cca}\}. These GCL methods can be divided into three categories based on their contrastive losses: BCE loss (DGI, MVGRL), InfoNCE loss (GRACE, GCA, GraphCL) and CCA loss (CCA-SSG). To verify the applicability of our SpCo, we select one baseline from each category (DGI, GRACE and CCA-SSG) to integrate with SpCo. The detailed descriptions of DGI, GRACE and CCA-SSG are given in Appendix~\ref{app_describ}. Experimental implementation details are given in Appendix~\ref{Implementation Details}.

\begin{table*}[h]
  \Huge
  \caption{Quantitative results (\%$\pm\sigma$) on node classification.}
  \label{fenlei}
  \resizebox{\textwidth}{!}{
  \renewcommand\arraystretch{1.5}
  \begin{tabular}{c|c|c|c|cc|c|cc|c|c|cc}
    \bottomrule
    Datasets & Metrics & GCN & GAT & DGI & \textbf{DGI+SpCo} & MVGRL & GRACE & \textbf{GRACE+SpCo} & GCA & GraphCL & CCA-SSG& \textbf{CCA+SpCo}\\
    \bottomrule
    \multirow{2}{*}{Cora}&
    {Ma-F1}&79.6±0.7&81.3±0.3&80.4±0.7&\textbf{81.1±0.5}&81.5±0.5&79.2±1.0&\textbf{80.3±0.8}&79.9±1.1&80.7±0.9&82.9±0.8&\textbf{83.6±0.4}\\
    \cline{2-13}
    &{Mi-F1}&80.7±0.6&82.3±0.2&82.0±0.5&\textbf{82.8±0.7}&82.8±0.4&80.0±1.0&\textbf{81.2±0.9}&81.1±1.0&82.3±0.9&83.6±0.9&\textbf{84.3±0.4}\\
    \hline
    \multirow{2}{*}{Citeseer}&
    {Ma-F1}&68.1±0.5&67.5±0.2&67.7±0.9&\textbf{68.3±0.5}&66.8±0.7&65.1±1.2&\textbf{65.1±0.8}&62.8±1.3&67.8±1.0&67.9±1.0&\textbf{68.5±1.0}\\
    \cline{2-13}
    &{Mi-F1}&70.9±0.5&72.0±0.9&71.7±0.8&\textbf{72.4±0.5}&72.5±0.5&68.7±1.1&\textbf{69.4±1.0}&65.9±1.0&71.9±0.9&73.1±0.7&\textbf{73.6±1.1}\\
    \hline
    \multirow{2}{*}{BlogCatalog}&
    {Ma-F1}&71.2±1.2&67.6±2.2&68.2±1.3&\textbf{71.5±0.8}&80.3±3.6&67.7±1.2&\textbf{68.2±0.4}&71.7±0.4&63.9±2.1&72.0±0.5&\textbf{72.8±0.3}\\
    \cline{2-13}
    &{Mi-F1}&72.1±1.3&68.3±2.2&68.8±1.4&\textbf{72.3±0.9}&80.9±3.6&68.5±1.3&\textbf{69.4±1.3}&72.7±0.5&64.6±2.1&73.0±0.5&\textbf{73.7±0.3}\\
    \hline
    \multirow{2}{*}{Flickr}&
    {Ma-F1}&48.9±1.6&35.0±0.8&31.2±1.6&\textbf{33.7±0.7}&31.2±2.9&35.7±1.3&\textbf{36.3±1.4}&41.2±0.5&32.1±1.1&37.0±1.1&\textbf{38.7±0.6}\\
    \cline{2-13}
    &{Mi-F1}&50.2±1.2&37.1±0.3&33.0±1.6&\textbf{35.2±0.7}&33.4±3.0&37.3±1.0&\textbf{38.1±1.3}&42.2±0.6&34.5±0.9&39.3±0.9&\textbf{40.4±0.4}\\
    \hline
    \multirow{2}{*}{PubMed}&
    {Ma-F1}&78.5±0.3&77.4±0.2&76.8±0.9&\textbf{77.6±0.6}&79.8±0.4&80.0±0.7&\textbf{80.3±0.3}&80.8±0.6&77.0±0.4&80.7±0.6&\textbf{81.3±0.3}\\
    \cline{2-13}
    &{Mi-F1}&78.9±0.3&77.8±0.2&76.7±0.9&\textbf{77.4±0.5}&79.7±0.3&79.9±0.7&\textbf{80.7±0.2}&81.4±0.6&76.8±0.5&81.0±0.6&\textbf{81.5±0.4}\\
    \hline
  \end{tabular}}
\end{table*}

\subsection{Node classification}
\label{Node classification}
To more comprehensively evaluate our model, we use two common evaluation metrics, including Macro-F1 and Micro-F1. The results are reported in Table~\ref{fenlei}, where the training set contains 20 nodes per class. As can be seen, the proposed SpCo can generally improve the performances of the corresponding original models on all datasets, which verifies that our SpCo is widely applicable and effective. We also choose 5 and 10 labeled nodes per class as training set respectively, which are reported in Appendix~\ref{node classification}.
\begin{figure}[htbp]
\centering
\subfigure[DGI: Citeseer]
{
    \begin{minipage}[b]{.3\linewidth}
        \centering
        \includegraphics[scale=0.23]{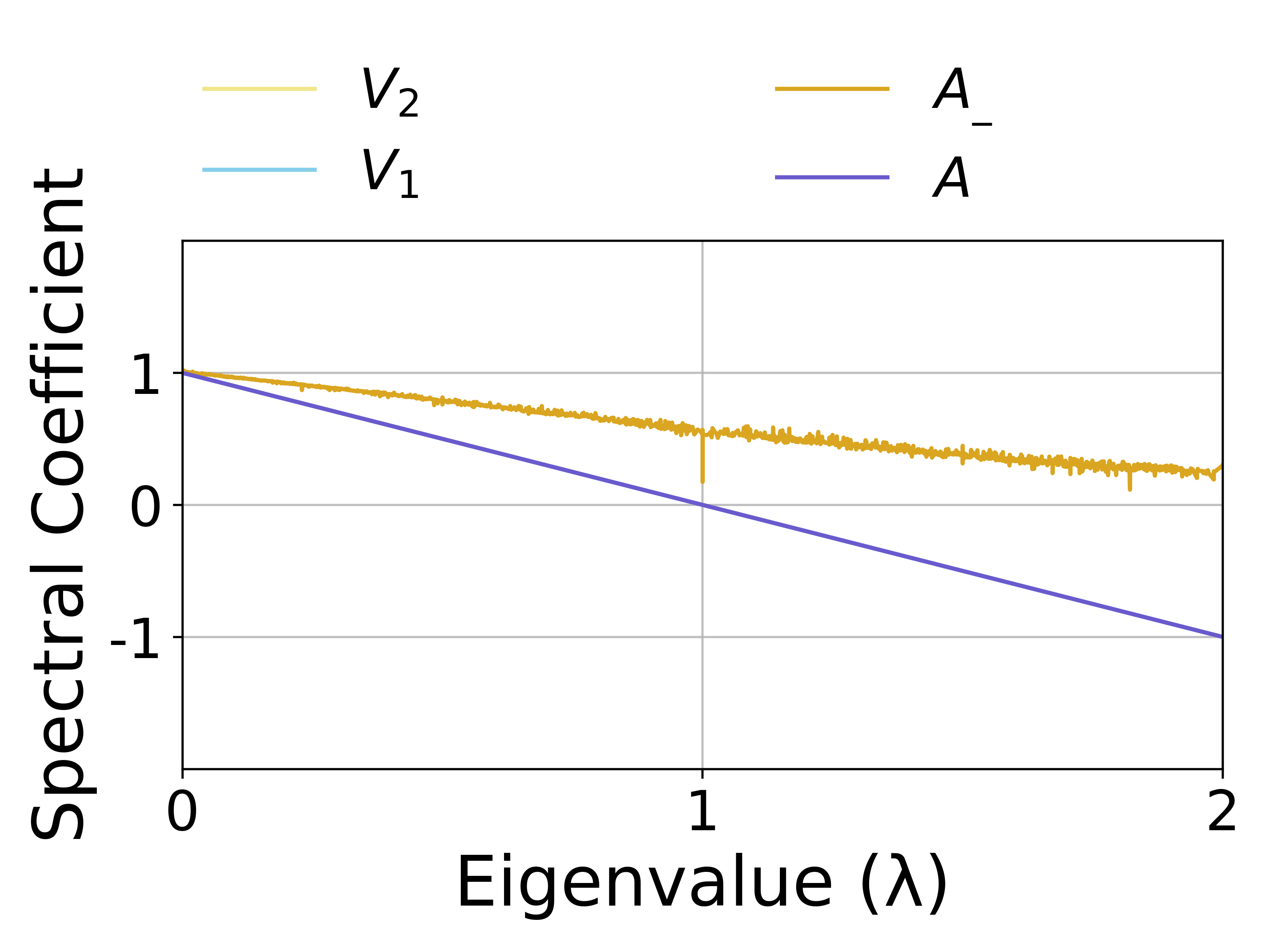}
    \end{minipage}
}
\subfigure[GRACE: Citeseer]
{
 	\begin{minipage}[b]{.3\linewidth}
        \centering
        \includegraphics[scale=0.23]{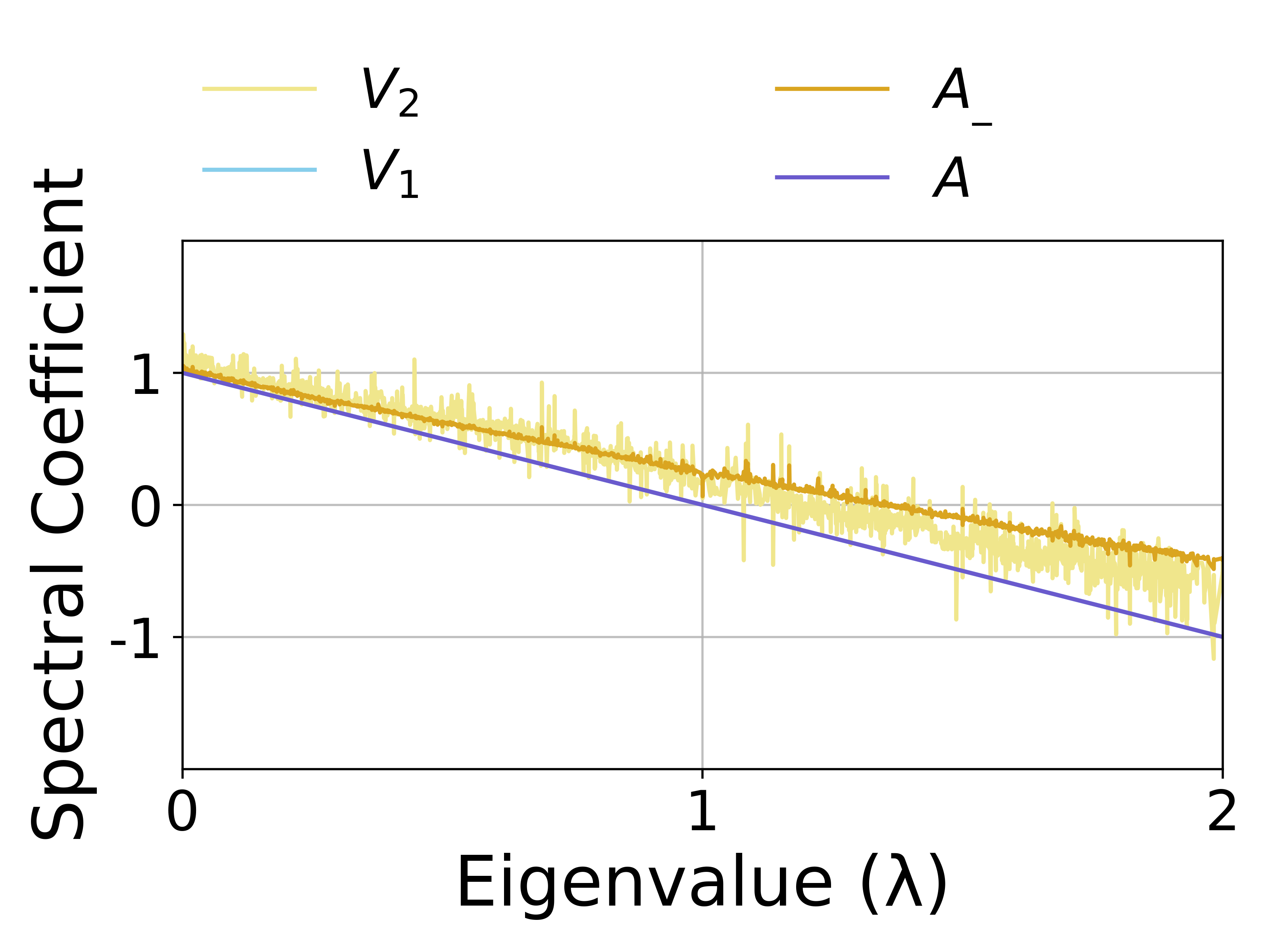}
    \end{minipage}
}
\subfigure[CCA-SSG: Citeseer]
{
 	\begin{minipage}[b]{.3\linewidth}
        \centering
        \includegraphics[scale=0.23]{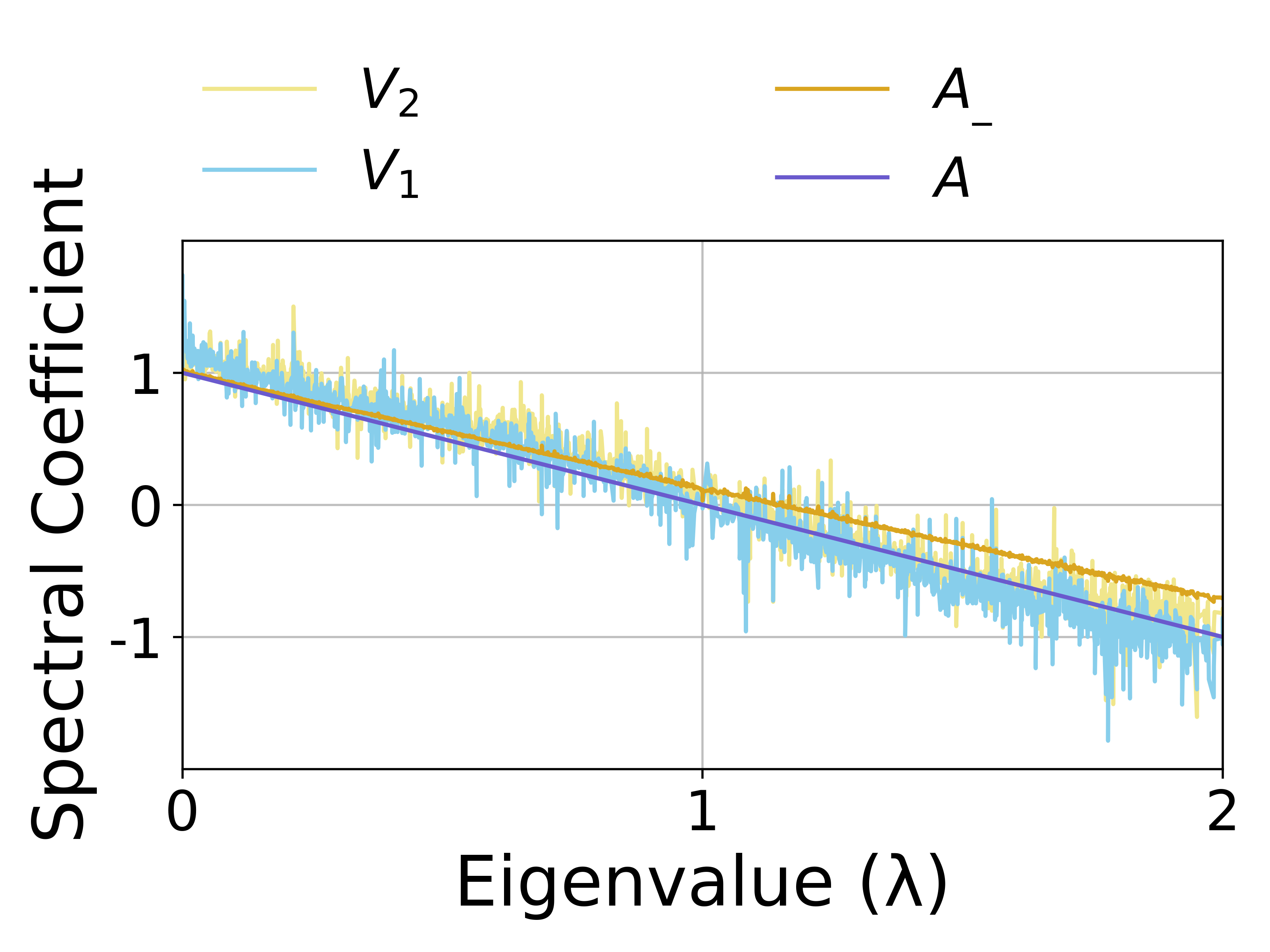}
    \end{minipage}
}
\caption{The visualisation of graph spectrum on Citeseer.}
\label{cora}
\end{figure}
\subsection{Visualisation of graph spectrum}
In this section, we test if the learnt view $\bm{A\_}$ and $\bm{A}$ meet the GAME rule. We plot the graph spectrum of $\bm{A\_}$, $\bm{A}$, $\bm{V_1}$ and $\bm{V_2}$ in one figure for each method on Citeseer, which are shown in Fig.~\ref{cora}. Here, we discard the impact of self-loop operation. For DGI, it does not use topological augmentation. \footnote{Although in DGI, the authors summary a vector to depict the global view, this summary vector does not reflect any graph structure, thus we think DGI does not have special augmentation strategies on topology.} Therefore, we only plot $\bm{A\_}$ and $\bm{A}$ for it. For GRACE, the augmentation strength of $\bm{V_1}$ is set to 0. Thus, the plot of $\bm{V_1}$ is same with $\bm{A}$. From the figures, we can see that the difference between $\bm{A\_}$ and $\bm{A}$ is smaller in $\mathcal{F_L}$ than in $\mathcal{F_H}$, which proves that they are optimal contrastive pair. Meanwhile, they can drive $\bm{V_1}$ and $\bm{V_2}$ also to obey the GAME rule, and thus boost the final results. More results on Cora are given in Appendix~\ref{isualisation of graph spectrum}.
\subsection{Hyper-parameter sensitivity}
In this subsection, we systematically investigate the sensitivity of two parameters: matrix $\bm{\mathcal{C}}$ and $\epsilon$. We conduct node classification on Cora and BlogCatalog datasets and report the Micro-F1 values. More experiments of hyper-parameters are given in Appendix~\ref{Hyper-parameter Sensitivity}.

\begin{wrapfigure}[10]{r}{0.6\textwidth}
    \centering
    \vspace{-15pt}
    \subfigure[Cora]
{
    \begin{minipage}[b]{.25\textwidth}
        \centering
        \includegraphics[width=\textwidth]{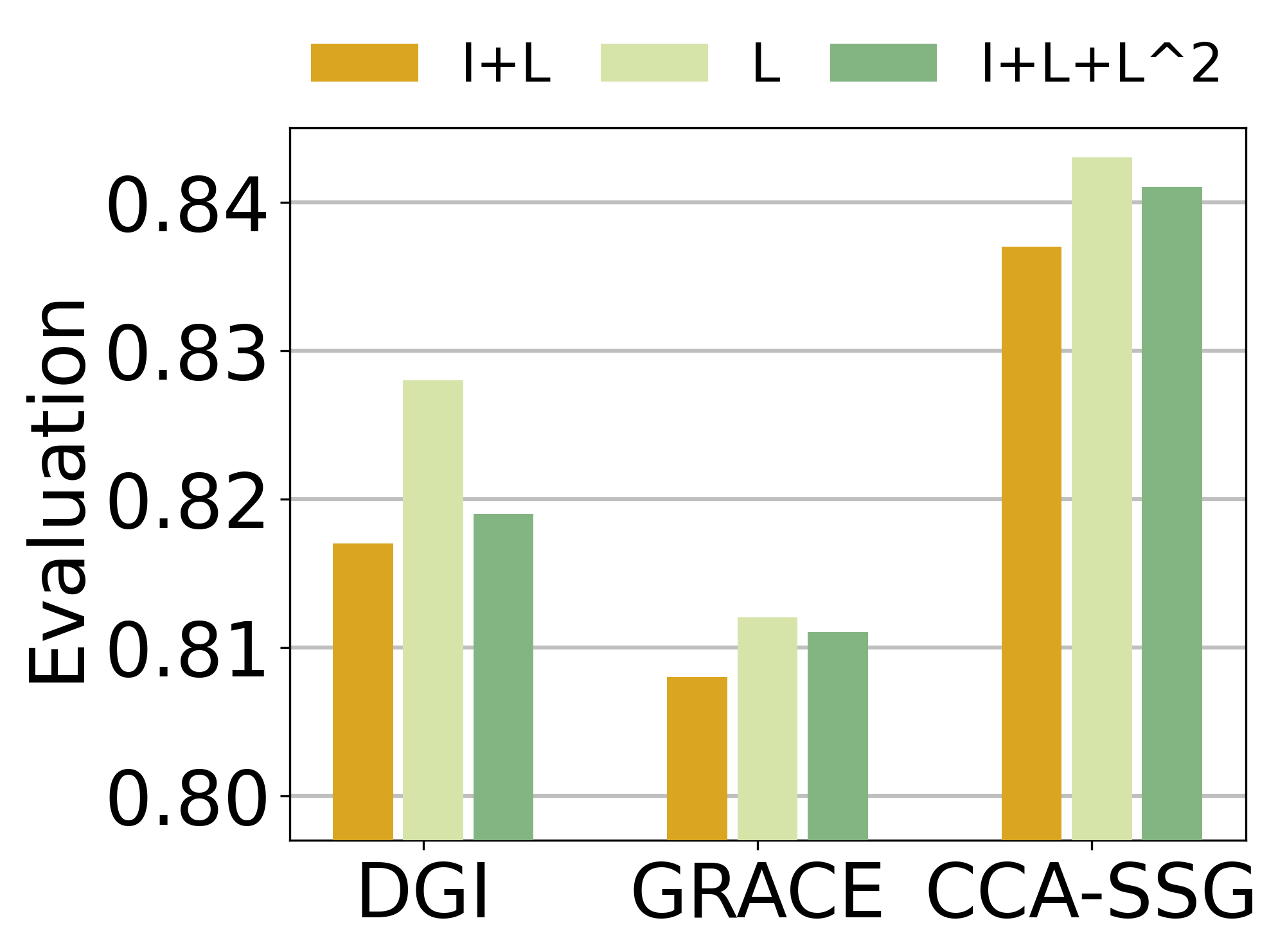}
    \end{minipage}
}
\subfigure[BlogCatalog]
{
 	\begin{minipage}[b]{.25\textwidth}
        \centering
        \includegraphics[width=\textwidth]{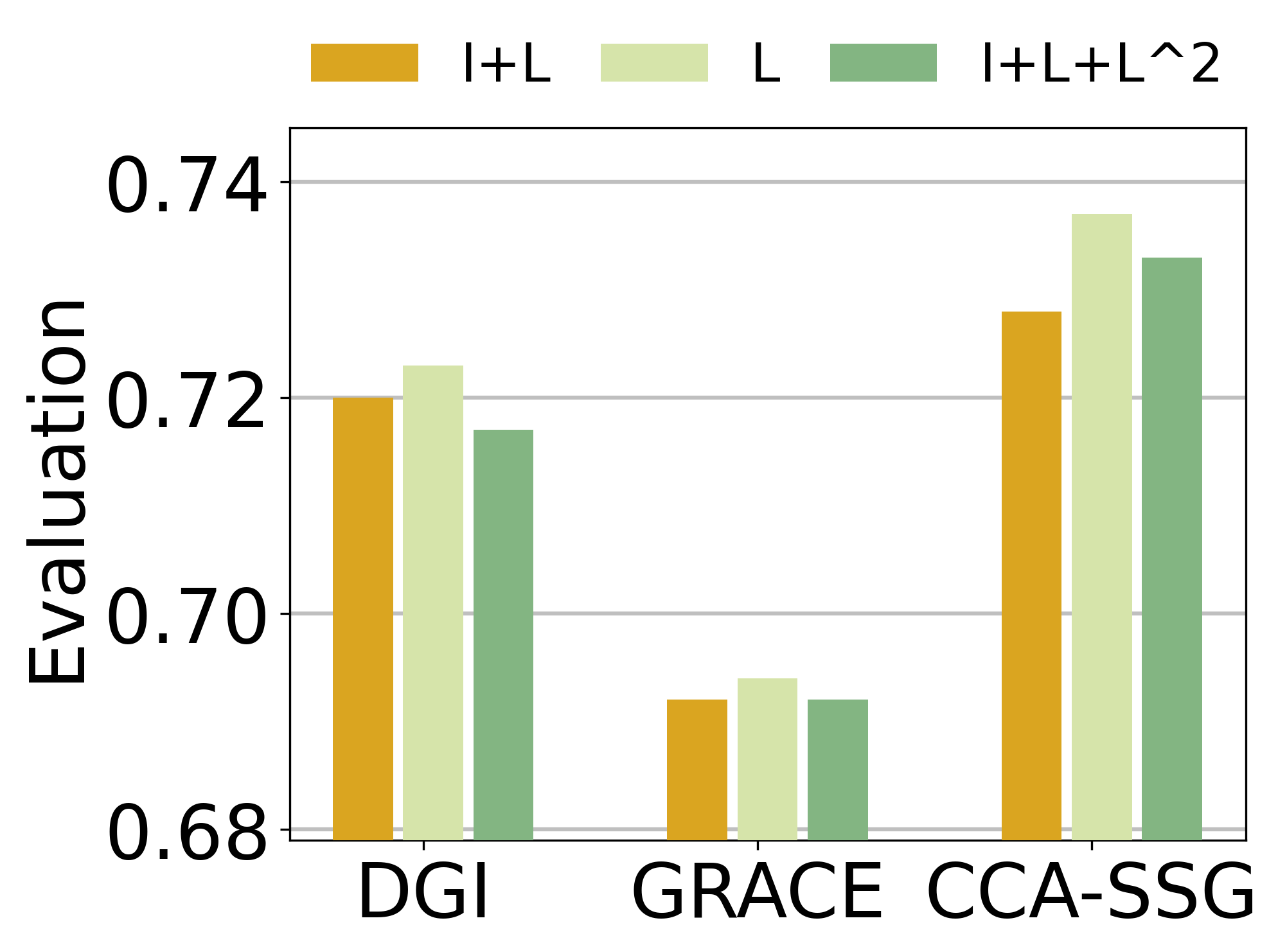}
    \end{minipage}
}
\vspace{-10pt}
\caption{The comparison between three candidates for $\bm{\mathcal{C}}$.}
\label{testc}
\end{wrapfigure} 
\textbf{Analysis of $\bm{\mathcal{C}}$.} The matrix $\bm{\mathcal{C}}$ directly affects the final structures of the $\Delta_{\bm{A+}}$ and $\Delta_{\bm{A-}}$. Therefore, we give three kinds of $\bm{\mathcal{C}}$: $\bm{\mathcal{I+L}}$, $\bm{\mathcal{L}}$ and $\bm{\mathcal{I+L+L}^2}$, and corresponding results are shown in Fig.~\ref{testc}. From the figures, we can see that $\bm{\mathcal{L}}$ is the best choice compared with two candidates. So, we use $\theta\bm{\mathcal{L}}$  as $\bm{\mathcal{C}}$. Other well-designed $\bm{\mathcal{C}}$ can also replace $\bm{\mathcal{L}}$ here.

\begin{figure}[htbp]
\centering
\subfigure[DGI: BlogCatalog]
{
    \begin{minipage}[b]{.3\linewidth}
        \centering
        \includegraphics[scale=0.25]{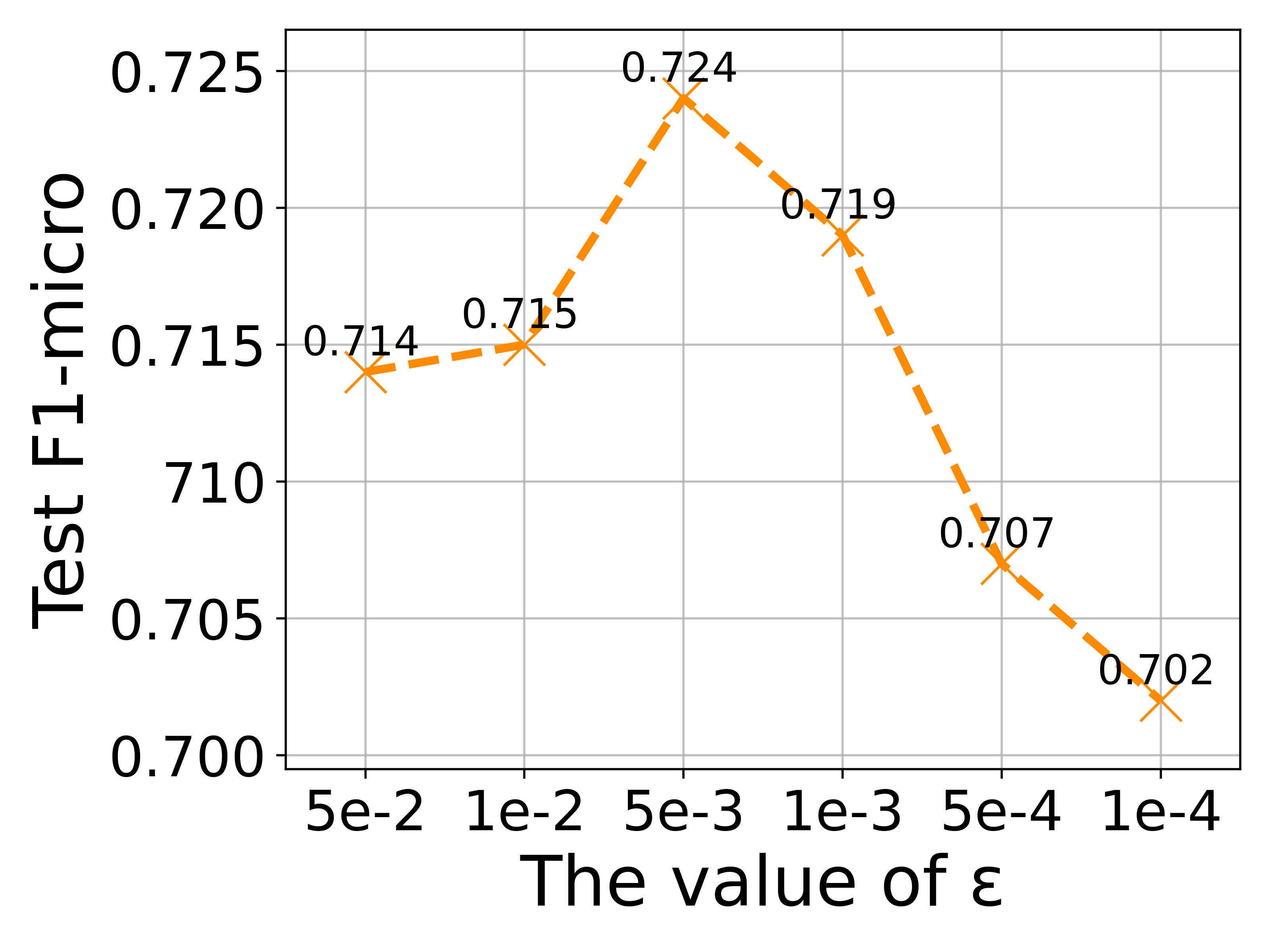}
    \end{minipage}
}
\subfigure[GRACE: BlogCatalog]
{
 	\begin{minipage}[b]{.3\linewidth}
        \centering
        \includegraphics[scale=0.25]{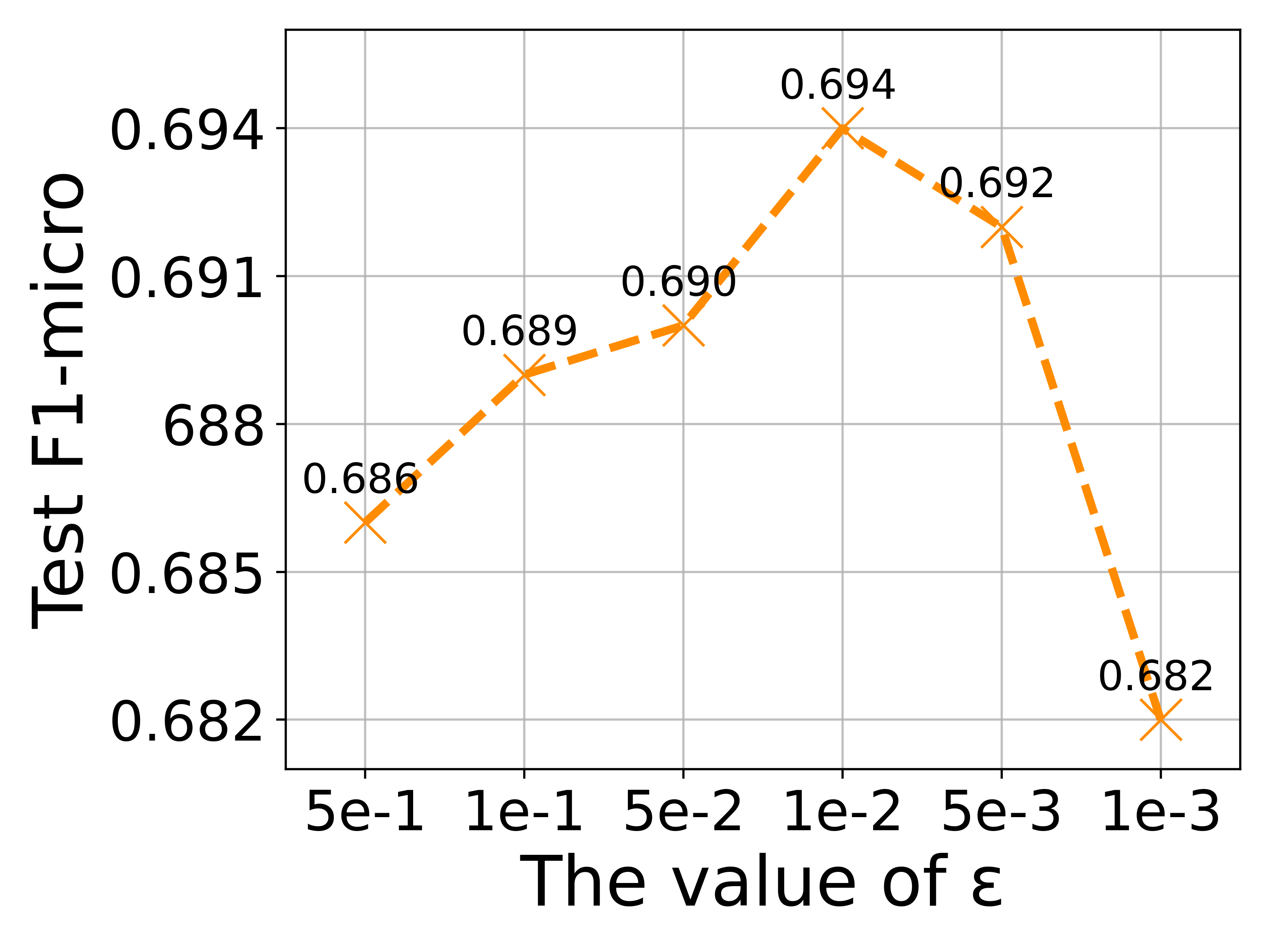}
    \end{minipage}
}
\subfigure[CCA-SSG: BlogCatalog]
{
 	\begin{minipage}[b]{.3\linewidth}
        \centering
        \includegraphics[scale=0.25]{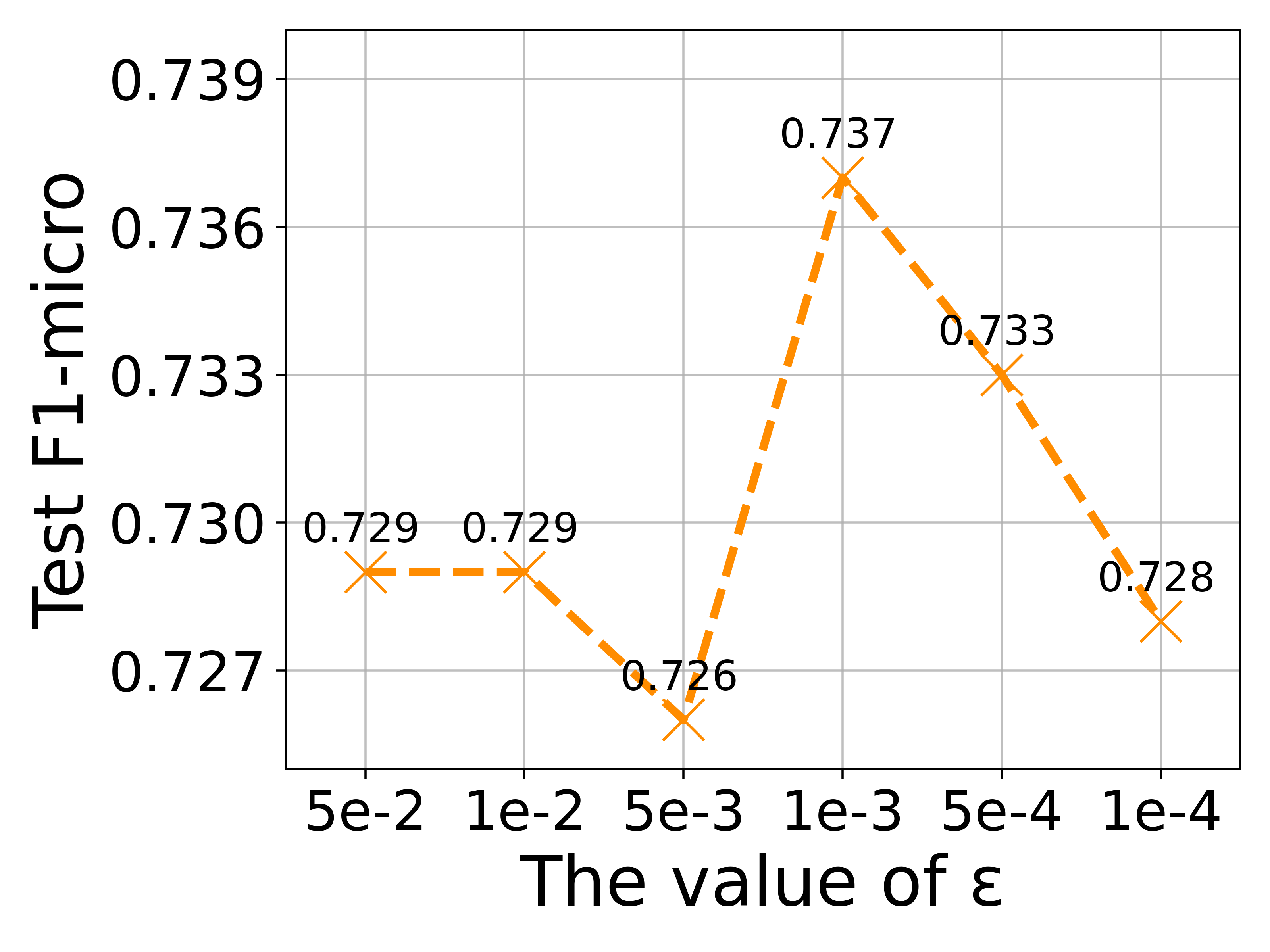}
    \end{minipage}
}
\caption{Analysis of the hyper-parameter $\epsilon$ on BlogCatalog.}
\label{epi_cora}
\end{figure}

\textbf{Analysis of $\epsilon$.} The $\epsilon$ in eq.~\eqref{target} controls the strength of entropy regularization, and in eq.~\eqref{solution} also controls the smoothness of exponential. We vary the value of it and plot the results on BlogCatalog in Fig.~\ref{epi_cora}. From the results, we know that $\epsilon$ is a sensitive parameter for SpCo. If $\epsilon$ is too small, the effect of entropy term will diminish. And if $\epsilon$ is too large, the entropy term will interfere the molding of new structure. More results on Cora are given in Appendix~\ref{Hyper-parameter Sensitivity}.
\section{Conclusion}
\label{conclusion}
In this paper, we fundamentally explore the topological augmentation of GCL from spectral domain. We propose contrastive invariance theorem, and discover a general augmentation (GAME) rule, which deepen our understanding of the essence of GCL. Then, we propose a general augmentation plug-in based on GAME rule, SpCo, to boost existing GCL methods. Extensive experiments verify the effectiveness of SpCo.

\textbf{Limitations and broader impact.} On potential limitation is that this work mainly focuses on the homophily graph, rather than the heterophily graphs \citep{bo2021beyond}, where high-frequency information is more useful. Despite the great development of GCL, some theoretical foundations are still lacking.  Our work points out the great potential of graph spectrum in GCL, and may open a new path to understand and design GCL. Other than that, we do not foresee any direct negative impacts on the society.
\begin{ack}
This work is supported in part by the National Natural Science Foundation of China (No. U20B2045, 62192784, 62172052, 62002029, U1936014).
\end{ack}

\medskip

% \setcitestyle{numbers}
\bibliographystyle{plain}
\bibliography{bib}

\section*{Checklist}

% %%% BEGIN INSTRUCTIONS %%%
% The checklist follows the references.  Please
% read the checklist guidelines carefully for information on how to answer these
% questions.  For each question, change the default \answerTODO{} to \answerYes{},
% \answerNo{}, or \answerNA{}.  You are strongly encouraged to include a {\bf
% justification to your answer}, either by referencing the appropriate section of
% your paper or providing a brief inline description.  For example:
% \begin{itemize}
%   \item Did you include the license to the code and datasets? \answerYes{See Section~\ref{gen_inst}.}
%   \item Did you include the license to the code and datasets? \answerNo{The code and the data are proprietary.}
%   \item Did you include the license to the code and datasets? \answerNA{}
% \end{itemize}
% Please do not modify the questions and only use the provided macros for your
% answers.  Note that the Checklist section does not count towards the page
% limit.  In your paper, please delete this instructions block and only keep the
% Checklist section heading above along with the questions/answers below.
% %%% END INSTRUCTIONS %%%

\begin{enumerate}

\item For all authors...
\begin{enumerate}
  \item Do the main claims made in the abstract and introduction accurately reflect the paper's contributions and scope?
    \answerYes{}
  \item Did you describe the limitations of your work?
    \answerYes{See Section~\ref{conclusion}.}
  \item Did you discuss any potential negative societal impacts of your work?
    \answerYes{See Section~\ref{conclusion}.}
  \item Have you read the ethics review guidelines and ensured that your paper conforms to them?
    \answerYes{}
\end{enumerate}

\item If you are including theoretical results...
\begin{enumerate}
  \item Did you state the full set of assumptions of all theoretical results?
    \answerYes{}
        \item Did you include complete proofs of all theoretical results?
    \answerYes{}
\end{enumerate}

\item If you ran experiments...
\begin{enumerate}
  \item Did you include the code, data, and instructions needed to reproduce the main experimental results (either in the supplemental material or as a URL)?
    \answerYes{We include the code, data, and instructions in the supplemental material}
  \item Did you specify all the training details (e.g., data splits, hyperparameters, how they were chosen)?
    \answerYes{See Section~\ref{experiments} and Appendix~\ref{Experimental Details}.}
        \item Did you report error bars (e.g., with respect to the random seed after running experiments multiple times)?
    \answerYes{See Section~\ref{Node classification} and Appendix~\ref{node classification}.}
        \item Did you include the total amount of compute and the type of resources used (e.g., type of GPUs, internal cluster, or cloud provider)?
    \answerYes{See Appendix~\ref{Operating Environment}.}
\end{enumerate}

\item If you are using existing assets (e.g., code, data, models) or curating/releasing new assets...
\begin{enumerate}
  \item If your work uses existing assets, did you cite the creators?
    \answerYes{}
  \item Did you mention the license of the assets?
    \answerNo{We were unable to find the license for the assets we used.}
  \item Did you include any new assets either in the supplemental material or as a URL?
    \answerYes{}
  \item Did you discuss whether and how consent was obtained from people whose data you're using/curating?
    \answerNA{The datasets are public benchmarks.}
  \item Did you discuss whether the data you are using/curating contains personally identifiable information or offensive content?
    \answerNA{The datasets are public benchmarks.}
\end{enumerate}

\item If you used crowdsourcing or conducted research with human subjects...
\begin{enumerate}
  \item Did you include the full text of instructions given to participants and screenshots, if applicable?
    \answerNA{}
  \item Did you describe any potential participant risks, with links to Institutional Review Board (IRB) approvals, if applicable?
    \answerNA{}
  \item Did you include the estimated hourly wage paid to participants and the total amount spent on participant compensation?
    \answerNA{}
\end{enumerate}

\end{enumerate}

\newpage
\appendix
\section{Proof and Derivation}
\subsection{Proof of Contrastive Invariance}
\label{proof_1}
\begin{proof}
We start our proof from the contrastive loss in eq.~\eqref{contra}, rewritten as follows:
\begin{equation}
    \mathcal{L}(\bm{h_i^{V_1}},\bm{ h_i^{V_2}})=\log\frac{\exp(\theta\bm{(h_i^{V_1}}, \bm{h_i^{V_2}})/\tau)}{\exp(\theta(\bm{h_i^{V_1}}, \bm{h_i^{V_2}})/\tau)+\sum\limits_{k\neq i}\exp(\theta(\bm{h_i^{V_1}}, \bm{h_k^{V_2}})/\tau)},
\end{equation}
Here, for simplification, we set $\tau=1$, $\theta$ is dot product similarity, the number of GCN layer is 1 and without non-linear activation. In the case study, we utilize augmentations $\bm{A}$ and $\bm{V}$ as two inputs. And then, for the optimization of node $i$, we have:
\begin{subequations}
\begin{align} 
\mathcal{L}(\bm{h_i^{A}},\bm{h_i^{V}})&=\log \frac{\exp(\bm{h_i^{A}}\bm{h_i^{V^\top}})}{\sum\limits_{k}\exp(\bm{h_i^{A}}\bm{h_k^{V^\top}})} \notag \\
&=\bm{h_i^{A}}\bm{h_i^{V^\top}}-\log\sum\limits_{k}\exp(\bm{h_i^{A}}\bm{h_k^{V^\top}}) \notag \\
&\leq \bm{h_i^{A}}\bm{h_i^{V^\top}}-\log N\cdot\exp\left(\frac{\sum\limits_{k}\bm{h_i^{A}}\bm{h_k^{V^\top}}}{N}\right) \label{1_1} \\
&= \bm{h_i^{A}}\bm{h_i^{V^\top}}-\log N-\frac{1}{N}\sum\limits_{k}\bm{h_i^{A}}\bm{h_k^{V^\top}} \notag \\
&\Rightarrow  \bm{h_i^{A}}\bm{h_i^{V^\top}}-\frac{1}{N}\sum\limits_{k}\bm{h_i^{A}}\bm{h_k^{V^\top}}. \label{1_2}
\end{align}
\end{subequations}
In formula~\eqref{1_1} we utilize inequality of arithmetic and geometric means, where given any N positive numbers, they meet $\frac{x_1+\dots+x_N}{N}\geq \sqrt[N]{x_1\dots x_N}$. And in formula~\eqref{1_2}, we neglect the constant $\log N$. Then, we gather the losses of all nodes as follows:

\begin{equation}
\begin{aligned} 
\label{asrairiro}
\mathcal{L} &= \sum\limits_i \mathcal{L}(\bm{h_i^{A}},\bm{h_i^{V}})\\
&\leq \sum\limits_i\bm{h_i^{A}}\bm{h_i^{V^\top}}-\sum\limits_i\frac{1}{N}\sum\limits_{k}\bm{h_i^{A}}\bm{h_k^{V^\top}} \\
&= tr(\bm{H^{A}}\bm{H^{V^\top}})-\frac{1}{N}sum(\bm{H^{A}}\bm{H^{V^\top}}),
\end{aligned}
\end{equation}

where $\bm{H^{A}}$ and $\bm{H^{V^\top}}$ denote the embeddings matrices for all nodes under $\bm{A}$ and $\bm{V}$, respectively. $tr(\bm{U})$ means the trace of matrix $\bm{U}$, and $sum(\bm{U})$ means the sum of all elements in matrix $\bm{U}$. Please recall that $\bm{V}$ is composed of different eigenspaces of $\bm{A}$, so we can have that $\bm{A=U\Lambda U^\top}$ and $\bm{V=U\Gamma U^\top}$, where $\bm{U}$ is the collection of eigenspaces, and $\bm{\Lambda}=diag(\lambda_1,\dots,\lambda_N)$ and $\bm{\Gamma}=diag(\gamma_1,\dots,\gamma_N)$ are the diagonal weight matrices of different eigenspaces. Because we simplify that only one layer GCN without activation function is used, we have the following formula:
\begin{equation}
\label{zxoari}
    \bm{H^{A}}\bm{H^{V^\top}}=\bm{AXWWXV},
\end{equation}
where $\bm{W}$ is learnable parameters of the encoder. Then, we set $\bm{M=XWWX}$. And thus, we can view $\bm{M}$ as the similarities of features between every two nodes after projection. As mentioned in previous studies, linked nodes often have similar features \citep{mcpherson2001birds}, and thus their similarity should be also higher after projection. Therefore w.l.o.g., we assume that transformed features between nodes present a kind of high-order proximity, given in \citep{DBLP:conf/kdd/ZhangCWPY018}:
\begin{Assumption}
\textbf{(High-order Proximity)} $\bm{M}=w_0+w_1\bm{A}+w_2\bm{A^2}+\dots+w_q\bm{A^q}$, where $\bm{A^i}$ means matrix multiplications between $i$ $\bm{A}$s, and $w_i$ is the weight of that term.
\end{Assumption}
In other words, this assumption aims to expand $\bm{M}$ with the weighted sum of different orders of $\bm{A}$. Furthermore, we have that:

\begin{Theorem}
\label{emakelm}
When $q\geq N-1$, $\bm{M=U\Theta U^\top}$, where $\Theta=diag(\theta_1,\dots,\theta_N)$. And $\{\theta_1,\dots,\theta_N\}$ are N different parameters, if $\{\lambda_1,\dots,\lambda_N\}$ are N different frequency amplitudes.
\end{Theorem}

\begin{proof}
\begin{equation}
\begin{aligned}
\bm{M}&=w_0+w_1\bm{A}+w_2\bm{A}^2+\dots+w_q\bm{A}^q\\
&=\bm{U}w_0\bm{I}\bm{U}^\top+\bm{U}w_1\bm{\Lambda} \bm{U}^\top+\bm{U}w_2\bm{\Lambda}^2 \bm{U}^\top+\dots+\bm{U}w_q\bm{\Lambda}^q\bm{U}^\top\\
&=\bm{U}[w_0\bm{I}+w_1\bm{\Lambda}+w_2\bm{\Lambda}^2+\dots+w_q\bm{\Lambda}^q]\bm{U}^\top\\
&=\bm{U}\begin{bmatrix} \sum\limits_{i=0}^q w_i\lambda_1^i &  & &\\  & \sum\limits_{i=0}^q w_i\lambda_2^i & &\\  &  &\dots & \\ &  & &\sum\limits_{i=0}^q w_i\lambda_N^i\end{bmatrix}\bm{U}^\top\\
\end{aligned}
\end{equation}
We firstly give the following equation set:
\begin{equation}
\label{set}
\left\{
\begin{aligned}
w_0+w_1\lambda_1+&\dots+w_q\lambda_1^q=\theta_1\\
w_0+w_1\lambda_2+&\dots+w_q\lambda_2^q=\theta_2\\
&\dots\\
w_0+w_1\lambda_N+&\dots+w_q\lambda_N^q=\theta_N,\\
\end{aligned}
\right.\\
\end{equation}

where $\{\theta_1,\dots,\theta_N\}$ are N randomly given parameters. And the determinant of coefficient of this equation set is $\bm{\mathbb{D}}=\left |\begin{array}{cccc}
1 &\lambda_1&\dots   & \lambda_1^q \\
1 &\lambda_2&\dots   & \lambda_2^q \\
 &&\dots   &\\
1 &\lambda_N&\dots   & \lambda_N^q \\
\end{array}\right|$. From the linear algebra, we know that if (1) $q=N-1$, $\bm{\mathbb{D}}$ is the Vandermonde determinant, and then $\bm{\mathbb{D}}=\prod_{N\geq i\geq j\geq1}(\lambda_i-\lambda_j)$. If $\forall i,j\in[1, N]$, we all have $\lambda_i\neq\lambda_j$, so $\bm{\mathbb{D}}\neq0$. In this case, the coefficient matrix is a full-rank matrix, and there will be one unique solution of the set~\eqref{set}. If $\exists$ one pair of $i,j\in[1, N]$ and $\lambda_i=\lambda_j$, we can force $\theta_i$ and $\theta_j$ to be equal, which means that the encoder deploys the same effect on the repeated frequencies. The number of repeated frequencies is considerably smaller that that of different frequencies, so we neglect this case; (2) $q>N-1$, the number of unknown numbers is larger than that of equations in set~\eqref{set}, so there are infinite solutions. Above all, theorem~\ref{emakelm} is proved.
\end{proof}
Theorem~\ref{emakelm} points that $\bm{M}$ can be easily rewritten in $U\Theta U^\top$, if we use sufficient orders of $\bm{A}$ to approach $\bm{M}$. Combined with this theorem, we resume the derivation of eq.~\eqref{zxoari} as follows:
\begin{equation}
\begin{aligned}
\bm{H^{A}}\bm{H^{V^\top}}&=\bm{AMV}\\
& =\bm{U}\bm{\Lambda} \bm{U}^\top \bm{M}\bm{U}\bm{\Gamma} \bm{U}^\top=\bm{U}\bm{\Lambda}\bm{\Theta}\bm{\Gamma} \bm{U}\\
& =\bm{U}\begin{bmatrix} \lambda_1\theta_1\gamma_1 &  & &\\  & \lambda_2\theta_2\gamma_2 & &\\  &  &\dots & \\ &  & &\lambda_n\theta_n\gamma_n\end{bmatrix}\bm{U}^\top\\
& =\lambda_1\theta_1\gamma_1\bm{u_1u_1}^\top+\lambda_2\theta_2\gamma_2\bm{u_2u_2}^\top+\dots+\lambda_N\theta_N\gamma_N\bm{u_Nu_N}^\top
\end{aligned}
\end{equation}
Therefore, we have:
\begin{equation}
\begin{aligned}
tr(\bm{H^{A}}\bm{H^{V^\top}})=\sum\limits_i \lambda_i\theta_i\gamma_i,\ \text{and}\   sum(\bm{H^{A}}\bm{H^{V^\top}})=\sum\limits_i \lambda_i\theta_i\gamma_i sum(\bm{u}_i\bm{u}_i^\top)
\end{aligned}
\end{equation}
Finally, plug this equation into eq.~\eqref{asrairiro}, we can get:

\begin{subequations}
\begin{align} 
\mathcal{L} &\leq  tr(\bm{H^{A}}\bm{H^{V^\top}})-\frac{1}{N}sum(\bm{H^{A}}\bm{H^{V^\top}}) \notag\\
&=\ \sum\limits_i \lambda_i\theta_i\gamma_i[1-\frac{1}{N}sum(\bm{u}_i\bm{u}_i^\top)] \notag \\
&\leq (1+N)\sum\limits_i \lambda_i\theta_i\gamma_i \label{1_3}\\
&=\frac{1+N}{2}\sum\limits_i \theta_i[\lambda_i^2+\gamma_i^2-(\lambda_i-\gamma_i)^2]  \notag\\
&\leq \frac{1+N}{2}\sum\limits_i \theta_i[2-(\lambda_i-\gamma_i)^2].  \label{1_4}
\end{align}
\end{subequations}
In eq.~\eqref{1_3}, we utilize the following lemma:
\begin{Lemma}
$sum(\bm{u}_i\bm{u}_i^\top)\geq -N^2$.
\end{Lemma}
\begin{proof}
We have known that $\bm{u_i^\top u_i}=u_{i1}^2+\dots+u_{iN}^2=1$, which implies that $\forall j\in[1, N], u_{ij}\in(-1, 1)$. Thus, $\forall j,k\in[1, N], u_{ij}u_{ik}\in(-1, 1)$. Finally, $sum(\bm{u}_i\bm{u}_i^\top)=\sum\limits_{j,k}u_{ij}u_{ik}>\sum\limits_{j,k}(-1)=-N^2$.
\end{proof}
And in eq.~\eqref{1_4}, we utilize the fact that for $\bm{A}$, amplitudes of its frequencies lay in [-1, 1], while for $\bm{V}$, amplitudes of its all frequencies equal to 1. Therefore, $\lambda_i^2\leq1$ and $\gamma_i^2\leq1$. Similarly for $\mathcal{L}(\bm{h_i^{V_2}},\bm{ h_i^{V_1}})$, we also have the same upper bound. Therefore, we have the following results:
\begin{equation}
    \mathcal{L}_{InfoNCE}=\sum\limits_i \frac{1}{2}\left(\mathcal{L}(\bm{h_i^{V_1}}, \bm{h_i^{V_2}})+\mathcal{L}(\bm{h_i^{V_2}}, \bm{h_i^{V_1}})\right)\leq \frac{1+N}{2}\sum\limits_i \theta_i[2-(\lambda_i-\gamma_i)^2].
\end{equation}
So, the contrastive invariance is proved.
\end{proof}

\subsection{Derivation of optimization objective}
\label{Derivation of optimization objective}
In this subsection, we provide detailed derivation of optimization objective in eq.~\eqref{target}. As shown in section~\ref{mmmmmmmm}, we attempt to learn an augmented graph from original graph based on the GAME rule. Based on eigenvalue perturbation formula~\eqref{pertu} removing the high-order term $\mathcal{O}(||\Delta\bm{A}||)$, we have:
\begin{subequations}
\begin{align} 
\Delta \lambda_i& \approx \bm{u}_i^\top\Delta\bm{A}\bm{u}_i-\lambda_i\bm{u}_i^\top\Delta\bm{D}\bm{u}_i \notag\\
& = \sum_{m,n} [(\bm{u}_i\bm{u}_i^\top)*\Delta \bm{A}]_{m,n} - \lambda_i \bm{u}_i^\top\Delta \bm{D} \bm{u}_i \label{b}\\
& = <\bm{S}_i, \Delta \bm{A}> - \lambda_i \bm{u}_i^\top\Delta \bm{D} \bm{u}_i. \label{c}
\end{align}
\end{subequations}
In eq.~\eqref{b}, we use '*' to represent element-wise product, and use $<\bm{P}, \bm{Q}>$ to represent the sum of all elements of $\bm{P} * \bm{Q}$ in eq.~\eqref{c}. Before further derivation, we give a lemma as following:
\begin{Lemma}
\label{lemma}
$|\lambda_i\bm{u}_i^\top\Delta\ \bm{D}\bm{u}_i|\ \leq N|\lambda_i|$.
\end{Lemma}
This lemma is proved in Appendix~\ref{proof_lemma}. Then with Lemma~\ref{lemma}, we gather the changes of all eigenvalues together, and have the following two theorems:
\begin{Theorem}
Given $\Delta_{total} =\sum_i \alpha_i|\Delta \lambda_i|$, we have
\label{lllalla}
\begin{align*}
\Delta_{total} \iff \sum_i \alpha_i\left|<\bm{S}_i, \Delta \bm{A}>\right|\geq \sum_i\alpha_i\left|<\bm{S}_i, \Delta_{\bm{A+}}>\right|-\sum_j a_j\left|<\bm{S_j}, \Delta_{\bm{A-}}>\right|=\Delta_+-\Delta_-,
\end{align*}
where $\alpha_i\geq 0$ is the weight for the change of $\lambda_i$. Formally, we have $\Delta\bm{A}=\Delta_{\bm{A+}}-\Delta_{\bm{A-}}$, where $\Delta_{\bm{A+}}$ and $\Delta_{\bm{A-}}$ indicate which edge is added and deleted, respectively. 
\end{Theorem}

\begin{Theorem}
\label{close}
Given N eigenspaces $\left[\bm{S_1},\dots,\bm{S_N}\right]$, if $|<\bm{S}_i,\Delta \bm{A}>|$ is maximized, then $\forall j,\ |<\bm{S_j},\ \Delta \bm{A}>| \to 0$.
\end{Theorem}
The proof is given in Appendix~\ref{proof_2} and~\ref{proof_3}. Theorem~\ref{close} implies that $\Delta \bm{A}$ can only be related to one of these eigenspaces. Please recall that we need $|\Delta \lambda_i|$ in $\mathcal{F_L}$ is smaller than that in $\mathcal{F_H}$. Therefore, we first make $\alpha_i$ is larger for $\Delta \lambda_i$ in $\mathcal{F_H}$ than in $\mathcal{F_L}$. Then, we maximize $\Delta_{total}$, and $\Delta \bm{A}$ will only be related to $\bm{S}_i$ in $\mathcal{F_H}$. That means $|\Delta \lambda_i|$ in $\mathcal{F_H}$ will be larger. Particularly, Theorem~\ref{lllalla} indicates that we can maximize $\Delta_+$ and minimize $\Delta_-$ simultaneously to maximize $\Delta_{total}$. For $\Delta_+$, we can maximize it by following formula:
\begin{subequations}
\begin{align} 
\Delta_+ &=\sum_i \alpha_i\left|<\bm{S}_i, \Delta_{\bm{A+}}>\right| = |\sum_i \alpha_i<\bm{S}_i, \Delta_{\bm{A+}}>|  \notag\\
&=|<\sum_i\alpha_i\bm{u}_i\bm{u}_i^\top,\ \Delta_{\bm{A+}}>| =\left|<\bm{U}g(\lambda)\bm{U}^\top, \ \Delta_{\bm{A+}}>\right|=\left|<\bm{\mathcal{C}}, \ \Delta_{\bm{A+}}>\right|.  \label{bb}
\end{align}
\end{subequations}
In eq.~\eqref{bb}, we define $\bm{\mathcal{C}}=\bm{U}g(\lambda)\bm{U}^\top$ and $\alpha_i=g(\lambda_i)$, where $g(\lambda)$ should be a monotone increasing function. As can be seen, if $\Delta_{\bm{A+}}$ is learnt to be closed to some predefined $\bm{\mathcal{C}}$, $\Delta_+$ will increase.

\textbf{Setting $\bm{\mathcal{C}}$} We notice that the graph spectrum of Laplacian $\bm{\mathcal{L}}$ does meet our need about $g(\lambda)$. Therefore, we simply set $\bm{\mathcal{C}}=\Theta\bm{\mathcal{L}}$, where $\Theta$ is a parameter. We change $\Theta$ as the training goes.

With eq.~\eqref{bb}, we give the following optimization objective:
\begin{equation}
    \label{targett}
    \mathcal{J} = <\bm{\mathcal{C}}, \ \Delta_{\bm{A+}}>^2+  \epsilon H(\Delta_{\bm{A+}})+<\bm{f},\Delta_{\bm{A+}}\mathbbm{1}_n-\bm{a}>+<\bm{g},\Delta_{\bm{A+}}^\top\mathbbm{1}_n-\bm{b}>,
\end{equation}
where $H(\bm{P})$ is the entropy regularization, defined as $-\sum_{i, j}\bm{P}_{i, j}(\log(\bm{P}_{i, j})-1)$ \citep{ot}, and $\epsilon$ is the weight of this term. This term exposes more edges in $\Delta_{\bm{A+}}$ to optimization. The last two terms are Lagrange constraint conditions, where $\bm{f}\in \mathbb{R}^{N}$ and $\bm{g}\in \mathbb{R}^{N}$ are Lagrange multipliers, and $\bm{a}\in \mathbb{R}^{N\times 1}$ and $\bm{b}\in \mathbb{R}^{N\times 1}$ are distributions that the row and column sums of $\Delta_{\bm{A+}}$ should meet.

\subsubsection{Proof of Lemma~\ref{lemma}}
\label{proof_lemma}
\begin{proof}
\begin{align}
|\lambda_i\bm{u}_i^\top\Delta \bm{D}\bm{u}_i|&=|\lambda_i| |\bm{u}_i^\top \bm{diag}(d_1,\dots,d_N)\bm{u}_i|=|\lambda_i|\left[|d_1|\bm{u_{i1}}^2+\dots+|d_N|\bm{u_{iN}}^2\right] \notag \\
&\leq |d|_{max}|\lambda_i|\left[\bm{u_{i1}}^2+\dots+\bm{u_{iN}}^2\right]=|d|_{max}|\lambda_i| \leq N|\lambda_i|. \notag
\end{align}
\end{proof}
In this proof, we utilize two facts: (1) the length of any eigenvector is 1, and (2) the change of degree matrix does not exceed N (in the limiting case, one node becomes from isolated to reaching all other nodes, where $d_{max}$ equals to N).

\subsubsection{Proof of Theorem~\ref{lllalla}}
\label{proof_2}
\begin{proof}
\begin{subequations}
\begin{align}
\Delta_{total} &=\sum_i a_i|\Delta \lambda_i| \notag\\
&=\sum_i a_i\left|<\bm{S}_i, \Delta \bm{{A}}> - \lambda_i \bm{u}_i^T\Delta \bm{D} \bm{u}_i\right| \notag\\
&\geq\sum_i a_i\left|<\bm{S}_i, \Delta \bm{{A}}>\right|-\left|\lambda_i \bm{u}_i^T\Delta \bm{D} \bm{u}_i\right| \notag\\
&=\sum_i a_i\left|<\bm{S}_i, \Delta \bm{A}>\right|-\sum_k\left|\lambda_k \bm{u_k}^T\Delta \bm{D} \bm{u_k}\right| \notag\\
&\geq \sum_i a_i\left|<\bm{S}_i, \Delta \bm{A}>\right|-N\sum_k |\lambda_k| \label{1_5}\\
&\geq \sum_i a_i\left|<\bm{S}_i, \Delta \bm{A}>\right|-N^2 \label{1_6}\\
&\iff \sum_i a_i\left|<\bm{S}_i, \Delta \bm{A}>\right|  \label{1_7}\\
&\geq \sum_i a_i|<\bm{S}_i, \Delta_{\bm{A+}}>|-\sum_j a_j|<\bm{S}_i, \Delta_{\bm{A-}}>|.  \notag
\end{align}
\end{subequations}
In eq.~\eqref{1_5}, we utilize the result of lemma~\ref{lemma}, and in eq.~\eqref{1_6} we again utilize that the amplitudes of frequencies of $\bm{A}$ are between [-1, 1]. For simplification, we ignore the constant $N^2$ in eq.~\eqref{1_7}.
\end{proof}

\subsubsection{Proof of Theorem~\ref{close}}
\label{proof_3}
\begin{proof}
To prove the Theorem~\ref{close}, we firstly prove the following lemma:
\begin{Lemma}
$<\bm{S}_i, \bm{S_j}>=0$, when $i\neq j$.
\end{Lemma}
\begin{proof}
We have that $<\bm{S}_i, \bm{S_j}>=<\bm{u_iu_i^\top}, \bm{u_ju_j^\top}>=\sum\limits_{m, n}(\bm{u_iu_i^\top})* (\bm{u_ju_j^\top})=\bm{u_i^\top u_ju_j^\top u_i}=0$, where we take advantage of any two different eigenvectors are orthogonal, or $\bm{u_i^\top u_j}=0$.
\end{proof}
With this lemma, we give the following derivation:
\begin{equation}
    2<\bm{S}_i,\ \Delta \bm{A}> =2\Delta \bm{A}_{11}\bm{S}_{i11}+2\Delta \bm{A}_{12}\bm{S}_{i12}+\dots+2\Delta \bm{A}_{NN}\bm{S}_{iNN}.
\end{equation}
Then we conduct category discussion on $<\bm{S}_i,\ \Delta \bm{A}>$:

1. For $<\bm{S}_i,\ \Delta \bm{A}>$ > 0, we have:
\begin{subequations}
\begin{align}
2|<\bm{S}_i,\ \Delta \bm{A}>|& =2\Delta \bm{A}_{11}\bm{S}_{i11}+2\Delta \bm{A}_{12}\bm{S}_{i12}+\dots+2\Delta \bm{A}_{NN}\bm{S}_{iNN} \notag \\
&=\sum\limits_{m,n}\Delta \bm{A}_{mn}^2+\bm{S}_{imn}^2-(\Delta \bm{A}_{mn}-\bm{S}_{imn})^2 \notag \\
&=\sum\limits_{m,n}\Delta \bm{A}_{mn}^2+||\bm{S}_i||_F^2-\sum\limits_{m,n}(\Delta \bm{A}_{mn}-\bm{S}_{imn})^2  \notag\\
&\leq N^2+||\bm{S}_i||_F^2-\sum\limits_{m,n}(\Delta \bm{A}_{mn}-\bm{S}_{imn})^2. \label{1_8}
\end{align}
\end{subequations}
In eq.~\eqref{1_8}, we make use of that the change of one element in $\Delta \bm{A}$ is in $[-1, 1]$. From eq.~\eqref{1_8}, we can infer that if $\Delta \bm{A}$ is learnt to be similar with some eigenspace $\bm{S}_i$, maximizing the left hand of above equation, then $\sum\limits_{m,n}(\Delta \bm{A}_{mn}-\bm{S}_{imn})^2$ will be minimized, which means $\Delta \bm{A}_{mn}\rightarrow\bm{S}_{imn}$. Then, if we test the similarity between $\Delta \bm{A}$ and another eigenspace $\bm{S}_{j}$, we have that:
\begin{subequations}
\begin{align}
<\bm{S_j},\ \Delta \bm{A}>&=\Delta \bm{A}_{11}\bm{S}_{j11}+\Delta \bm{A}_{12}\bm{S}_{j12}+\dots+\Delta \bm{A}_{NN}\bm{S}_{jNN} \notag \\
&\Rightarrow \bm{S}_{i11}\bm{S}_{j11}+\bm{S}_{i12}\bm{S}_{j12}+\dots+\bm{S}_{iNN}\bm{S}_{jNN} \notag \\
&=<\bm{S_j},\ \bm{S}_i>=0. \notag
\end{align}
\end{subequations}
2. For $<\bm{S}_i,\ \Delta \bm{A}>$ < 0, we have:
\begin{subequations}
\begin{align}
2|<\bm{S}_i,\ \Delta \bm{A}>|& =-2\Delta \bm{A}_{11}\bm{S}_{i11}-2\Delta \bm{A}_{12}\bm{S}_{i12}-\dots-2\Delta \bm{A}_{NN}\bm{S}_{iNN} \notag \\
&=\sum\limits_{m,n}\Delta \bm{A}_{mn}^2+\bm{S}_{imn}^2-(\Delta \bm{A}_{mn}+\bm{S}_{imn})^2 \notag \\
&=\sum\limits_{m,n}\Delta \bm{A}_{mn}^2+||\bm{S}_i||_F^2-\sum\limits_{m,n}(\Delta \bm{A}_{mn}+\bm{S}_{imn})^2  \notag\\
&\leq N^2+||\bm{S}_i||_F^2-\sum\limits_{m,n}(\Delta \bm{A}_{mn}+\bm{S}_{imn})^2. \notag
\end{align}
\end{subequations}
In this case, $\sum\limits_{m,n}(\Delta \bm{A}_{mn}+\bm{S}_{imn})^2$ will be minimized, which means $\Delta \bm{A}_{mn}\rightarrow-\bm{S}_{imn}$. Then, for another eigenspace $\bm{S}_{j}$, we have:
\begin{subequations}
\begin{align}
<\bm{S_j},\ \Delta \bm{A}>&=\Delta \bm{A}_{11}\bm{S}_{j11}+\Delta \bm{A}_{12}\bm{S}_{j12}+\dots+\Delta \bm{A}_{NN}\bm{S}_{jNN} \notag \\
&\Rightarrow -\bm{S}_{i11}\bm{S}_{j11}-\bm{S}_{i12}\bm{S}_{j12}-\dots-\bm{S}_{iNN}\bm{S}_{jNN} \notag \\
&=-<\bm{S_j},\ \bm{S}_i>=0. \notag
\end{align}
\end{subequations}
Above all, we prove that $\Delta \bm{A}$ can only be related to part of eigenspaces, rather than all of them.
\end{proof}

\subsection{Proof of Theorem~\ref{theorem 4}}
\label{proof_4}
\begin{proof}
Here, we start from eq.~\eqref{bbb}, and replace $(\Delta_{\bm{A+}})_{ij}$ with $\Delta_{ij}$, shown as following:
\begin{equation}
\begin{aligned}
\label{bbb_}
    f(\Delta_{ij})&=\frac{\partial \mathcal{J}}{\partial (\Delta_{\bm{A+}})_{ij}}=m_{ij}+2\bm{\mathcal{C}}_{ij}^2(\Delta_{\bm{A+}})_{ij}-\epsilon \log (\Delta_{\bm{A+}})_{ij}+\bm{f}_i+\bm{g}_j \\
    &=m_{ij}+2\bm{\mathcal{C}}_{ij}^2\Delta_{ij}-\epsilon\log{\Delta_{ij}}+\bm{f}_i+\bm{g}_j,
\end{aligned}
\end{equation}
where the domain of definition is $\Delta_{ij}\in(0, 1)$. We observe eq.~\eqref{bbb_} that $f(0)\rightarrow +\infty$ and $f(1)= m_{ij}+2\bm{\mathcal{C}}_{ij}^2+\bm{f}_i+\bm{g}_j$. Then, we conduct category discussion.

1. If $f(1)<0$, we have $\bm{\mathcal{C}}_{ij}^2 < -\frac{\bm{f}_i+\bm{g}_j+m_{ij}}{2}$,  and \ $\bm{f}_i+\bm{g}_j+m_{ij}<0$. In this case, by Existence Theorem of Zero Points, we know that there must be a point making $f(\Delta_{ij})=0$, and the loss $\mathcal{J}$ can obtain the extremum. This is the condition (1) in Theorem~\ref{theorem 4}.

2. If $f(1)>0$, we must let the minimal value of $f(\Delta_{ij})<0$, in order to guarantee there exists Zero Points. To get the minimal value, we need take the derivative about $f(\Delta_{ij})$ as follows:
\begin{equation}
    \frac{\partial f(\Delta_{ij})}{\partial \Delta_{ij}}=2\bm{\mathcal{C}}_{ij}^2-\frac{\epsilon}{\Delta_{ij}}.
\end{equation}
From above formula, the minimal value is $\Delta_{ij}=\frac{\epsilon}{2C_{ij}^2}$ by letting $\frac{\partial f(\Delta_{ij})}{\partial \Delta_{ij}}=0$. Then, we need to meet the following two conditions:
\begin{equation}
    \frac{\epsilon}{2\bm{\mathcal{C}}_{ij}^2}<1\ and \ f(\frac{\epsilon}{2C_{ij}^2})<0.
\end{equation}
The solution of above two conditions is the condition (2) in Theorem~\ref{theorem 4}.
\end{proof}

\subsection{Proof of Theorem~\ref{theorem 5}}
\label{proof_5}
\begin{proof}
To begin with, we introduce some basic definitions, which will be used in this part of proof.

\begin{Definition}
\textbf{(Hilbert's projective metric)}\citep{bushell1973hilbert} Given $\forall (x, x')\in(\mathbb{R}^n_{+})^2$, then the Hilbert's projective metric on $\mathbb{R}^n_{+}$ is defined as: \\
\centerline{$d(x, x')=\log \max\limits_{i,k}\frac{x_ix_k'}{x_i'x_k}$.}
\end{Definition}

This is an explicitly defined distance function on a bounded convex subset of the n-dimensional Euclidean space, which can greatly simplified the global convergence analysis of Sinkhorn \citep{franklin1989scaling}. This metric satisfies the triangular inequality, and $d(x, x')=0$ if $\exists s>0, x = sx'$.

\begin{Definition}
\textbf{(Contraction ratio)}\citep{franklin1989scaling} Given a positive n $\times$ n matrix $\bm{A}$, then the contraction ratio of $\bm{A}$ is defined as:\\
\centerline{$\gamma=\kappa({\bm{A}})=\sup \left\{\frac{d(\bm{Ay, Ay'})}{d(\bm{y, y'})}:\ \bm{y, y'}\in \mathbb{R}^n_+,\ \bm{y}\neq\alpha\bm{y'} \right\}$}
\end{Definition}
Specially, if we set $\aleph=\sup \left\{d(\bm{Ay, Ay'}):\ \bm{y, y'}\in \mathbb{R}^n_+,\ \bm{y}\neq\alpha\bm{y'} \right\}$, then $\gamma=\frac{\aleph^{1/2}-1}{\aleph^{1/2}+1}$.

With these two definitions, the authors in \citep{peyre2019computational} give the following lemma:

\begin{Lemma}
One has 
\begin{equation}
\label{llleee}
    ||\log(\bm{P}^{(l)})-\log(\bm{P^\star})||_{\infty} \leq \frac{1}{1-\gamma}\{d(\bm{r}^{(l)}, \bm{a})+d(\bm{c}^{(l)}, \bm{b})\},
\end{equation}
where $\bm{P}^{(l)}$ and $\bm{P^\star}$ are the intermediate result after $l$th Sinkhorn's Iteration and the final result, respectively. $\bm{r}^{(l)}$ and $\bm{c}^{(l)}$ are the row and column sum vectors of $\bm{P}^{(l)}$, and $\bm{a}$ and $\bm{b}$ are two predefined distributions.  
\end{Lemma}

Now, please recall that we utilize the trick to operate Sinkhorn's Iteration in eq.~\eqref{solution}. In this case, we set $l$ in eq.~\eqref{llleee} as 0, so the left hand of equation will be changed as:
\begin{equation}
    ||\log(\bm{P}^{(0)})-\log(\bm{P^\star})||_{\infty}
    =||\log(\bm{K_{+}})-\log(\Delta_{\bm{A+}})||_{\infty}.
\end{equation}
This is because before iteration, $\bm{P}^{(0)}$ is actually $\bm{K_{+}}$, and $\Delta_{\bm{A+}}$ is the final result of iteration. Then, we have the following derivation:

\begin{equation}
\begin{aligned}
\left|\log((\Delta_{\bm{A+}})_{ij})-\log((\bm{K_{+}})_{ij})\right| &\leq ||\log(\Delta_{\bm{A+}})-\log(\bm{K_{+}})||_{\infty} \\
& \leq \frac{1}{1-\gamma}\{d(\bm{r}^{(0)}, \bm{a})+d(\bm{c}^{(0)}, \bm{b})\}   \\
\left|\epsilon\log((\Delta_{\bm{A+}})_{ij})-2<\bm{\mathcal{C}}, \Delta_{\bm{A+}}'>\bm{\mathcal{C}}_{ij}\right| & \leq \frac{1}{\epsilon(1-\gamma)}\{d(\bm{r}^{(0)}, \bm{a})+d(\bm{c}^{(0)}, \bm{b})\}  \\
\left|\epsilon\log((\Delta_{\bm{A+}})_{ij})-2\bm{\mathcal{C}}_{ij}^2(\Delta_{\bm{A+}}')_{ij}-m_{ij}\right| & \leq \frac{1}{\epsilon(1-\gamma)}\{d(\bm{r}^{(0)}, \bm{a})+d(\bm{c}^{(0)}, \bm{b})\}  \\
\left|\epsilon((\Delta_{\bm{A+}})_{ij}-1)-2\bm{\mathcal{C}}_{ij}^2(\Delta_{\bm{A+}}')_{ij}-m_{ij}\right| & \leq \frac{1}{\epsilon(1-\gamma)}\{d(\bm{r}^{(0)}, \bm{a})+d(\bm{c}^{(0)}, \bm{b})\}  \\
\because \left|\epsilon(\Delta_{\bm{A+}})_{ij}-2\bm{\mathcal{C}}_{ij}^2(\Delta_{\bm{A+}}')_{ij}\right|-\epsilon-|m_{ij}| & \leq |\epsilon((\Delta_{\bm{A+}})_{ij}-1)-2\bm{\mathcal{C}}_{ij}^2(\Delta_{\bm{A+}}')_{ij}-m_{ij}| \\
\therefore \left|\epsilon(\Delta_{\bm{A+}})_{ij}-2\bm{\mathcal{C}}_{ij}^2(\Delta_{\bm{A+}}')_{ij}\right| &\leq \frac{1}{\epsilon(1-\gamma)}\{d(\bm{r}^{(0)}, \bm{a})+d(\bm{c}^{(0)}, \bm{b})\}+\epsilon+|m_{ij}|\\
\left|\frac{\epsilon}{2\bm{\mathcal{C}}_{ij}^2}(\Delta_{\bm{A+}})_{ij}-(\Delta_{\bm{A+}}')_{ij}\right| &\leq \frac{1}{2\bm{\mathcal{C}}_{ij}^2\epsilon(1-\gamma)}\{d(\bm{r}^{(0)}, \bm{a})+d(\bm{c}^{(0)}, \bm{b})\}\\
&+\frac{\epsilon+|m_{ij}|}{2\bm{\mathcal{C}}_{ij}^2}.
\end{aligned}
\end{equation}

If we set $\alpha=\frac{\epsilon}{2\bm{\mathcal{C}}_{ij}^2}$, we have:
\begin{equation}
\left|\alpha(\Delta_{\bm{A+}})_{ij}-(\Delta_{\bm{A+}})_{ij}'\right|\leq\frac{\alpha}{\epsilon^2(1-\gamma)}\{d(r^{(0)},\bm{a})+d(c^{(0)},\bm{b})\}+\alpha(1+\frac{|m_{ij}|}{\epsilon}).
\end{equation}
\end{proof}

\section{More details of Section \ref{case study}}
\subsection{Experimental settings}
\label{Experimental settingss}
In Fig.~\ref{case_study_model}, we show the GCL model used in following case study. Here, we provide more experimental settings about case study. As description, we use a shared GCN to encode $\bm{A}$ and $\bm{V}$ and get their nodes embeddings as $\bm{H_A}$ and $\bm{H_V}$. In the training, we set dimensions of $\bm{H_A}$ and $\bm{H_V}$ as 8, learning rate as 0.001, $\tau$ in eq.~\ref{contra} as 0.5 and weight decay as 0. We train the model 300 epochs, and then test the quality of $\bm{H_A}$. In the test phrase, we follow DGI, and the split of each dataset is consistent with that in table~\ref{statistics} in Appendix~\ref{app_dataset}, and we only test the case where training set contains 20 nodes per class.

\subsection{More results of case study}
\label{More results of case study}
\begin{figure}[htbp]
\centering
\begin{minipage}[h]{0.4\linewidth}
\centering
\includegraphics[scale=0.13]{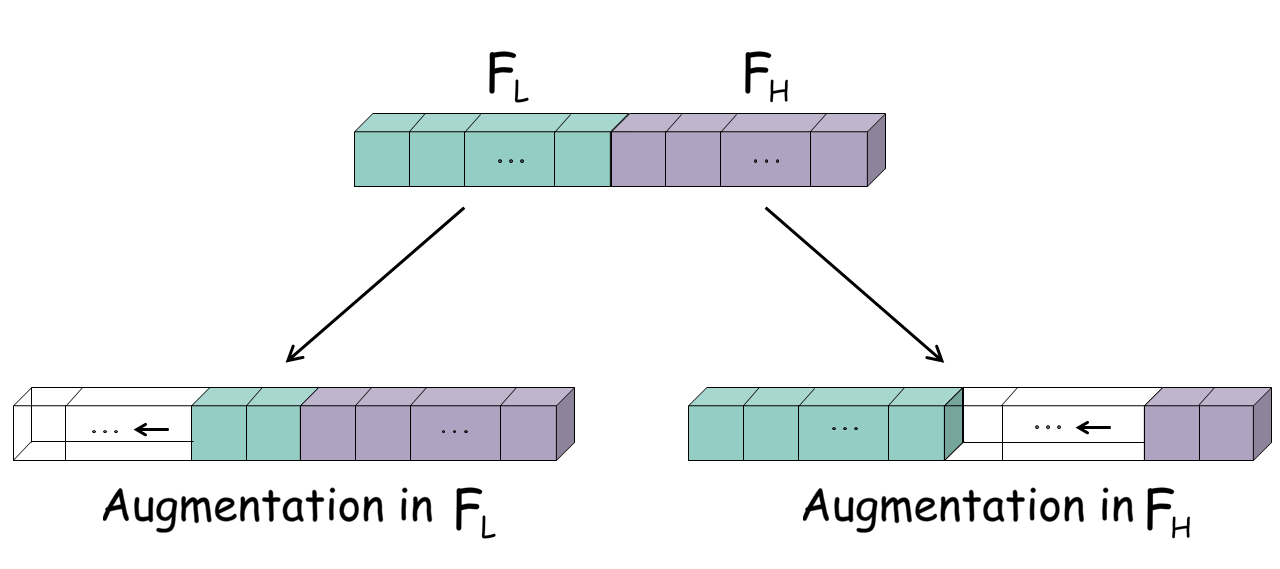}
\caption{New generation of V, where we follow high-to-low frequency order in both low-frequency and high-frequency part.}\label{New generation of V}
\end{minipage}
\hspace{50pt}
\begin{minipage}[h]{0.4\linewidth}
\centering
\includegraphics[scale=0.04]{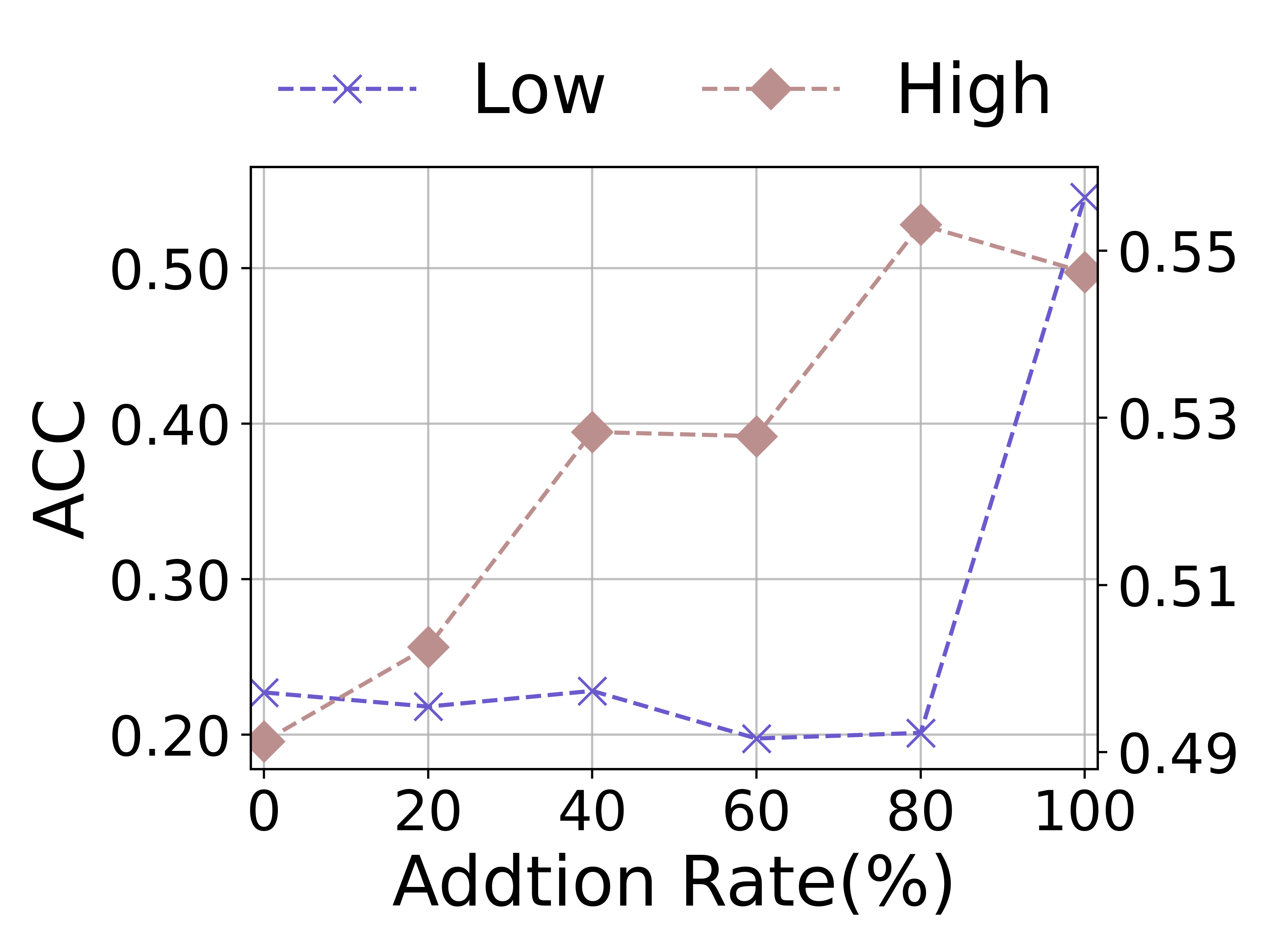}
\caption{The results of inverse adding order on Cora.}\label{The results of inverse}
\end{minipage}
\end{figure}
In section \ref{case study}, we design a simple GCL framework to contrast $\bm{A}$ and generated $\bm{V}$. When constructing $\bm{V}$, we conduct augmentations in $\mathcal{F_L}$ (or $\mathcal{F_H}$) by adding eigenspaces in $\mathcal{F_L}$ (or $\mathcal{F_H}$) back in low-to-high frequency order. Here, we consider the inverse high-to-low frequency order. The process is shown in Fig.~\ref{New generation of V}, where $\bm{V}$ augmenting 20\% in $\mathcal{F_L}$ is $\bm{u_{0.8*N/2}u_{0.8*N/2}}^\top+\dots+\bm{u_{(N-1)/2}u_{(N-1)/2}}^\top+\bm{u_{N/2}u_{N/2}}^\top+\dots+\bm{u_Nu_N}^\top$. Similarly, $\bm{V}$ augmenting 20\% in $\mathcal{F_H}$ is $\bm{u_1u_1}^\top+\dots+\bm{u_{N/2}u_{N/2}}^\top+\bm{u_{1.8*N/2}u_{1.8*N/2}}^\top+\dots+\bm{u_Nu_N}^\top$. We take Cora as an example, and report the results in Fig.~\ref{The results of inverse}. As can be observed, we get the similar conclusions as in section~\ref{case study}: (1) From the curve marked by “Low”, the performances remain stably low until the lowest-frequency components are involved; (2)  From the curve marked by “High”, the performances continually rise, as more high-frequency components are involved. Therefore, based on observations, we emphasize the importance of lowest-frequency and more high-frequency information in GCL. So, whether we incrementally add in the low-to-high or high-to-low frequency order, as long as lowest-frequency and more high-frequency information is preserved in $\bm{V}$, the results will be better. 

\section{Another Experimental Analysis of GAME Rule}
\label{another}
\begin{wrapfigure}[11]{r}{0.4\textwidth}
    \centering
    \includegraphics[width=0.3\textwidth]{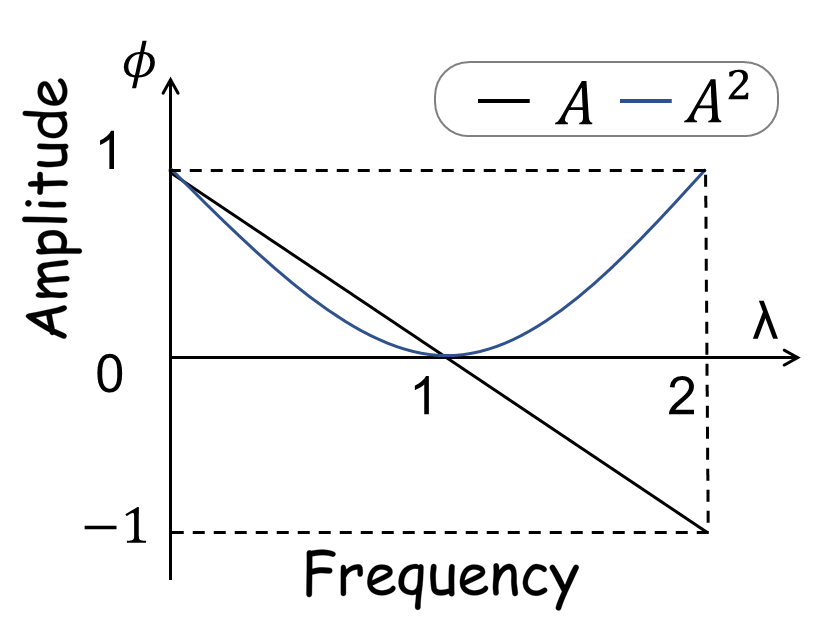}
    \caption{The spectrum of $\bm{A}$ and $\bm{A}^2$.}\label{spectrum_2}
\end{wrapfigure}
We substitute $\bm{A}^2$ (two-hop of $\bm{A}$) for augmentation $\bm{V}$ in the case study. In Fig.~\ref{spectrum_2}, we simultaneously plot the graph spectra of $\bm{A}$ and $\bm{A}^2$. It can be seen that $\bm{A}$ and $\bm{A}^2$ meet GAME rule. Next, we compare the performances of this pair with two other pairs: $\bm{A}$ and $\bm{A}$, $\bm{A}^2$ and $\bm{A}^2$, where the spectra of two augmentations are the same and not meet GAME rule in each pair. We randomly run 10 times for each pairs, and report the average accuracy on four datasets. The results are shown in Table~\ref{three}, where we can see that the pair of $\bm{A}$ and $\bm{A}^2$ actually excels the other two pairs. 

\begin{table*}[h]
    \centering
      \caption{Performance of there different pairs of augmentations to verify the GAME rule.}
      \label{three}
    \begin{tabular}{c|c|c|c|c}
      \bottomrule
         Pair of Augmentations   & Cora & Citeseer & BlogCatalog & Flickr\\
      \bottomrule
         $\bm{A}\ \&\ \bm{A}$& 37.0$\pm$6.1 & 35.4$\pm$3.9 &50.6$\pm$3.2 &26.6$\pm$2.6\\
         $\bm{A}\ \&\ \bm{A}^2$& \textbf{53.7$\pm$3.2} &\textbf{44.7$\pm$5.0} &\textbf{63.1$\pm$4.6} &\textbf{33.7$\pm$2.3}\\
         $\bm{A}^2\ \&\ \bm{A}^2$& 33.3$\pm$2.1 & 35.8$\pm$4.1 & 56.2$\pm$2.1 &28.2$\pm$1.6\\
        \hline
      \end{tabular}
\end{table*}

\section{Experimental Details}
\label{Experimental Details}
\subsection{Implementation details}
\label{Implementation Details}
For two semi-supervised GNN models (GCN, GAT), we utilize the original codes from their authors, and train the models in an end-to-end way. For six graph contrastive learning methods, we obey the traditional GSL setting: we first train each model by optimizing corresponding objective, and get the embeddings for all nodes; then, we feed the node embeddings into a logistic regression to evaluate the quality of embeddings. We use the source codes provided by authors, and follow the settings in their original papers with carefully tune.

For the proposed SpCo, we search on combination coefficient $\eta$ from 0.1 to 1.2 with step 0.1. For $\epsilon$, we test it ranging from \{1e-1, 1e-2, 1e-3, 1e-4\}. We tune the iteration $Iter$ for Sinkhorn's Iteration in algorithm~\ref{Sinkhorn's Iteration Process} from \{1, 2, 3\}. Finally, we carefully select total epochs $T$ and update epochs $\Omega$ for target model, and tune the scope $\mathbb{S}$ from one-hop or two-hop neighbors.

For fair comparisons, we randomly run 10 times and report the average results for all methods. For the reproducibility, we report the related parameters in Appendix~\ref{app_parameter}.

\subsection{Datasets and baselines}
\label{app_dataset}
We choose the five commonly used Cora, Citeseer, Pubmed \citep{gcn}, BlogCatalog and Flickr \citep{meng2019co} for evaluation. The first three datasets are citation networks, where nodes represent papers, edges are the citation relationship between papers, node features are comprised of bag-of-words vector of the papers and labels represent the fields of papers. For them, we choose 500 nodes for validation, 1000 nodes for test. The BlogCatalog dataset is a social network with bloggers and their social relationships from the BlogCatalog website. Node features are constructed by the keywords of user profiles, and the labels represent the topic categories provided by the authors. The Flickr dataset is an image and video hosting website, where users interact with each other via photo sharing. It is a social network where nodes represent users and edges represent their relationships. For these two methods, We choose 1000 nodes for validation, 1000 nodes for test. We uniformly select three label rates for the training set (i.e., 5, 10, 20 labeled nodes per class) for all datasets. Please notice that we adopt classical splits \citep{gcn} on the first three datasets for the 20 labeled nodes per class setting. The details of these datasets are summarized in table~\ref{statistics}.

\begin{table}[h]
  \centering
  \caption{The statistics of the datasets.}
  \label{statistics}
  \begin{tabular}{c|ccccccc}
    \bottomrule
    Dataset & Nodes & Edges & Classes & Features & Training & Validation & Test\\
    \bottomrule
    Cora & 2708 & 10556 & 7 & 1433 & 35/70/140 & 500 & 1000\\
    Citeseer & 3327 & 9228 & 6 & 3703 & 30/60/120 & 500 & 1000\\
    BlogCatalog & 5196 & 343486 & 6 & 8189 & 30/60/120 & 1000 & 1000\\
    Flickr & 7575 & 479476 & 9 & 12047 & 45/90/180 & 1000 & 1000\\
    Pubmed & 19717 & 88651 & 3 & 500 & 15/30/60 & 500 & 1000\\
    \bottomrule
\end{tabular}
\end{table}
These five datasets used in experiments can be found in these URLs:
\begin{itemize}
    \item Cora, Citeseer, Pubmed: \url{https://github.com/tkipf/gcn}
    \item BlogCatalog, Flickr: \url{https://github.com/mengzaiqiao/CAN}
\end{itemize}
And the publicly available implementations of Baselines can be found at the following URLs:
\begin{itemize}
    \item GCN: \url{https://github.com/tkipf/gcn}
    \item GAT: \url{https://github.com/PetarV-/GAT}
    \item DGI: \url{https://github.com/PetarV-/DGI}
    \item MVGRL: \url{https://github.com/kavehhassani/mvgrl}
    \item GRACE: \url{https://github.com/CRIPAC-DIG/GRACE}
    \item GCA: \url{https://github.com/CRIPAC-DIG/GCA}
    \item GraphCL: \url{https://github.com/Shen-Lab/GraphCL}
    \item CCA-SSG: \url{https://github.com/hengruizhang98/CCA-SSG}
\end{itemize}

\subsection{Hyper-parameters settings}
\label{app_parameter}
We implement SpCo in PyTorch, and list some important parameter values in our model in table~\ref{para_1}.

\begin{table}[h]
  \caption{The values of parameters used in SpCo on Cora, Citeseer, BlogCatalog and Flickr.}
  \label{para_1}
  \resizebox{\textwidth}{!}{
  \begin{tabular}{c|c|cccccc}
        \bottomrule
       Method & Dataset & $T$ / patience for early stopping & $\Omega$ & $\mathbb{S}$(hop) & $\eta$ & $\epsilon$ & $Iter$ \\
        \bottomrule
       \multirow{4}{*}{DGI+SpCo}
       & Cora & 40 (patience) & 20 & 1 & 0.1 & 1.0 & 3 \\
       & Citeseer & 30 (patience) & 20 & 1 & 0.5 & 1.0 & 3 \\
       & BlogCatalog & 30 (patience) & 30 & 2 & 0.3 & 1e-2 & 3\\
       & Flickr & 30 (patience) & 30 & 1 & 0.3 & 1e-2 & 2\\
        \hline
       \multirow{4}{*}{GRACE+SpCo}
       & Cora & 300 ($T$)& 30 & 1 & 0.5 & 1.0 & 3 \\
       & Citeseer & 150 ($T$) & 20 & 1 & 1.0 & 1e-2 & 3 \\
       & BlogCatalog & 800 ($T$) & 300 & 1 & 1.0 & 1e-2 & 3\\
       & Flickr & 1300 ($T$) & 300 & 1 & 1.0 & 1e-1 & 2\\
        \hline
       \multirow{4}{*}{CCA-SSG+SpCo}
       & Cora & 40 ($T$)& 15 & 1 & 0.5 & 1e-2 & 3 \\
       & Citeseer & 15 ($T$) & 5 & 1 & 0.1 & 1e-1 & 3 \\
       & BlogCatalog & 75 ($T$) & 10 & 1 & 0.1 & 1e-1 & 3\\
       & Flickr & 100 ($T$) & 30 & 1 & 0.5 & 1e-2 & 2\\
        \bottomrule
        
  \end{tabular}
  }
\end{table}

During experiments, we notice that Pubmed is a large graph, which will consume much time to operate Sinkhorn's Iteration. Meanwhile, we find that our SpCo is model-agnostic, which means that the Sinkhorn's Iteration is independent of some specific model. Therefore, we choose to calculate the iteration results for Pubmed and store them for utilization. Specifically, we firstly calculate 10 intermediate iteration results for Pubmed, and then for each method, we take $num$ intermediate results at equal intervals, and feed them one by one into the target model every $epoch\_inter$ epochs. Therefore, the relative parameter values are shown in table~\ref{para_2}.
\begin{table}[h]
  \caption{The values of parameters used in SpCo on Pubmed.}
  \label{para_2}
  \resizebox{\textwidth}{!}{
  \begin{tabular}{c|ccccccc}
        \bottomrule
       Method & $T$ / patience for early stopping & $num$ & $epoch\_inter$ & $\mathbb{S}$(hop) & $\eta$ & $\epsilon$ & $Iter$ \\
        \bottomrule
        DGI+SpCo & 30 (patience) & 3 & 100 &1& 0.1 & 1e-2 & 1 \\
        \hline
        GRACE+SpCo & 1500 ($T$)& 7 & 100 &1 & 1.0 & 1e-2 & 1 \\
        \hline 
       CCA-SSG+SpCo & 170 ($T$)& 7 & 15 &1 & 0.1 & 1e-2 & 1 \\
        \bottomrule
        
  \end{tabular}
  }
\end{table}

\subsection{More experiment results}
\subsubsection{Node classification}
\label{node classification}
In this section, we provide more experimental results on node classification, where we choose 5 and 10 labeled nodes per class as training set respectively, and keep the same validation and test set. The relative results are given in Table~\ref{fenlei_2}. Again, we can see that additional SpCo generally improves the performance of corresponding baselines on all datasets, which comprehensively demonstrates that SpCo can boost node classification in an effective way. Especially, we notice that the loss used in CCA-SSG is originated from canonical correlation analysis, which is definitely different from InfoNCE loss~\eqref{contra}. But, we can still boost the results of CCA-SSG by plugging SpCo, indicating the necessity to design effective augmentations.
\begin{table*}[t]
  \Huge
  \caption{Quantitative results (\%$\pm\sigma$) on node classification. The better results compared with target model are in bold.}
  \label{fenlei_2}
  \resizebox{\textwidth}{!}{
  \renewcommand\arraystretch{1.5}
  \begin{tabular}{c|c|c|c|c|cc|c|cc|c|c|cc}
    \bottomrule
    Datasets & Metrics & Splits &GCN & GAT & DGI & \textbf{DGI+SpCo} & MVGRL & GRACE & \textbf{GRACE+SpCo} & GCA & GraphCL & CCA-SSG& \textbf{CCA+SpCo}\\
    \bottomrule
    \multirow{4}{*}{Cora}&
    \multirow{2}{*}{Ma-F1}
    &5&68.4±1.4&74.4±0.3&76.1±0.4&\textbf{76.2±0.3}&76.2±0.5&70.5±2.0&\textbf{71.4±2.1}&70.7±2.8&75.1±0.4&73.4±0.7&\textbf{73.7±0.6}\\
    &&10&72.9±0.5&75.7±0.5&77.8±0.7&\textbf{78.2±0.4}&78.4±0.8&73.9±1.5&\textbf{75.7±1.6}&75.5±1.4&77.8±0.7&77.1±0.4&\textbf{78.4±1.0}\\
    \cline{2-14}
    &\multirow{2}{*}{Mi-F1}
    &5&69.6±1.5&75.8±0.4&77.3±0.5&\textbf{77.4±0.4}&77.0±0.7&70.7±1.8&\textbf{71.5±2.1}&71.5±2.8&76.4±0.3&73.6±0.7&\textbf{73.8±0.8}\\
    &&10&73.5±0.6&76.5±0.5&79.1±0.6&\textbf{79.3±0.5}&79.4±0.8&74.9±1.6&\textbf{76.5±1.3}&76.4±1.7&79.0±0.8&77.4±0.5&\textbf{78.7±1.5}\\
    \hline
    \multirow{4}{*}{Citeseer}&
    \multirow{2}{*}{Ma-F1}
    &5&48.2±1.5&55.3±0.4&62.6±1.2&\textbf{63.3±0.6}&58.3±0.9&58.1±1.5&\textbf{59.2±1.3}&52.3±2.5&62.1±1.1&57.2±1.4&\textbf{59.0±2.1}\\
    &&10&63.8±0.7&64.6±0.5&64.9±0.7&\textbf{65.0±0.5}&62.8±0.8&61.9±0.9&\textbf{62.1±0.5}&59.3±1.4&64.4±0.6&63.6±1.1&\textbf{65.2±0.5}\\
    \cline{2-14}
    &\multirow{2}{*}{Mi-F1}
    &5&53.4±1.1&61.8±1.9&67.4±1.5&\textbf{68.0±0.6}&63.0±1.1&62.5±1.5&\textbf{63.9±1.6}&54.6±3.0&66.8±1.3&61.4±1.5&\textbf{62.5±2.1}\\
    &&10&66.8±0.8&67.9±0.6&68.9±0.5&\textbf{69.2±0.4}&67.2±0.8&65.2±0.8&\textbf{66.0±1.0}&61.8±1.7&68.5±0.7&67.5±1.1&\textbf{69.7±0.9}\\
    \hline
    \multirow{4}{*}{BlogCatalog}&
    \multirow{2}{*}{Ma-F1}
    &5&61.4±2.8&55.7±3.9&60.2±1.8&\textbf{65.9±1.3}&69.0±7.1&56.6±1.5&\textbf{58.4±0.6}&62.5±0.8&55.6±3.1&67.9±0.4&\textbf{68.2±0.5}\\
    &&10&68.2±1.8&65.0±1.3&67.1±1.1&\textbf{70.5±0.4}&76.2±4.0&64.9±0.8&\textbf{65.8±0.9}&69.9±0.4&63.3±1.6&\textbf{67.6±0.5}&67.5±0.3\\
    \cline{2-14}
    &\multirow{2}{*}{Mi-F1}
    &5&63.3±2.7&58.1±3.1&61.8±1.7&\textbf{67.2±1.1}&69.9±7.1&58.9±1.5&\textbf{60.4±0.6}&64.4±0.7&57.2±3.3&69.0±0.4&\textbf{69.2±0.4}\\
    &&10&69.5±1.8&65.9±1.5&67.7±1.2&\textbf{71.3±0.5}&76.7±3.9&65.8±0.8&\textbf{66.8±1.0}&70.8±0.5&63.9±1.7&68.3±0.5&\textbf{68.4±0.3}\\
    \hline
    \multirow{4}{*}{Flickr}&
    \multirow{2}{*}{Ma-F1}
    &5&32.4±1.6&27.9±1.3&20.9±1.3&\textbf{24.3±1.0}&21.7±3.0&21.0±5.5&\textbf{31.3±1.0}&34.6±0.4&22.3±1.3&30.6±1.1&\textbf{31.4±1.0}\\
    &&10&42.8±1.1&32.8±2.3&26.2±2.3&\textbf{26.8±0.6}&29.1±3.8&32.6±1.7&\textbf{33.5±4.1}&39.3±0.6&25.9±2.2&32.3±0.6&\textbf{33.9±0.2}\\
    \cline{2-14}
    &\multirow{2}{*}{Mi-F1}
    &5&35.4±1.0&30.9±0.2&21.9±1.0&\textbf{26.6±1.1}&23.0±3.0&23.6±4.2&\textbf{32.7±0.9}&35.7±0.4&23.4±1.1&33.4±0.8&\textbf{33.8±0.1}\\
    &&10&44.2±1.0&35.3±1.4&27.1±2.0&\textbf{29.8±0.6}&30.5±3.8&34.9±1.0&\textbf{35.9±2.4}&40.4±0.5&27.3±1.9&35.2±0.9&\textbf{36.2±0.3}\\
    \hline
    \multirow{4}{*}{PubMed}&
    \multirow{2}{*}{Ma-F1}
    &5&69.3±0.7&70.1±0.4&72.3±1.0&\textbf{73.0±0.4}&73.8±0.7&\textbf{75.5±0.7}&75.4±0.6&70.9±1.5&70.7±0.4&72.1±0.8&\textbf{73.1±0.4}\\
    &&10&73.1±1.1&72.5±0.3&73.2±0.7&\textbf{74.2±0.5}&72.9±0.5&73.2±1.6&\textbf{73.4±1.4}&75.0±0.1&72.3±0.4&73.0±1.0&\textbf{74.2±0.6}\\
    \cline{2-14}
    &\multirow{2}{*}{Mi-F1}
    &5&68.4±0.8&69.8±0.4&72.2±1.2&\textbf{73.0±0.6}&73.2±1.2&75.5±0.7&\textbf{76.0±0.6}&70.5±1.8&70.4±0.6&71.5±0.8&\textbf{72.7±0.4}\\
    &&10&72.7±1.3&72.2±0.4&72.5±0.9&\textbf{73.3±0.6}&71.9±0.6&72.1±1.9&\textbf{72.9±1.6}&74.6±0.2&71.3±0.5&71.9±1.2&\textbf{73.3±0.7}\\
    \hline
  \end{tabular}}
\end{table*}

\subsubsection{Visualisation of graph spectrum}
\label{isualisation of graph spectrum}
\begin{figure}[htbp]
\centering
\subfigure[DGI: Cora]
{
    \begin{minipage}[b]{.3\linewidth}
        \centering
        \includegraphics[scale=0.23]{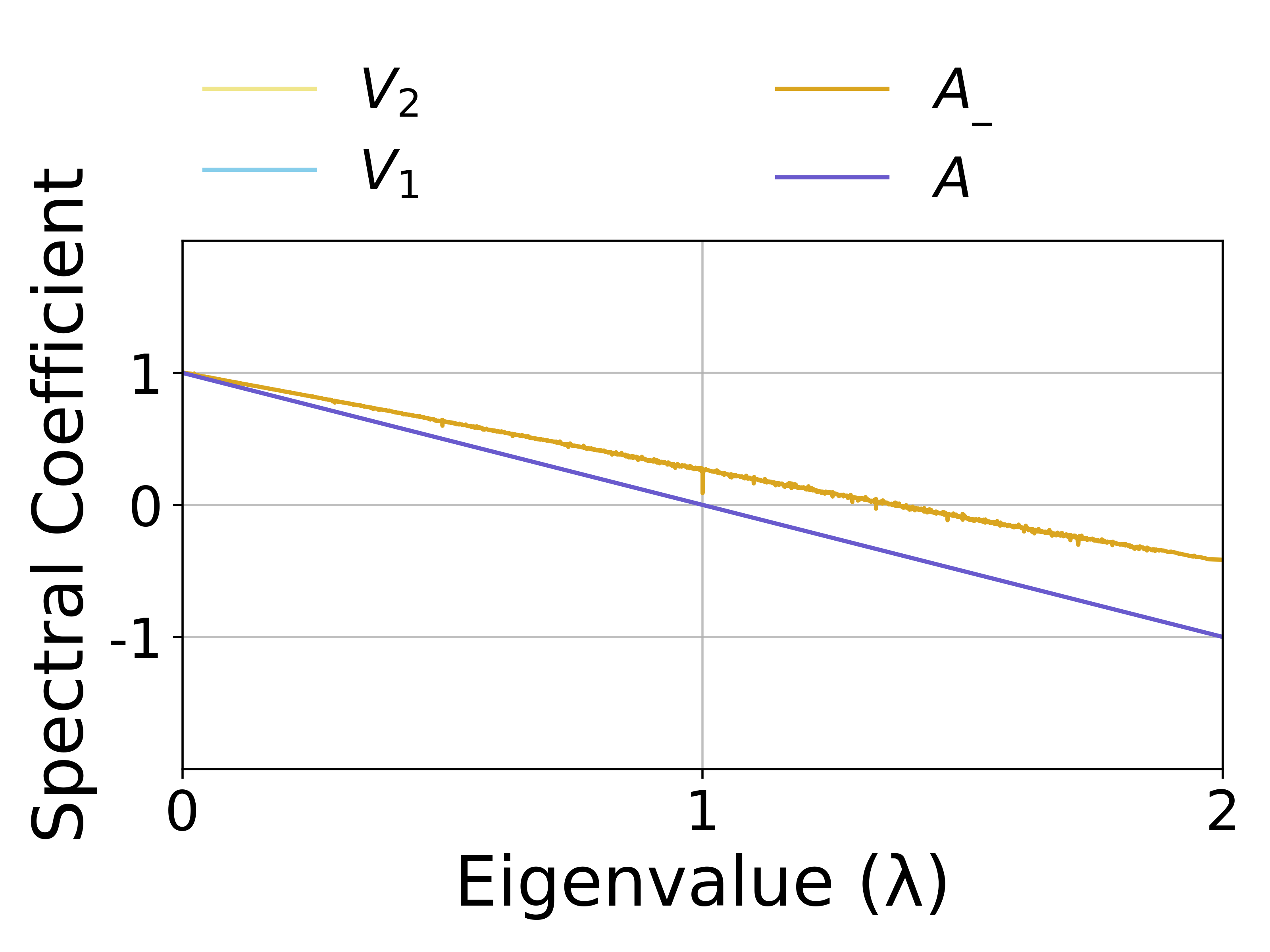}
    \end{minipage}
}
\subfigure[GRACE: Cora]
{
 	\begin{minipage}[b]{.3\linewidth}
        \centering
        \includegraphics[scale=0.23]{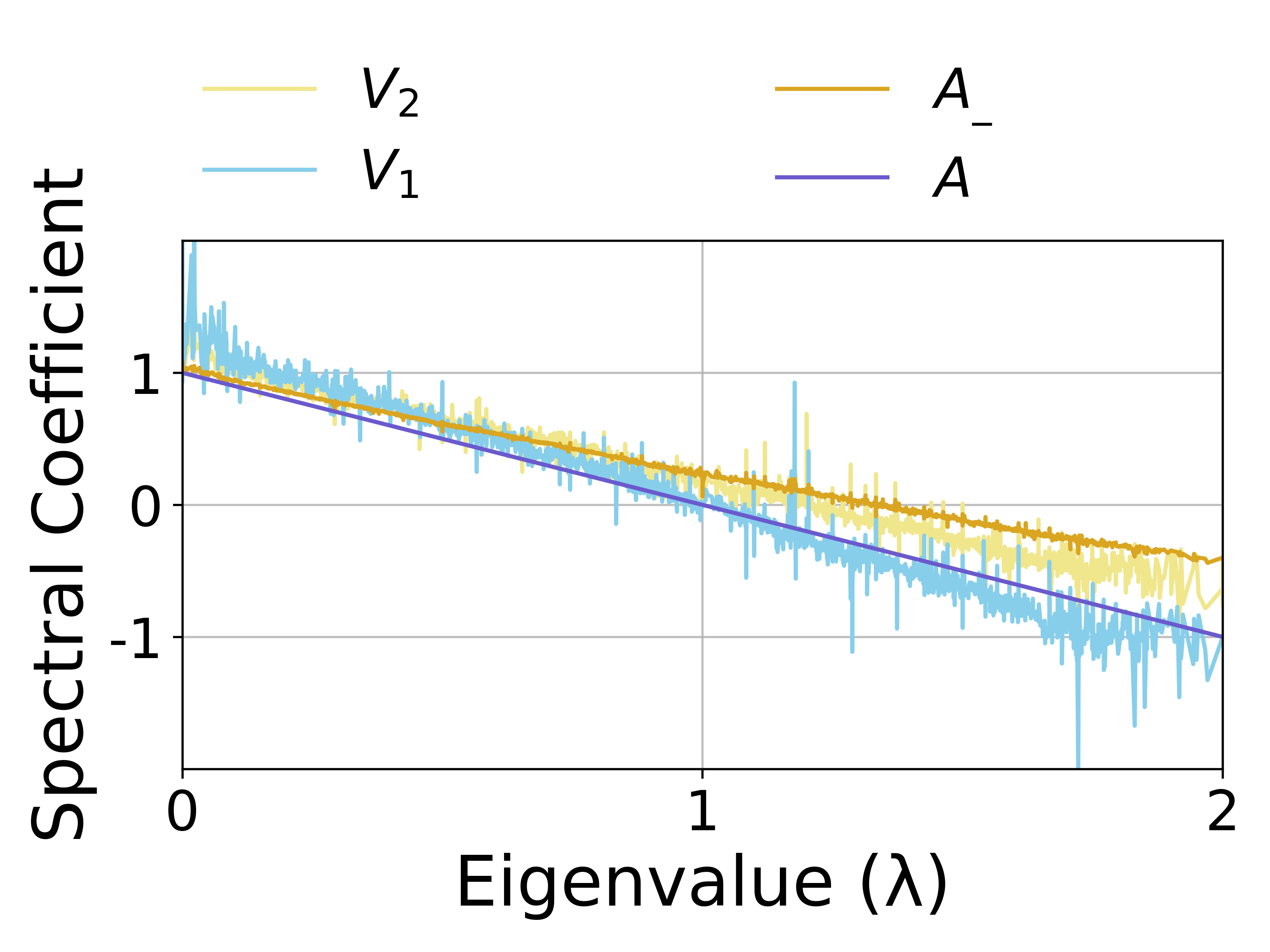}
    \end{minipage}
}
\subfigure[CCA-SSG: Cora]
{
 	\begin{minipage}[b]{.3\linewidth}
        \centering
        \includegraphics[scale=0.23]{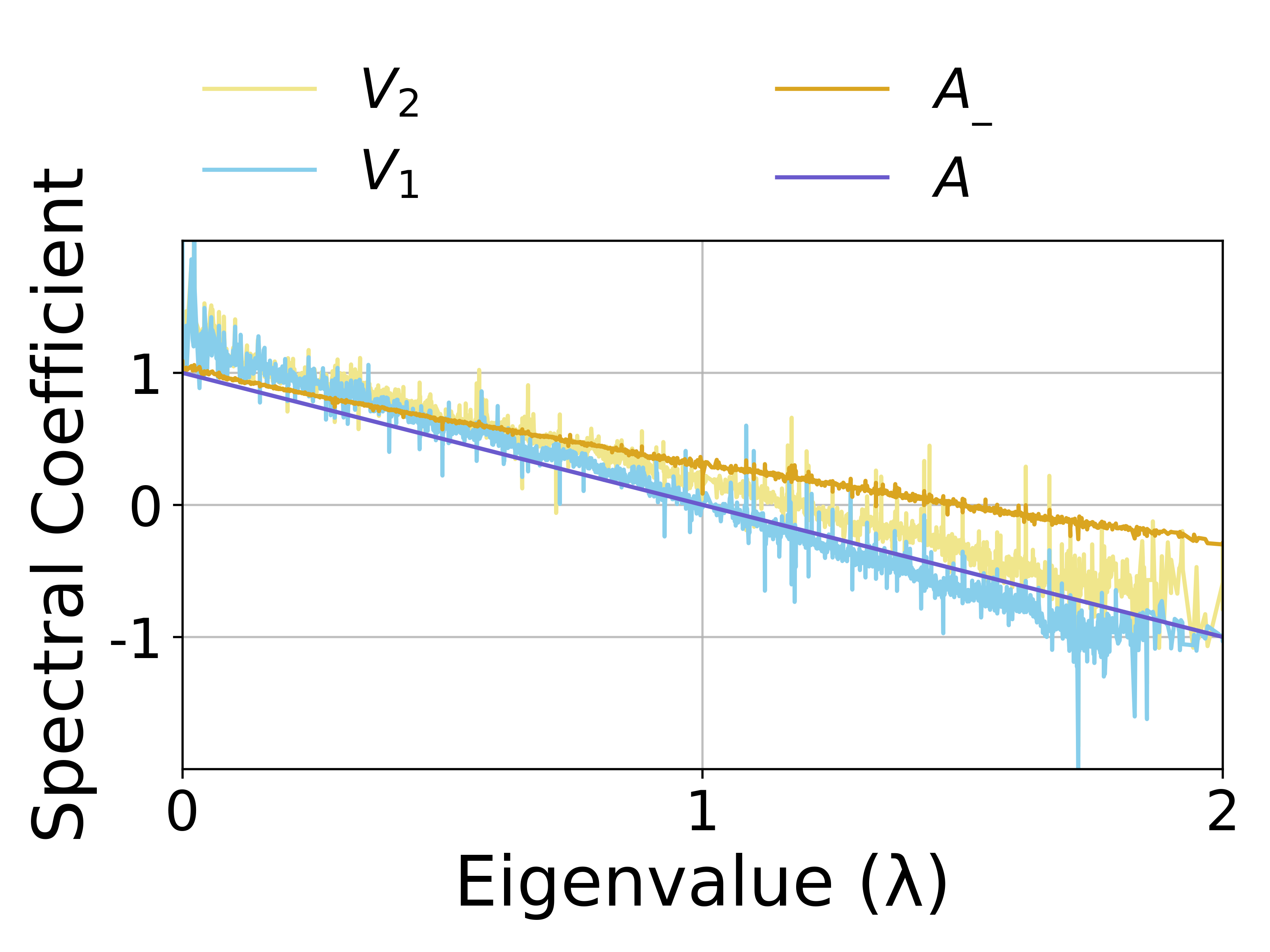}
    \end{minipage}
}
\caption{The visualisation of graph spectrum on Cora.}
\label{citeseer}
\end{figure}

In this section, we mainly supplement the visualisation of graph spectrum on Cora, shown in Fig.~\ref{citeseer}. Again, $\bm{A\_}$ and $\bm{A}$ have a smaller difference in $\mathcal{F_L}$ than in $\mathcal{F_H}$, indicating they satisfy the GAME rule. And $\bm{V_1}$ and $\bm{V_2}$ got from them also present the same pattern. Therefore, with contrastive invariance theorem~\ref{case_theo}, they will instruct encoder to capture more low-frequency information, and hence improve the results.

\subsubsection{Hyper-parameter sensitivity}
\label{Hyper-parameter Sensitivity}
\begin{figure}[htbp]
\centering
\subfigure[DGI: Cora]
{
    \begin{minipage}[b]{.3\linewidth}
        \centering
        \includegraphics[scale=0.25]{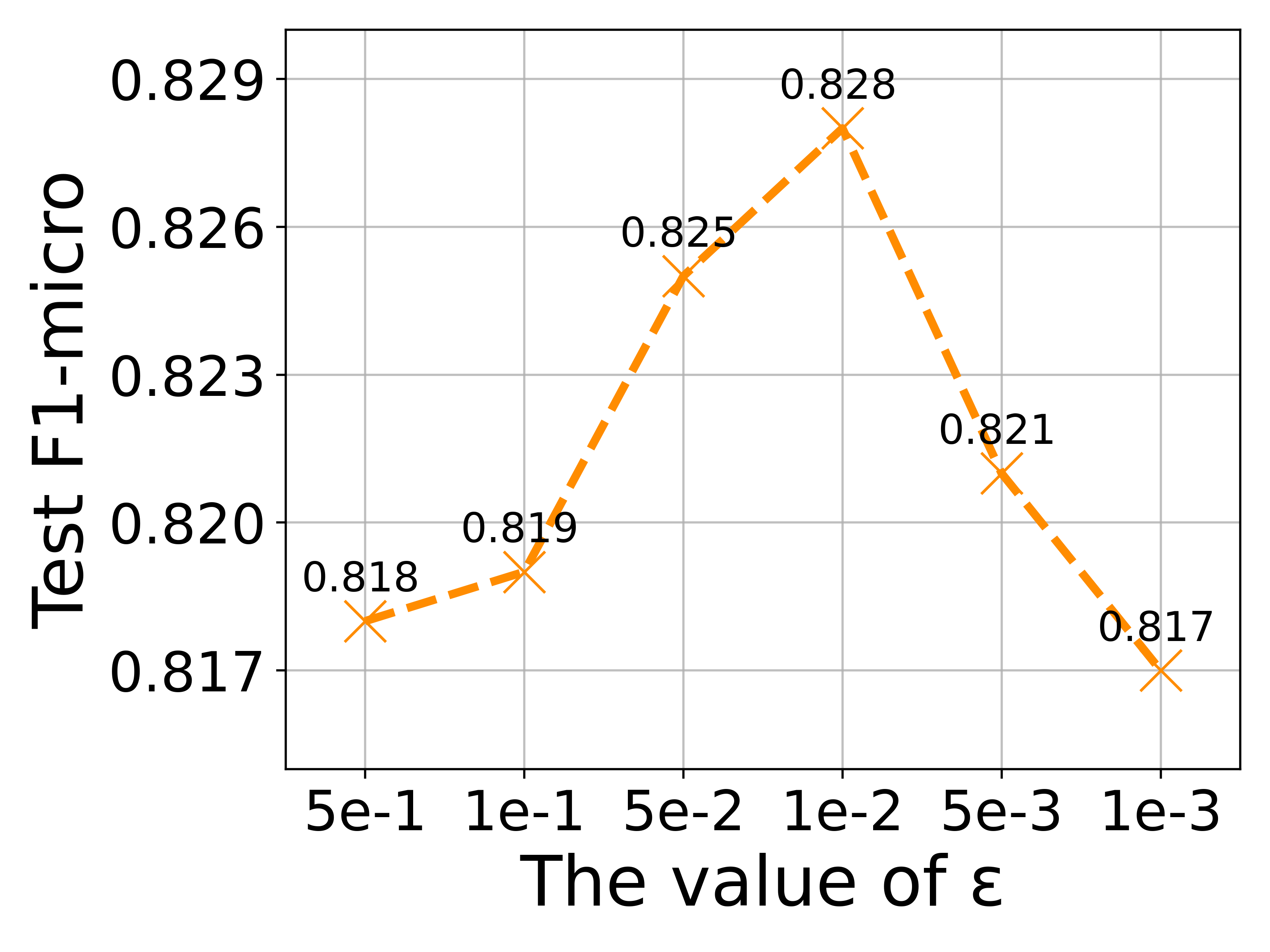}
    \end{minipage}
}
\subfigure[GRACE: Cora]
{
 	\begin{minipage}[b]{.3\linewidth}
        \centering
        \includegraphics[scale=0.25]{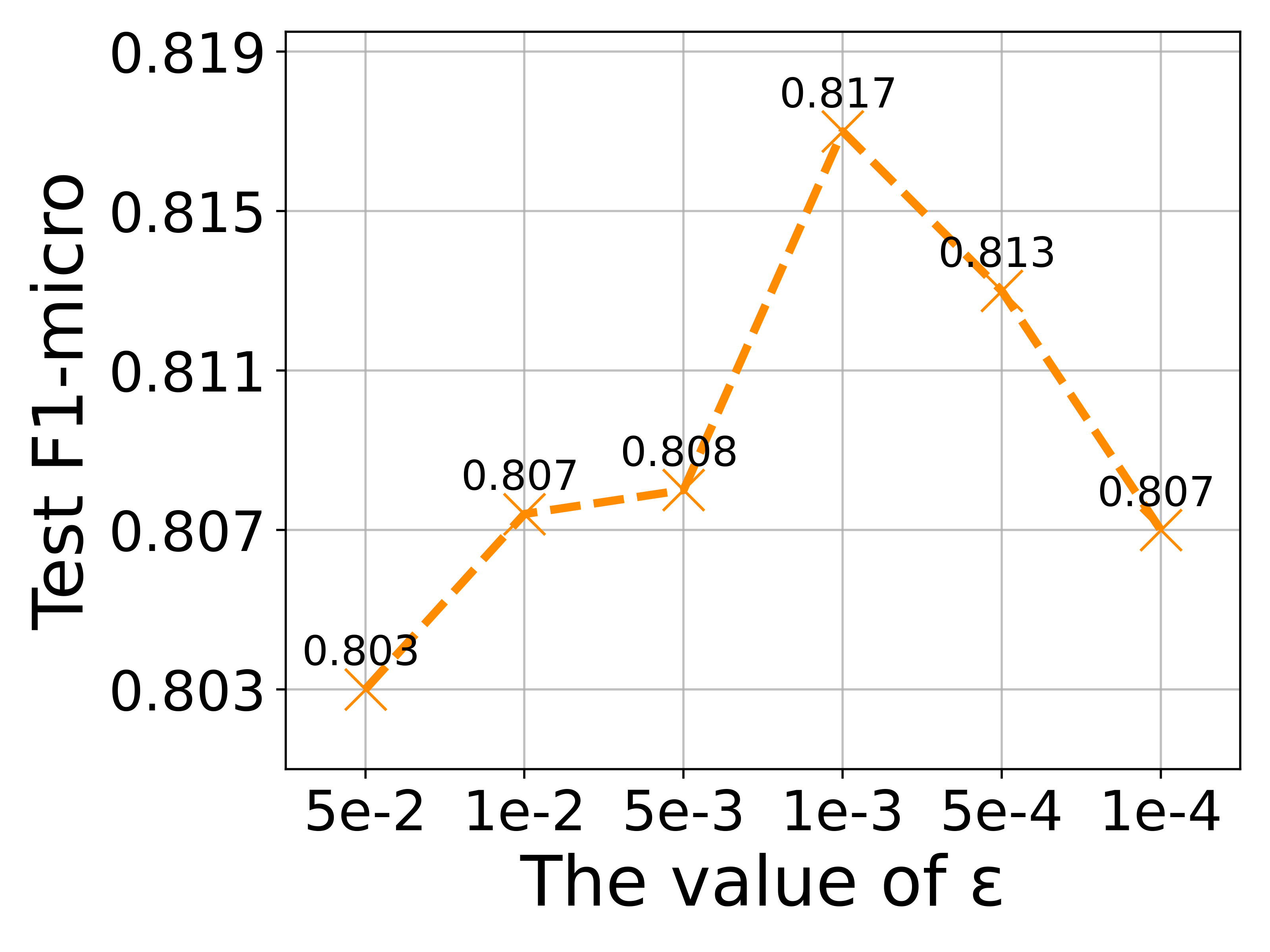}
    \end{minipage}
}
\subfigure[CCA-SSG: Cora]
{
 	\begin{minipage}[b]{.3\linewidth}
        \centering
        \includegraphics[scale=0.25]{fig/epi_blog_cca.png}
    \end{minipage}
}
\caption{Analysis of the hyper-parameter $\epsilon$ on Cora.}
\label{epi_cora_}
\end{figure}

In this section, we show more results about $\epsilon$ on Cora, shown in Fig.~\ref{epi_cora_}. Then, we give more analyses about hyper-parameter $\eta$, and test its sensitivity on Cora and BlogCatalog.

\begin{figure}[htbp]
\centering
\subfigure[DGI: Cora]
{
    \begin{minipage}[b]{.3\linewidth}
        \centering
        \includegraphics[scale=0.25]{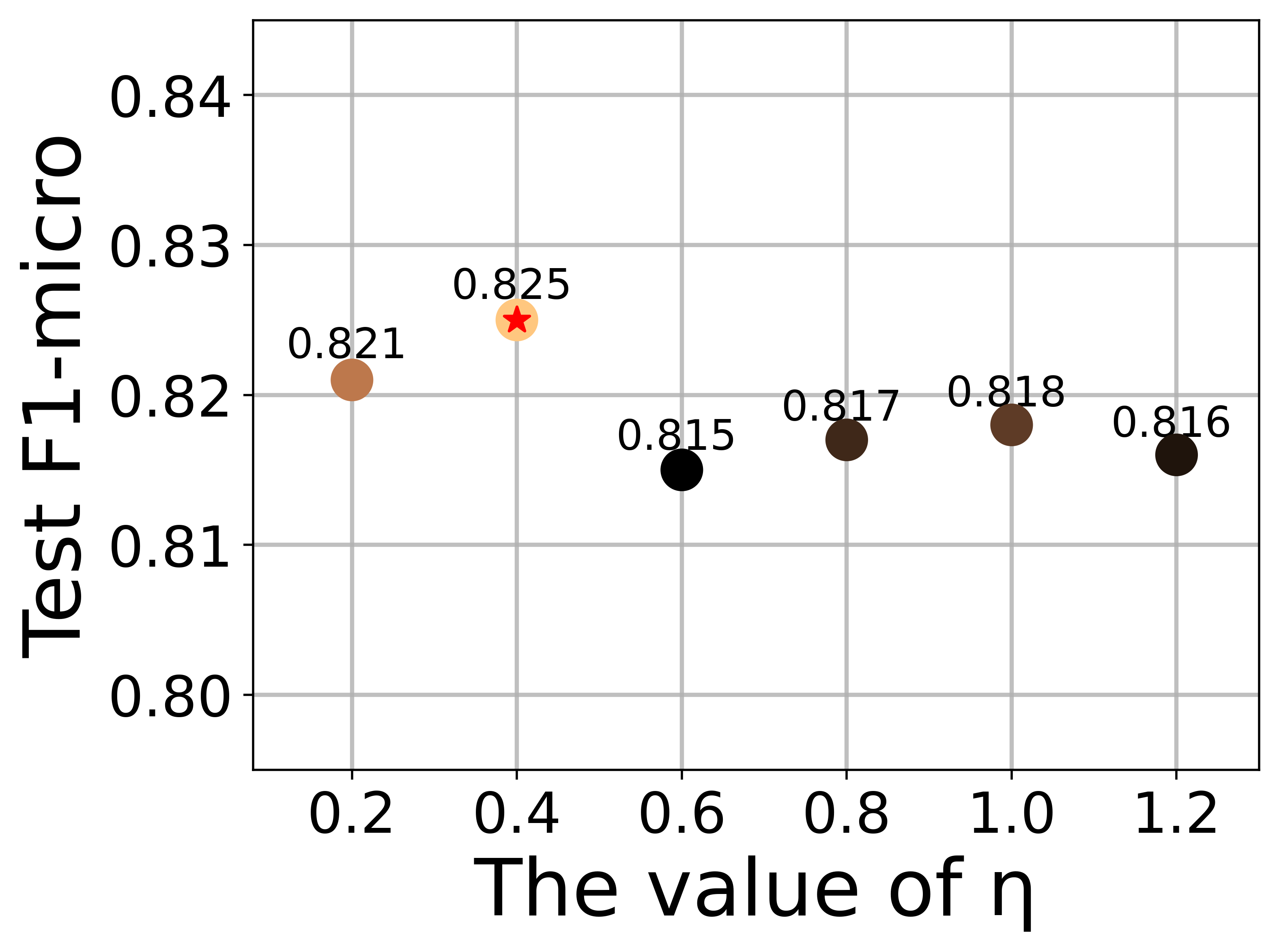}
    \end{minipage}
}
\subfigure[GRACE: Cora]
{
 	\begin{minipage}[b]{.3\linewidth}
        \centering
        \includegraphics[scale=0.25]{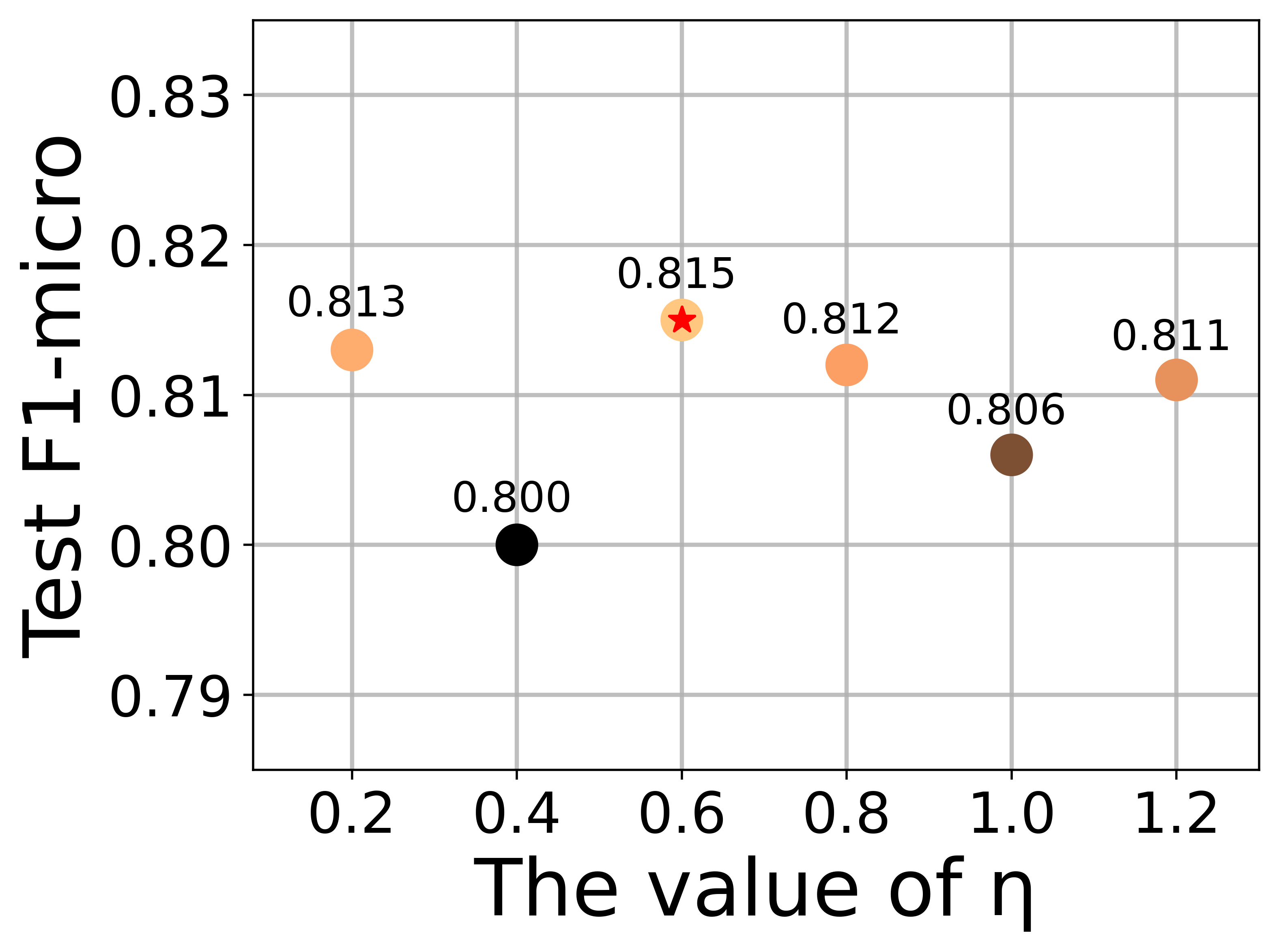}
    \end{minipage}
}
\subfigure[CCA-SSG: Cora]
{
 	\begin{minipage}[b]{.3\linewidth}
        \centering
        \includegraphics[scale=0.25]{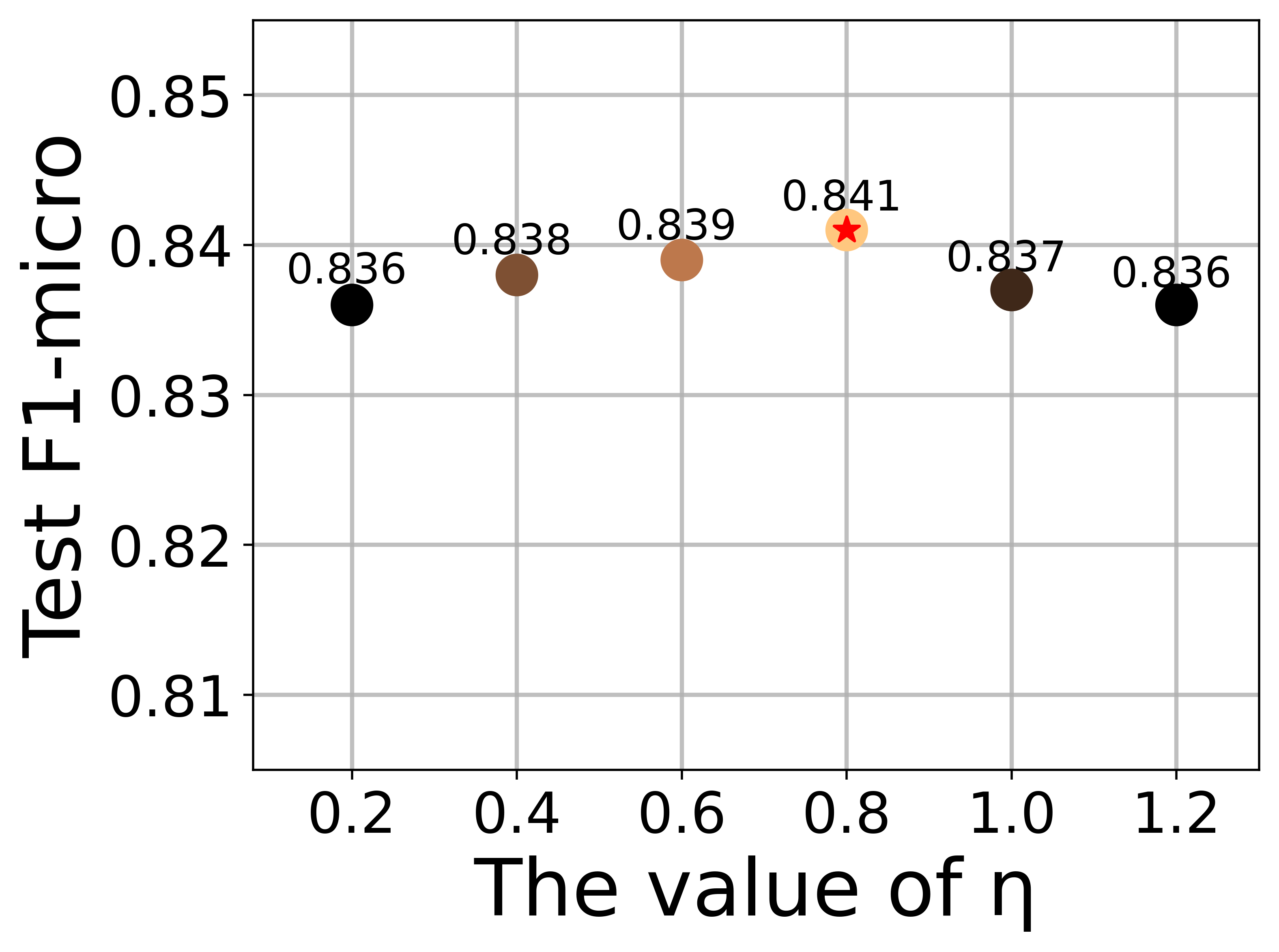}
    \end{minipage}
}
\subfigure[DGI: BlogCatalog]
{
    \begin{minipage}[b]{.3\linewidth}
        \centering
        \includegraphics[scale=0.25]{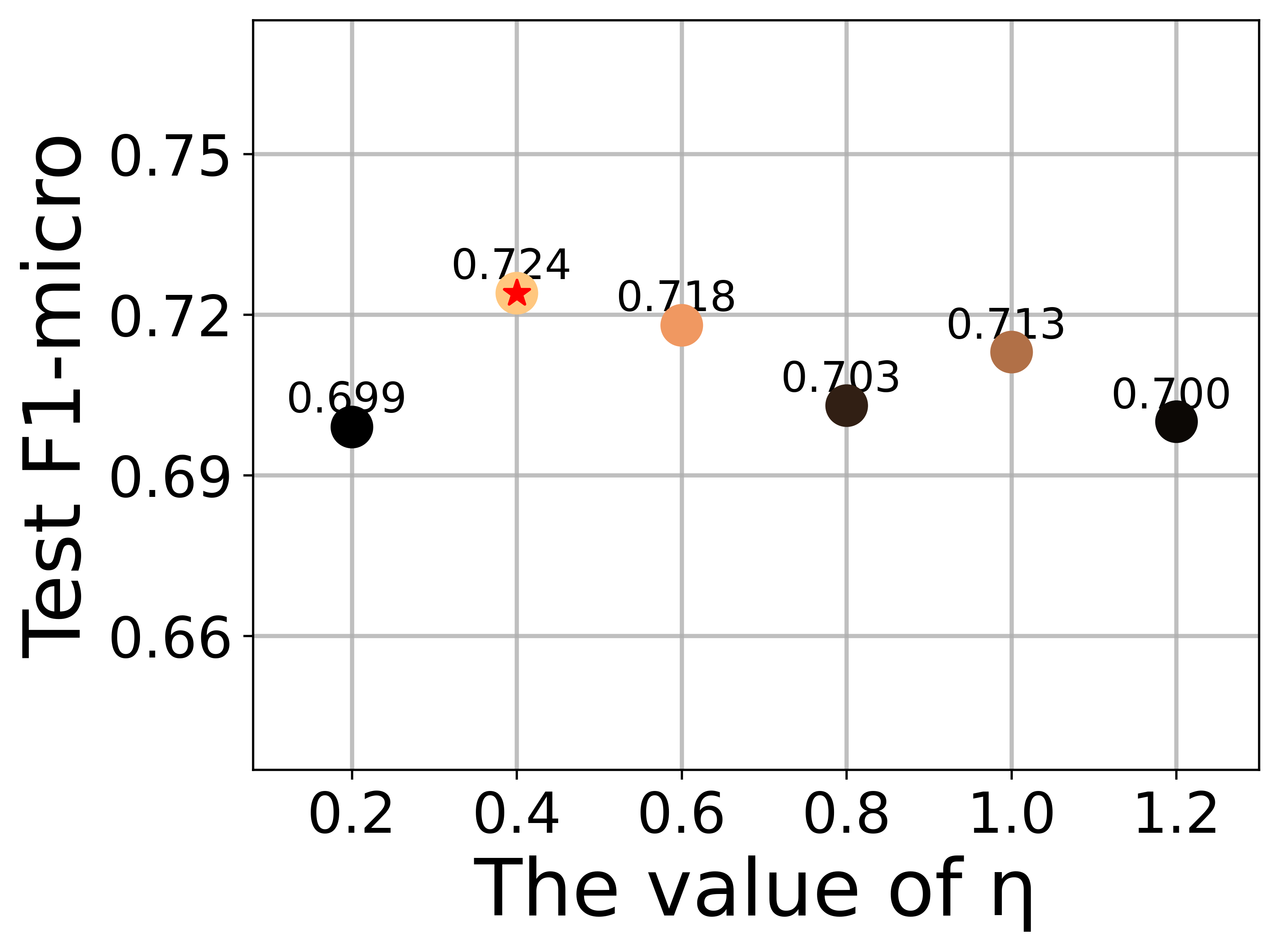}
    \end{minipage}
}
\subfigure[GRACE: BlogCatalog]
{
 	\begin{minipage}[b]{.3\linewidth}
        \centering
        \includegraphics[scale=0.25]{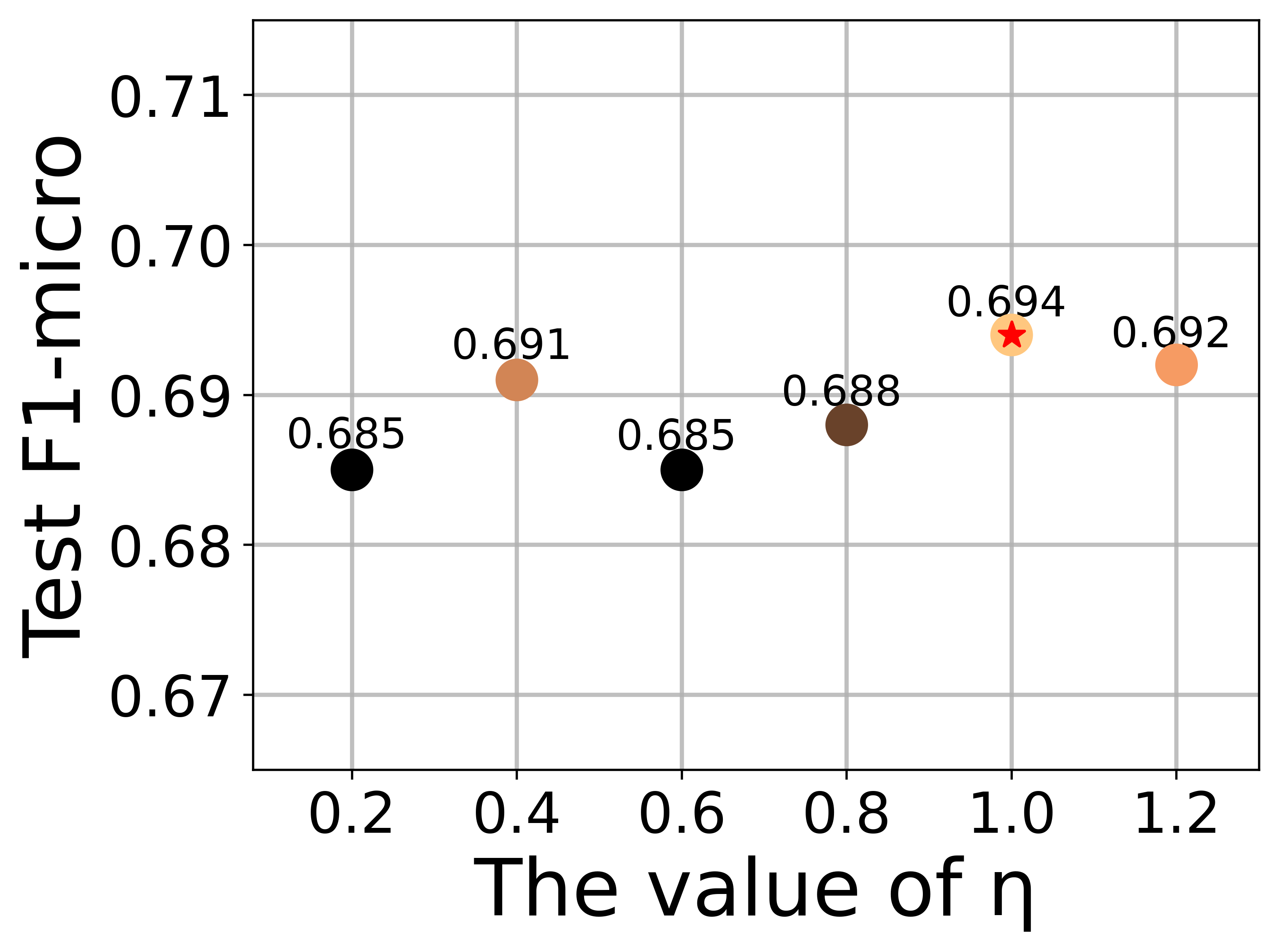}
    \end{minipage}
}
\subfigure[CCA-SSG: BlogCatalog]
{
 	\begin{minipage}[b]{.3\linewidth}
        \centering
        \includegraphics[scale=0.25]{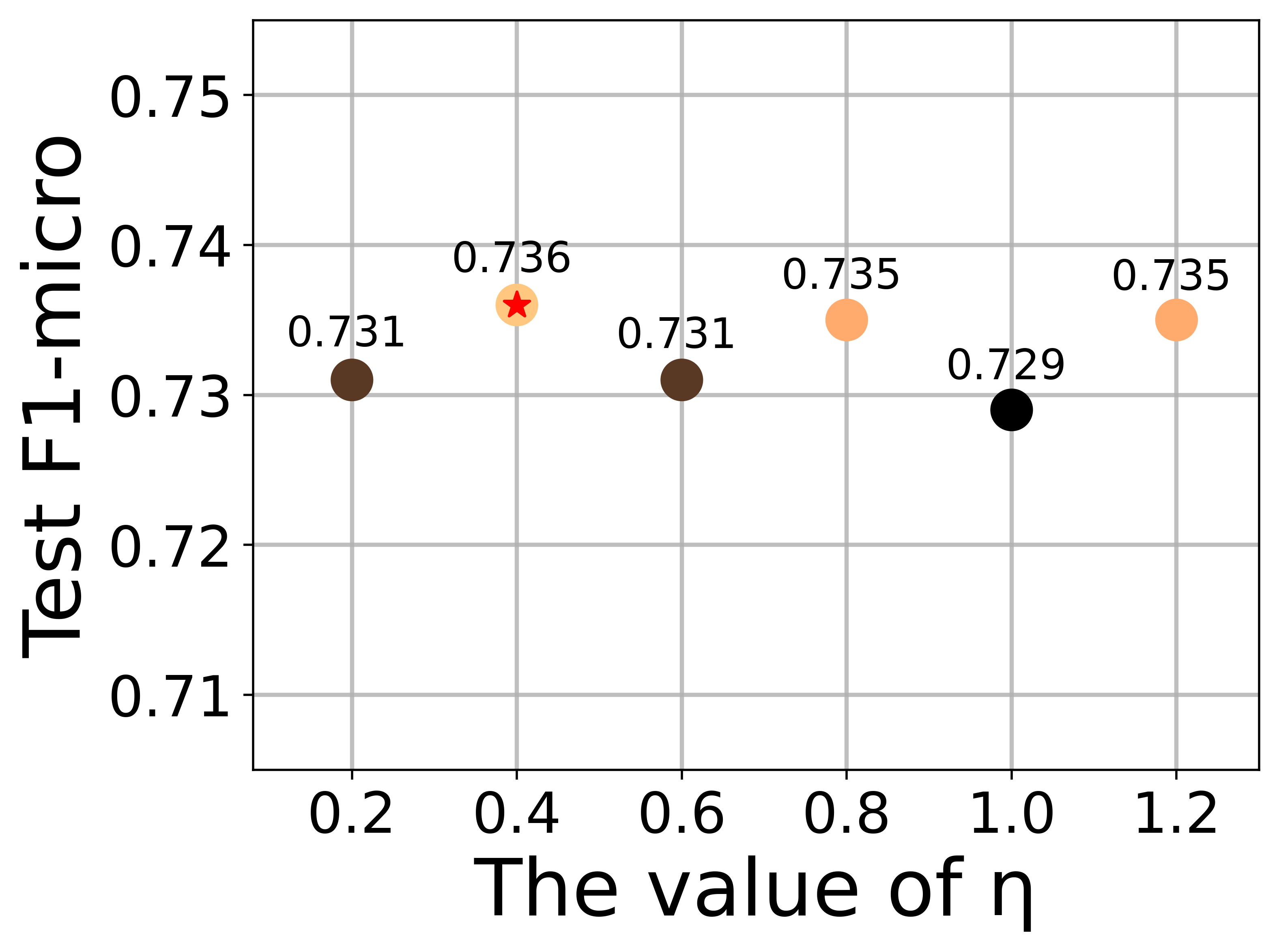}
    \end{minipage}
}
\caption{Analysis of the hyper-parameter $\eta$ on Cora and BlogCatalog.}
\label{lam_cora}
\end{figure}

\textbf{Analysis of $\eta$.} The $\eta$ in eq.~\eqref{ete} controls the strength that $\Delta_{\bm{A}}$ changes the original $\bm{A}$. If $\eta$ is larger, $\Delta_{\bm{A}}$ will give more impact on $\bm{A}$, and the margin between $\bm{A\_}$ and $\bm{A}$ is larger in $\mathcal{F_H}$. We again range the value of $\eta$ from 0.2 to 1.2, and the results are shown in Fig~\ref{lam_cora}. In the figure, the shallower color is, the better performance is. And the maximal value is marked as red star. We can see that the performances are relative stable in this range. Please note that, this range is reasonable. If the lower $\eta$ is tested, our SpCo will have a smaller influence. And if the higher $\eta$ is tested, our SpCo will dominate the combination in eq.~\eqref{ete}, which will destroy the original $\bm{A}$ heavily.

\subsection{Operating environment}
\label{Operating Environment}
The environment where our code runs is shown as follows:
\begin{itemize}
    \item Operating system: Linux version 3.10.0-693.el7.x86\_64
    \item CPU information: Intel(R) Xeon(R) Silver 4210 CPU @ 2.20GHz
    \item GPU information: GeForce RTX 3090
\end{itemize}

\subsection{Detailed descriptions of target models}
\label{app_describ}
As introduced in experiments, we plug our SpCo into three existing GCL frameworks, including DGI \citep{dgi}, GRACE \citep{grace} and CCA-SSG \citep{cca}, to roundly verify the effectiveness of SpCo. Therefore, it is necessary to briefly introduce their mechanisms in this section.
\begin{itemize}
    \item \textbf{DGI}: This method deploys contrastive learning between local patch and summary vector. Specifically, the authors utilize GCN to encode the original graph and get all node embeddings, which is viewed as local patch. Then, they design a readout function to summarize node embeddings into one vector, which represents the global information of the graph. In optimization, DGI views every node in the graph as the positive sample of the summary vector, while views nodes in corrupted graph as negative samples. Different from InfoNCE loss, DGI uses BCE loss as the objective.
    \item \textbf{GRACE}: This method uses feature mask and random edge dropping as two augmentation strategies from feature and topology levels. Specifically, the authors perform two strategies at the same time but with different ratios to generate two augmentations, respectively. Then, they obey traditional InfoNCE loss, where the positive sample of one node is just it own embedding in the other augmentation, and other nodes are viewed as negative samples.
    \item \textbf{CCA-SSG}: This method inherits previous augmentation strategies, like feature mask and random edge dropping. However, the main contribution of this work is to involve  canonical correlation analysis into GCL framework and propose a new objective, which maximizes the correlation between two views and prevent degenerated solutions simultaneously. 
\end{itemize}

\section{Related Work}
\label{related work}
\textbf{Graph Neural Networks.}
Recently, Graph Neural Networks (GNNs) have attracted considerable attentions, which can be broadly divided into two categories, spectral-based and spatial-based. Spectral-based GNNs are inheritance of graph signal processing, and define graph convolution operation in spectral domain. For example, \cite{spe1} utilizes Fourier bias to decompose graph signals; \cite{spe2} employs the Chebyshev expansion of the graph Laplacian to improve the efficiency. For another line, spatial-based GNNs greatly simplify above convolution by only focusing on neighbors. For example, GCN \cite{gcn} simply averages information of one-hop neighbors. GraphSAGE \cite{graphsage} only randomly fuses a part of neighbors with various poolings. GAT \cite{gat} assigns different weights to different neighbors. More detailed surveys can be found in \cite{wu2021comprehensive}.

\textbf{Graph Contrastive Learning.}
Graph Contrastive Learning has shown its distinguished capacity in unsupervised setting, and many studies have been proposed. In this paper, we mainly focus on the various augmentation strategies. Specifically, DGI \citep{dgi} contrasts between local node embedding and global summary vector. MVGRL \citep{mvgrl} proposes several strategies based on diffusion or  distance matrix. For \{GRACE \citep{grace}, GCA \citep{gca}, GraphCL \citep{graphcl}, CCA-SSG \citep{cca}\}, they can roughly be gathered into the same category: the random edge and node perturbation. There are some frameworks aiming to learn an adaptive augmentation with the help of different principles, such as AD-GCL \citep{adgcl}, InfoGCL \citep{infogcl}, DGCL \citep{dgcl} and GASSL \cite{yang2021graph}. But these studies mainly focus on graph classification task. Besides, we also notice that some papers propose augmentation free \citep{lee2021augmentation}.
\end{document}